\documentclass{article}

\usepackage[final]{neurips_data_2021}

\usepackage{listings}
\usepackage{booktabs} 
\usepackage[utf8]{inputenc} %
\usepackage[T1]{fontenc}    %
\usepackage{hyperref}       %
\usepackage{url}            %
\usepackage{booktabs}       %
\usepackage{amsfonts}       %
\usepackage{nicefrac}       %
\usepackage{microtype}      %
\usepackage{float}
\usepackage{wrapfig} %
\usepackage{balance} %
\usepackage{multicol}

\usepackage{amsmath,amsfonts,bm}

\def\eqref#1{equation~\ref{#1}}

\def\1{\bm{1}}

\DeclareMathAlphabet{\mathsfit}{\encodingdefault}{\sfdefault}{m}{sl}
\SetMathAlphabet{\mathsfit}{bold}{\encodingdefault}{\sfdefault}{bx}{n}

\usepackage{cleveref}
\usepackage{ctable}
\usepackage{tabularx,rotating}
\usepackage[frozencache=true,cachedir=_minted-main]{minted} 

\usepackage{tablefootnote}

\crefname{figure}{figure}{figures}
\crefname{equation}{equation}{equations}
\crefname{table}{table}{tables}
\crefname{section}{section}{sections}
\crefname{appendix}{appendix}{appendices}
\crefformat{footnote}{#2\footnotemark[#1]#3}
\usepackage{pifont}
\usepackage{gensymb}
\usepackage{array}
\usepackage{pythonhighlight}
\usepackage{xcolor}
\usepackage{graphicx}
\usepackage{subcaption}
\newcommand{\cmark}{\ding{51}}
\newcommand{\xmark}{\ding{55}}
    \usepackage[
        separate-uncertainty=true,
        detect-weight,
        table-number-alignment = center,
    	table-text-alignment = center,
    	table-format=3.2(4),
        table-sign-mantissa
    ]{siunitx}
\usepackage{etoolbox}
\BeforeBeginEnvironment{wrapfigure}{\setlength{\intextsep}{0pt}}

\usepackage{multirow}
\newcommand*{\myrulefill}[3][]{%
  \makebox[#2]{#1%
    \enskip{#3}\enskip%
    \leaders\hrule height \dimexpr.5ex+.2pt\relax depth \dimexpr -.5ex+.2pt\relax \hfill\kern0pt}%
}

\newcommand{\specialcell}[2][c]{%
  \begin{tabular}[#1]{@{}c@{}}#2\end{tabular}}

\newcommand{\BashFancyFormatLine}{%
  \def\FancyVerbFormatLine##1{\$\,##1}%
}

\usepackage{pdflscape}
\usepackage{afterpage}

\title{Benchmarking Multi-Agent Deep Reinforcement Learning Algorithms in Cooperative Tasks}

\author{%
    Georgios Papoudakis \thanks{Equal Contribution}  \\
    School of Informatics \\
    University of Edinburgh \\
    \texttt{g.papoudakis@ed.ac.uk}
    \And
    Filippos Christianos $^*$ \\
    School of Informatics \\
    University of Edinburgh \\
    \texttt{f.christianos@ed.ac.uk}
    \AND
    Lukas Sch\"afer \\
    School of Informatics\\
    University of Edinburgh \\
    \texttt{l.schaefer@ed.ac.uk}
    \And
    Stefano V. Albrecht \\
    School of Informatics\\
    University of Edinburgh \\
    \texttt{s.albrecht@ed.ac.uk}
}
\begin{document}

\maketitle

\begin{abstract}
Multi-agent deep reinforcement learning (MARL) suffers from a lack of commonly-used evaluation tasks and criteria, making comparisons between approaches difficult. In this work, we provide a systematic evaluation and comparison of three different classes of MARL algorithms (independent learning, centralised multi-agent policy gradient, value decomposition) in a diverse range of cooperative multi-agent learning tasks. Our experiments serve as a reference for the expected performance of algorithms across different learning tasks, and we provide insights regarding the effectiveness of different learning approaches. We open-source EPyMARL, which extends the PyMARL codebase to include additional algorithms and allow for flexible configuration of algorithm implementation details such as parameter sharing. Finally, we open-source two environments for multi-agent research which focus on coordination under sparse rewards.
\end{abstract}

\section{Introduction}

Multi-agent reinforcement learning (MARL) algorithms use RL techniques to co-train a set of agents in a multi-agent system. Recent years have seen a plethora of new  MARL algorithms which integrate deep learning techniques \citep{papoudakis2019dealing,hernandez2019survey}.
However, comparison of MARL algorithms is difficult due to a lack of established benchmark tasks, evaluation protocols, and metrics. While several comparative studies exist for single-agent RL \citep{duan2016benchmarking,henderson2018deep,wang2019benchmarking}, we are unaware of such comparative studies for recent MARL algorithms. \citet{ar2012aamas} compare several MARL algorithms but focus on the application of classic (non-deep) approaches in simple matrix games. Such comparisons are crucial in order to understand the relative strengths and limitations of algorithms, which may guide practical considerations and future research.

We contribute a \textbf{comprehensive empirical comparison of nine MARL algorithms in a diverse set of cooperative multi-agent tasks}. We compare three classes of MARL algorithms: independent learning, which applies single-agent RL algorithms for each agent without consideration of the multi-agent structure \citep{tan1993multi}; centralised multi-agent policy gradient \citep{lowe2017multi,foerster2018counterfactual, yu2021surprising}; and value decomposition \citep{sunehag2017value,rashid2018qmix} algorithms. The two latter
classes of algorithms follow the Centralised Training Decentralised Execution (CTDE) paradigm.
These algorithm classes are frequently used in the literature either as baselines or building blocks for more complex algorithms \citep{he2016opponent,sukhbaatar2016learning,foerster2016learning,raileanu2018modeling,jaques2018social,iqbal2018actor,du2019liir,ryu2019multi}.
We evaluate algorithms in two matrix games and four multi-agent environments,
in which we define a total of 25 different cooperative learning tasks. Hyperparameters of each algorithm are optimised separately in each environment using a grid-search, and we report the maximum and average evaluation returns during training. We run experiments with shared and non-shared parameters between agents, a common implementation detail in MARL that has been shown to affect converged returns \citep{christianos2021scaling}. In addition to reporting detailed benchmark results, we analyse and discuss insights regarding the effectiveness of different learning approaches.

To facilitate our comparative evaluation, we created the open-source codebase \textbf{EPyMARL} (Extended PyMARL)\footnote{\url{https://github.com/uoe-agents/epymarl}}, an extension of PyMARL~\citep{samvelyan19smac} which is commonly used in MARL research. EPyMARL implements additional algorithms and allows for flexible configuration of different implementation details, such as whether or not agents share network parameters. Moreover, we have implemented and open-sourced \textbf{two new multi-agent environments}: Level-Based Foraging (LBF) and Multi-Robot Warehouse (RWARE). With these environments we aim to test the algorithms' ability to learn coordination tasks under sparse rewards and partial observability.

\section{Algorithms}

\subsection{Independent Learning (IL)}
For IL, each agent is learning independently and perceives the other agents as part of the environment.

\textbf{IQL:} In Independent Q-Learning (IQL) \citep{tan1993multi}, each agent has a decentralised state-action value function that is conditioned only on the local history of observations and actions of each agent. Each agent receives its local history of observations and updates the parameters of the Q-value network \citep{mnih2015human} by minimising the standard Q-learning loss \citep{watkins1992q}.

\textbf{IA2C:} Independent synchronous Advantage Actor-Critic (IA2C) is a variant of the commonly-used A2C algorithm \citep{mnih2016asynchronous, baselines} for decentralised training in multi-agent systems. Each agent has its own actor to approximate the policy and critic network to approximate the value-function. Both actor and critic are trained, conditioned on the history of local observations, actions and rewards the agent perceives, to minimise the A2C loss.

\textbf{IPPO:} Independent Proximal Policy Optimisation (IPPO) is a variant of the commonly-used PPO algorithm \citep{schulman2017proximal} for decentralised training in multi-agent systems. 
The architecture of IPPO is identical to IA2C. The main difference between PPO and A2C is that PPO uses a surrogate objective which constrains the relative change of the policy at each update, allowing for more update epochs using the same batch of trajectories. In contrast to PPO, A2C can only perform one update epoch per batch of trajectories to ensure that the training batch remains on-policy.

\subsection{Centralised Training Decentralised Execution (CTDE)}
 In contrast to IL, CTDE allows sharing of information during training, while policies are only conditioned on the agents' local observations enabling decentralised execution.
 
\paragraph{Centralised policy gradient methods}
One category of CTDE algorithms are centralised policy gradient methods in which each agent consists of a decentralised actor and a centralised critic, which is optimised based on shared information between the agents.

\textbf{MADDPG:} Multi-Agent DDPG (MADDPG) \citep{lowe2017multi} is a variation of the DDPG algorithm \citep{lillicrap2015continuous} for MARL.
The actor is conditioned on the history of local observations, while critic is trained on the joint observation and action to approximate the joint state-action value function.
Each agent individually minimises the deterministic policy gradient loss \citep{silver2014deterministic}.
A common assumption of DDPG (and thus MADDPG) is differentiability of actions with respect to the parameters of the actor, so the action space must be continuous. 
\citet{lowe2017multi} apply the Gumbel-Softmax trick \citep{jang2016categorical, maddison2016concrete} to learn in discrete action spaces.

\begin{table}[t]
\centering
\caption{\label{algs} Overview of algorithms and their properties.}
    \scalebox{1.0}{\begin{tabular}{@{}lccccc@{}}
        \toprule
                                & Centr. Training   & Off-/On-policy   & Value-based   & Policy-based   \\ \midrule
        IQL                     & \xmark & Off              & \cmark        & \xmark       \\
        IA2C                    & \xmark & On               & \cmark        & \cmark        \\
        IPPO                    & \xmark & On               & \cmark        & \cmark          \\
        MADDPG                  & \cmark & Off              & \cmark        & \cmark       \\
        COMA                    & \cmark & On               & \cmark        & \cmark        \\
        MAA2C                    & \cmark & On               & \cmark        & \cmark       \\
        MAPPO                   & \cmark & On               & \cmark        & \cmark        \\
        VDN                     & \cmark & Off              & \cmark        & \xmark         \\
        QMIX                    & \cmark & Off              & \cmark        & \xmark         \\ \bottomrule
    \end{tabular}}
\end{table}

\textbf{COMA:} In Counterfactual Multi-Agent (COMA) Policy Gradient, \citet{foerster2018counterfactual} propose a modification of the advantage in the actor's loss computation to perform counterfactual reasoning for credit assignment in cooperative MARL. 
The advantage is defined as the discrepancy between the state-action value of the followed joint action and a counterfactual baseline. The latter is given by the expected value of each agent following its current policy while the actions of other agents are fixed. %
The standard policy loss with this modified advantage is used to train the actor and the critic is trained using the TD-lambda algorithm \citep{sutton1988learning}.

\textbf{MAA2C:} Multi-Agent A2C (MAA2C) is an actor-critic algorithm in which 
the critic learns a joint state value function (in contrast, the critics in MADDPG and COMA are also conditioned on actions). It extends the existing on-policy actor-critic algorithm A2C by applying centralised critics conditioned on the state of the environment rather than the individual history of observations. It is often used as a baseline in MARL research and is sometimes referred to as Central-V, because it computes a centralised state value function. However, MAPPO also computes a centralised state value function, and in order to avoid confusion we refer to this algorithm as MAA2C.

\textbf{MAPPO:} Multi-Agent PPO (MAPPO) \citep{yu2021surprising} is an actor-critic algorithm (extension of IPPO) in which  the critic learns a joint state value function, similarly to MAA2C. In contrast to MAA2C, which can only perform one update epoch per training batch, MAPPO can utilise the same training batch of trajectories to perform several update epochs.

\paragraph{Value Decomposition}
Another recent CTDE research direction is the decomposition of the joint state-action value function into individual state-action value functions. 

\textbf{VDN:} Value Decomposition Networks (VDN) \citep{sunehag2017value} aim to learn a linear decomposition of the joint Q-value. Each agent maintains a network to approximate its own state-action values. VDN decomposes the joint Q-value into the sum of individual Q-values.
The joint state-action value function is trained using the standard DQN algorithm \citep{watkins1992q, mnih2015human}. During training, gradients of the joint TD loss flow backwards to the network of each agent. 

\textbf{QMIX:} QMIX \citep{rashid2018qmix} extends VDN to address a broader class of environments. To represent a more complex decomposition, a parameterised mixing network is introduced to compute the joint Q-value based on each agent's individual state-action value function.
A requirement of the mixing function is that the optimal joint action, which maximises the joint Q-value, is the same as the combination of the individual actions maximising the Q-values of each agent.
QMIX is trained to minimise the DQN loss and the gradient is backpropagated to the individual Q-values.

\section{Multi-Agent Environments}
\label{sec:envs}

We evaluate the algorithms in two finitely repeated matrix games and four multi-agent environments within which we define a total of 25 different learning tasks. All tasks are fully-cooperative, i.e.\ all agents receive identical reward signals.
These tasks range over various properties including the degree of observability (whether agents can see the full environment state or only parts of it), reward density (receiving frequent/dense vs infrequent/sparse non-zero rewards), and the number of agents involved. \Cref{envs} lists environments with properties, and we give brief descriptions below. 
We believe each of the following environments addresses a specific challenge of MARL. 
Full details of environments and learning tasks are provided in
\Cref{appendix:tasks}.\\

\subsection{Repeated Matrix Games}

We consider two cooperative matrix games proposed by \citet{claus1998dynamics}: the \emph{climbing} and \emph{penalty} game. The common-payoff matrices of the climbing and penalty game, respectively, are:
$$
\begin{bmatrix}
    0   & 6   & 5\\   
    -30 & 7   & 0 \\
    11  & -30 & 0
\end{bmatrix}
\hspace{2cm}
\begin{bmatrix}
    k   & 0   & 10\\
    0   & 2   & 0 \\
    10  & 0   & k
\end{bmatrix}
$$
where $k \leq 0$ is a penalty term.
We evaluate in the penalty game for $k \in \{-100, -75, -50, -25, 0\}$. The difficulty of this game strongly correlates with $k$: the smaller $k$, the harder it becomes to identify the optimal policy due to the growing risk of penalty $k$.  Both games are applied as repeated matrix games with an episode length of 25 and agents are given constant observations at each timestep. These matrix games are challenging due to the existence of local minima in the form of sub-optimal Nash equilibria \citep{nash1951non}. Slight deviations from optimal policies by one of the agents can result in significant penalties, so agents might get stuck in risk-free (deviations from any agent does not significantly impede payoff) local optima. 

\begin{table}[t]
    \centering
    \caption{\label{envs} Overview of environments and properties.}
    \scalebox{1.0}{\begin{tabular}{@{}llllll@{}}
        \toprule
                                         & Observability & Rew. Sparsity   & Agents & Main Difficulty \\ \midrule
        Matrix Games                     & Full           & Dense           & 2     &  Sub-optimal equilibria \\
        MPE                              & Partial / Full & Dense           & 2-3   &  Non-stationarity\\
        SMAC                             & Partial       & Dense           & 2-10    &  Large action space \\
        LBF                              & Partial / Full          & Sparse~\tablefootnote{Rewards in LBF are sparser compared to MPE and SMAC, but not as sparse as in RWARE.}          & 2-4 & Coordination    \\
        RWARE                            & Partial       &  Sparse       & 2-4    & Sparse reward\\ \bottomrule
    \end{tabular}}
    
\end{table}

\subsection{Multi-Agent Particle Environment}

The Multi-Agent Particle Environments (MPE)~\citep{mordatch2017emergence} consists of several two-dimensional navigation tasks. We investigate four tasks that emphasise coordination: Speaker-Listener, Spread, Adversary\footnote{\label{mpe_pretraining}Adversary and Predator-Prey are originally competitive tasks. The agents controlling the adversary and prey, respectively, are controlled by a pretrained policy obtained by training all agents with the MADDPG algorithm for 25000 episodes (see \Cref{appendix:tasks} for details).}, and Predator-Prey\cref{mpe_pretraining}. Agent observations consist of high-level feature vectors including relative agent and landmark locations. The actions allow for two-dimensional navigation. All tasks but Speaker-Listener, which also requires binary communication, are fully observable. MPE tasks serve as a benchmark for agent coordination and their ability to deal with non-stationarity~\citep{papoudakis2019dealing} due to significant dependency of the reward  with respect to joint actions. Individual agents not coordinating effectively can severely reduce received rewards.

\subsection{StarCraft Multi-Agent Challenge}

The StarCraft Multi-Agent Challenge (SMAC)~\citep{samvelyan19smac} simulates battle scenarios in which a team of controlled agents must destroy an enemy team using fixed policies. Agents observe other units within a fixed radius, and can move around and select enemies to attack. %
We consider five tasks in this environment which vary in the number and types of units controlled by agents. The primary challenge within these tasks is the agents' ability to accurately estimate the value of the current state under partial observability and a growing number of agents of diverse types across tasks. Latter leads to large action spaces for agents which are able to select other agents or enemy units as targets for healing or attack actions, respectively, depending on the controlled unit.

\subsection{Level-Based Foraging}
\label{sec:lbf}
In Level-Based Foraging (LBF)  \citep{albrecht2013game, as2017aamas} agents must collect food items which are scattered randomly in a grid-world. Agents and items are assigned levels, such that a group of one or more agents can collect an item if the sum of their levels is greater or equal to the item's level. Agents can move in four directions, and have an action that attempts to load an adjacent item (the action will succeed depending on the levels of agents attempting to load the particular item). 
LBF allows for many different tasks to be configured, including partial observability or a highly cooperative task where all agents must simultaneously participate to collect the items. We define seven distinct tasks with a variable world size, number of agents, observability, and cooperation settings indicating whether all agents are required to load a food item or not. We implemented the LBF environment which is publicly available on GitHub, under the MIT licence: \url{https://github.com/uoe-agents/lb-foraging}.

\subsection{Multi-Robot Warehouse}
\label{sec:rware}
The Multi-Robot Warehouse environment (RWARE) represents a cooperative, partially-observable environment with sparse rewards. RWARE simulates a grid-world warehouse in which agents (robots) must locate and deliver requested shelves to workstations and return them after delivery. Agents are only rewarded for completely delivering requested shelves and observe a $3\times3$ grid containing information about the surrounding agents and shelves. The agents can move forward, rotate in either direction, and load/unload a shelf. We define three tasks which vary in world size, number of agents and shelf requests. The sparsity of the rewards makes this a hard environment, since agents must correctly complete a series of actions before receiving any reward. Additionally, observations are sparse and  high-dimensional compared to the other environments. RWARE is the second environment we designed and open-source under the MIT licence: \url{https://github.com/uoe-agents/robotic-warehouse}.

We have developed and plan to maintain the LBF and RWARE environments as part of this work. They have already been used in other multi-agent research ~\citep{christianos2020shared,christianos2021scaling,rahman2021open,papoudakis2021local}. For more information including installation instructions, interface guides with code snippets and detailed descriptions, see \Cref{app:new_envs}.

\begin{figure}
    \centering
    \begin{subfigure}{.49\textwidth}
         \centering
         \includegraphics[width=.55\textwidth]{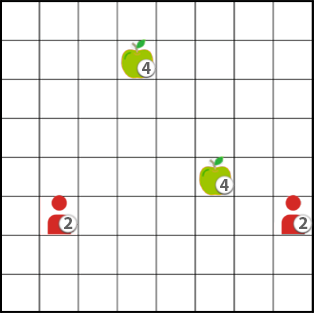}
         \caption{Level-Based Foraging (LBF)}
    \end{subfigure}
    \hfill
    \begin{subfigure}{.49\textwidth}
         \centering
         \includegraphics[width=.55\textwidth]{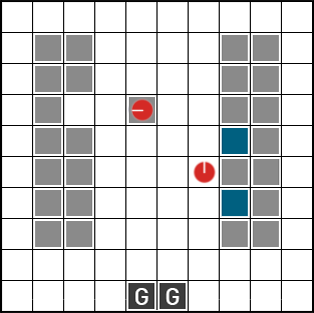}
         \caption{Multi-Robot Warehouse (RWARE)}
    \end{subfigure}
    \caption{Illustrations of the open-sourced multi-agent environments \citep{christianos2020shared}.}
    \label{fig:new_environments}
\end{figure}

\section{Evaluation}
\label{sec:evaluation}
\subsection{Evaluation Protocol}
To account for the improved sample efficiency of off-policy over on-policy algorithms and to allow for fair comparisons, we train on-policy algorithms for a factor of ten more samples than off-policy algorithms. In MPE and LBF we train on-policy algorithms for 20 million timesteps and off-policy algorithms for two million timesteps, while in SMAC and RWARE, we train on-policy and off-policy algorithms for 40 and four million timesteps, respectively.
By not reusing samples through an experience replay buffer, on-policy algorithms are less sample efficient, but not generally slower (if the simulator is reasonably fast) and thus this empirical adjustment is fair. 
We perform in total 41 evaluations of each algorithm at constant timestep intervals during training, and at each evaluation point we evaluate for 100 episodes.
In matrix games, we train off-policy algorithms for 250 thousand timesteps and on-policy algorithms for 2.5 million timesteps and evaluate every 2.5 thousand and 25 thousand timesteps, respectively, for a total of 100 evaluations.

\subsection{Parameter Sharing}
Two common configurations for training deep MARL algorithms are: without and with parameter sharing. Without parameter sharing, each agent uses its own set of parameters for its networks. Under parameter sharing, all agents share the same set of parameters for their networks.
In the case of parameter sharing, the policy and critic (if there is one) additionally receive the identity of each agent as a one-hot vector input. This allows for each agent to develop a different behaviour. The loss is computed over all agents and used to optimise the shared parameters.
In the case of varying input sizes across agents, inputs are zero-padded to ensure identical input dimensionality.
Similarly if agents have varying numbers of actions, action selection probabilities for invalid actions are set to 0.

\subsection{Hyperparameter Optimisation}

Hyperparameter optimisation was performed for each algorithm separately in each environment. From each environment, we selected one task and optimised the hyperparameters of all algorithms in this task.
In the MPE environment, we perform the hyperparameter optimisation in the Speaker-Listener task, in the SMAC environment in the ``3s5z'' task, in the LBF environment in the ``15x15-3p-5f'' task, and in the RWARE  environment in the ``Tiny 4p'' task. We train each combination of hyperparameters using three different seeds and compare the maximum evaluation returns. The best performing combination on each task is used for all tasks in the respective environment for the final experiments. In \Cref{sec:hyper}, we present the hyperparameters that were used in each environment and algorithm. 

\subsection{Performance Metrics}
\label{sec:metrics}

\textbf{Maximum returns:} For each algorithm, we identify the evaluation timestep during training in which the algorithm achieves the highest average evaluation returns across five random seeds. 
We report the average returns and the 95\% confidence interval across five seeds from this evaluation timestep. 
 
\textbf{Average returns:} We also report the average returns achieved throughout all evaluations during training. Due to this metric being computed over all evaluations executed during training, it considers learning speed besides final achieved returns.

\subsection{Computational Requirements}
\label{sec:computational_cost}
All experiments presented in this work were executed purely on CPUs. The experiments were executed in compute clusters that consist of several nodes. The main types of CPU models that were used for this work are Intel(R) Xeon(R) CPU E5-2630 v3 @ 2.40GHz and AMD EPYC 7502 32-Core processors. All but the SMAC experiments were executed using a single CPU core. All SMAC experiments were executed using 5 CPU cores. The total number of CPU hours that were spent for executing the experiments in this work (excluding the hyperparameter search) are 138,916.

\subsection{Extended PyMARL}

Implementation details in reinforcement learning significantly affect the returns that each algorithm achieves \citep{andrychowicz2020matters}. To enable consistent evaluation of MARL algorithms, we open-source the Extended PyMARL (EPyMARL) codebase. EPyMARL is an extension of the PyMARL codebase \citep{samvelyan19smac}. PyMARL provides implementations for IQL, COMA, VDN and QMIX. We increase the scope of the codebase to include five additional policy gradients algorithms: IA2C, IPPO, MADDPG, MAA2C and MAPPO. The original PyMARL codebase implementation assumes that agents share parameters and that all the agents' observation have the same shape. In general, parameter sharing is a commonly applied technique in MARL. However, it was shown that parameter sharing can act as an information bottleneck, especially in environments with heterogeneous agents \citep{christianos2021scaling}. EPyMARL allows training MARL algorithms without parameter sharing, training agents with observations of varying dimensionality, and tuning several implementation details such as reward standardisation, entropy regularisation, and the use of recurrent or fully-connected networks.
EPyMARL is publicly available on GitHub and distributed under the Apache License: \url{https://github.com/uoe-agents/epymarl}.

\section{Results}\label{sec:results}

\begin{figure}[t]
    \centering
    \begin{subfigure}{0.25\textwidth}
        \includegraphics[width=1\linewidth]{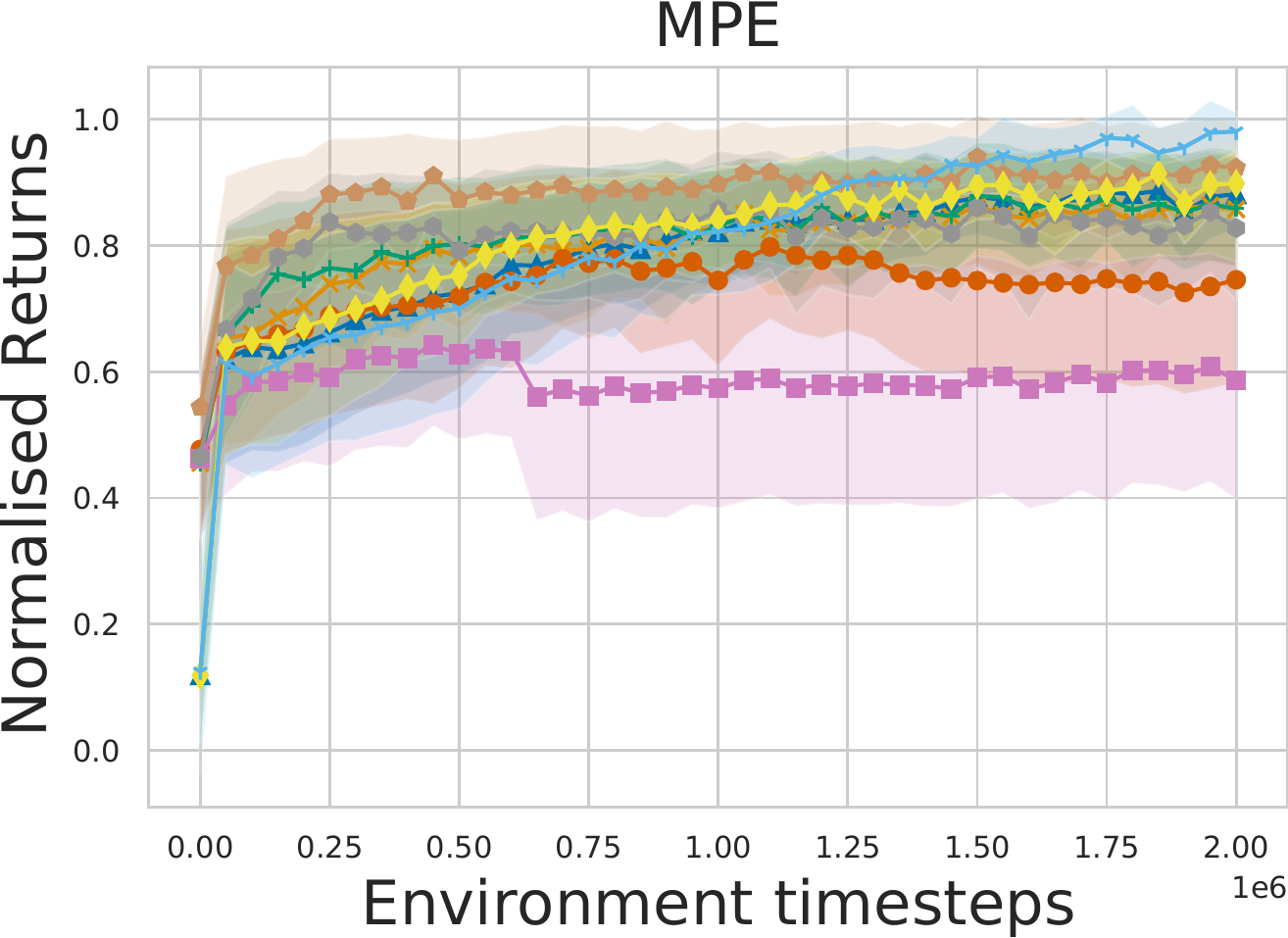}
    \end{subfigure}
     \hfill
    \begin{subfigure}{0.24\textwidth}
        \includegraphics[width=1\linewidth]{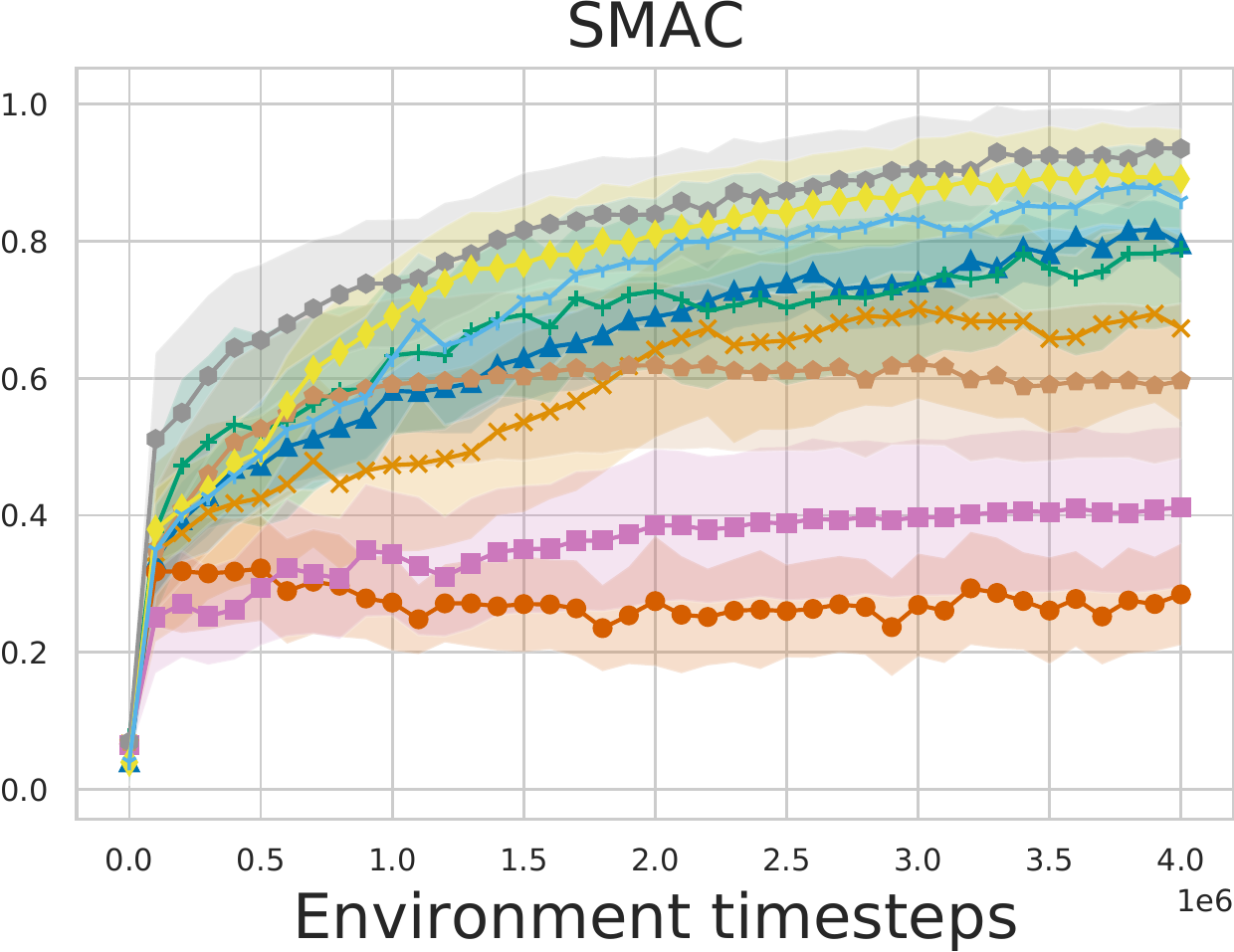}
    \end{subfigure}
     \hfill
    \begin{subfigure}{0.24\textwidth}
        \includegraphics[width=1\linewidth]{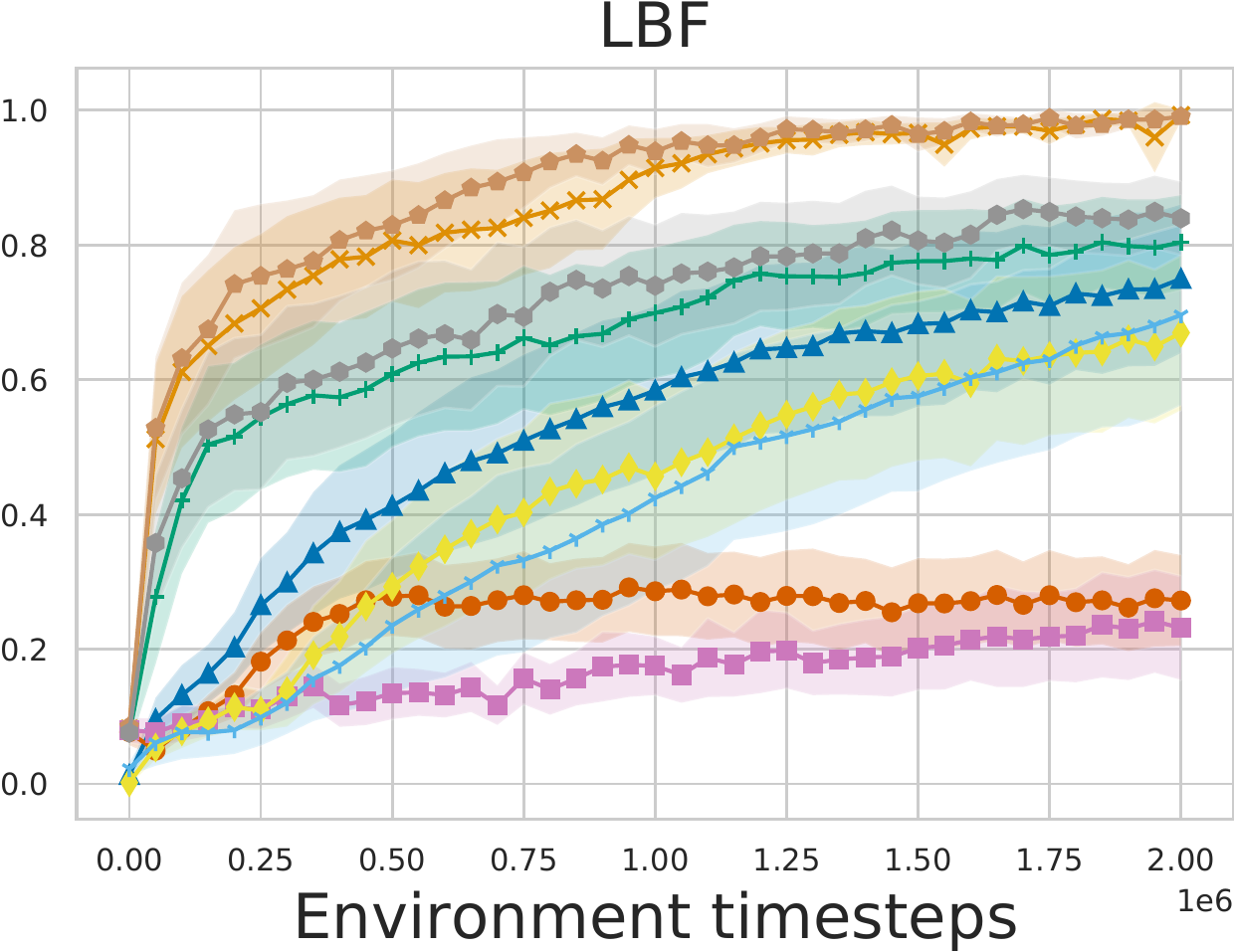}
    \end{subfigure}
     \hfill
    \begin{subfigure}{0.24\textwidth}
          \includegraphics[width=1\linewidth]{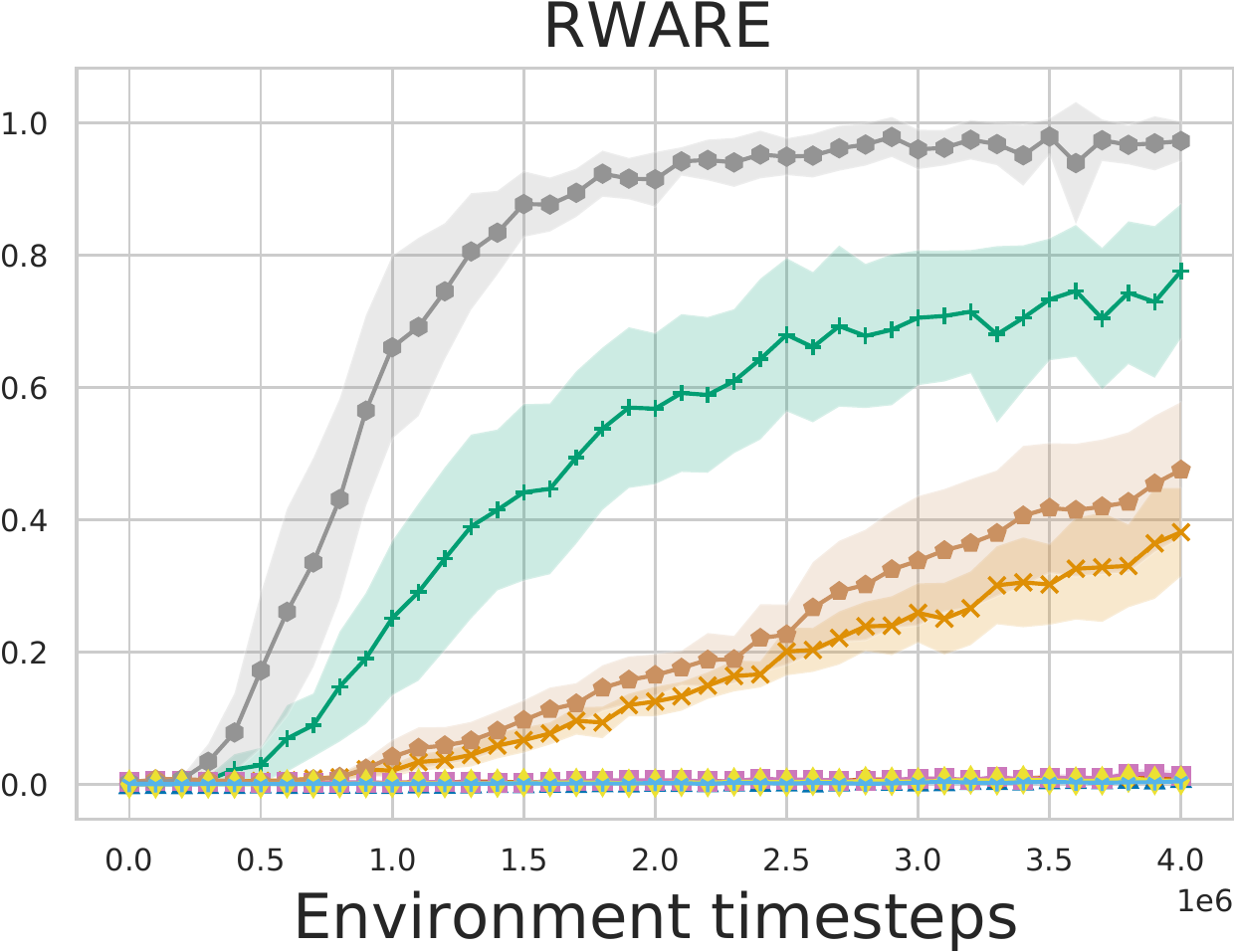}
    \end{subfigure}
    \begin{subfigure}{1\textwidth}
        \centering
        \includegraphics[trim={0cm 0cm 0.0cm, 10.1cm}, clip, width=0.5\textwidth]{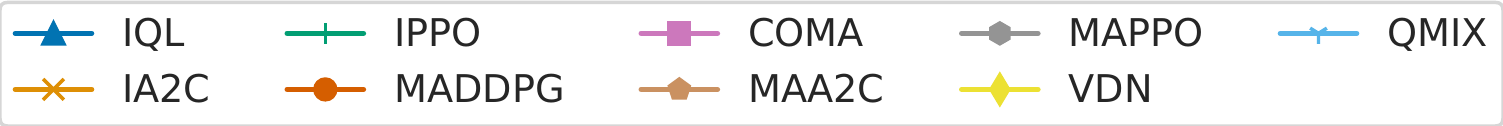}
    \end{subfigure}

\caption{\label{fig:norm_returns} Normalised evaluation returns averaged over the tasks in the all environments except matrix games. Shadowed part represents the 95\% confidence interval.}
\end{figure}

In this section we compile the results across all environments and algorithms. \Cref{fig:norm_returns} presents the normalised evaluation returns in all environments, except matrix games. We normalise the returns of all algorithms in each task in the $[0,1]$ range using the following formula:
$    \mathrm{norm}\_G_t^a = (G_t^a - \min(G_t))/(\max(G_t) - \min(G_t))
$, where $G_t^a$ is the return of algorithm $a$ in task $t$, and $G_t$ is the returns of all algorithms in task $t$.
\Cref{tab:max_returns} presents the maximum returns for the nine algorithms in all 25 tasks with parameter sharing. The maximum returns without parameter sharing, as well as the average returns  both with and without parameter sharing are presented in \Cref{sec:add_results}. 
In \Cref{tab:max_returns}, we highlight the highest mean in bold. We performed two-sided t-tests with a significance threshold of $0.05$ between the highest performing algorithm and each other algorithm in each task. If an algorithm's performance was \emph{not} statistically significantly different from the best algorithm, the respective value is annotated with an asterisk (i.e. bold or asterisks in the table show the best performing algorithms per task). In the SMAC tasks, it is a common practice in the literature to report the win-rate as a percentage and not the achieved returns. However, we found it more informative to report the achieved returns since it is the metric that the algorithms aim to optimise. Moreover, higher returns do not always correspond to higher win-rates which can make the interpretation of the performance metrics more difficult. For completeness, we report the win-rates achieved by all algorithms in \Cref{sec:smac_win}.

\begin{table}[t]
\centering
\caption{\label{tab:max_returns}
Maximum returns and 95\% confidence interval over five seeds for all nine algorithms with parameter sharing in all 25 tasks. The highest  value in each task is presented in bold. Asterisks denote the algorithms that are not significantly different from the best performing algorithm in each task.}
\resizebox{\linewidth}{!}{
\robustify\bf
\large
\begin{tabular}{p{1em} l S S  S  S  S  S  S S S S}
\toprule
& \textbf{Tasks \textbackslash Algs. }  &             {IQL} &           {IA2C} &             {IPPO} &              {MADDPG} &              {COMA} &              {MAA2C} &               {MAPPO} &             {VDN} &           {QMIX} \\ \\
\hline \\
\multirow{6}{*}{\rotatebox[origin=c]{90}{Matrix Games}} & Climbing    &  \bf 195.00(6782) &  175.00(0) &  175.00(0) &  170.00(1000) &  185.00(4899) &  175.00(0) &  175.00(0) &  175.00(5477) &  175.00(5477) \\
& Penalty k=0   &     \bf 250.00(0) &  \bf 250.00(0) &  \bf 250.00(0) &     249.98(4) &     \bf 250.00(0) &  \bf 250.00(0) &  \bf 250.00(0) &     \bf 250.00(0) &     \bf 250.00(0) \\
& Penalty k=-25  &      \bf 50.00(0) &   \bf 50.00(0) &   \bf 50.00(0) &      49.97(2) &     \bf  50.00(0) &   \bf 50.00(0) &   \bf 50.00(0) &      \bf 50.00(0) &      \bf 50.00(0) \\
& Penalty k=-50  &      \bf 50.00(0) &   \bf 50.00(0) &   \bf 50.00(0) &      49.98(2) &      \bf 50.00(0) &   \bf 50.00(0) &   \bf 50.00(0) &      \bf 50.00(0) &      \bf 50.00(0) \\
& Penalty k=-75  &      \bf 50.00(0) &   \bf 50.00(0) &   \bf 50.00(0) &      49.97(2) &      \bf 50.00(0) &   \bf 50.00(0) &   \bf 50.00(0) &      \bf 50.00(0) &      \bf 50.00(0) \\
& Penalty k=-100&      \bf 50.00(0) &   \bf 50.00(0) &   \bf 50.00(0) &      49.97(3) &      \bf 50.00(0) &   \bf 50.00(0) &   \bf 50.00(0) &      \bf 50.00(0) &      \bf 50.00(0) \\ \\
\hline \\
\multirow{4}{*}{\rotatebox[origin=c]{90}{MPE}} & Speaker-Listener&   -18.36(467) &   -12.60(362)* &   -13.10(350) &   -13.56(173) &   -30.40(518) &    -10.71(38)* &    \bf -10.68(30) &   -15.95(248) &    -11.56(53) \\
& Spread         &  -132.63(222) &  -134.43(115) &  -133.86(367) &  -141.70(174) &  -204.31(630) &  -129.90(163)* &  -133.54(308) &  -131.03(185) &  \bf -126.62(296) \\
& Adversary       &      9.38(91) &     12.12(44)* &     \bf 12.17(32) &      8.97(89) &      8.05(89) &     12.06(45)* &     11.30(38) &      9.28(90) &      9.67(66) \\
& Tag           &    22.18(283) &    17.44(131) &    19.44(294) &    12.50(630) &     8.72(442) &    19.95(715)* &    18.52(564) &    24.50(219) &    \bf 31.18(381) \\ \\
\hline \\
\multirow{5}{*}{\rotatebox[origin=c]{90}{SMAC}} 
& 2s\_vs\_1sc &     16.72(38) &      20.24(0) &     20.24(1) &  13.14(201) &   11.04(721) &   20.20(5)* &    \bf 20.25(0) &     18.04(33) &    19.01(40) \\
& 3s5z      &     16.44(15) &   18.56(131)* &   13.36(208) &   12.04(82) &  18.90(101)* &   19.95(5)* &  \bf 20.39(114) &    19.57(20)* &   19.66(14)* \\
& corridor  &    15.72(177) &  \bf 18.59(62) &  17.97(344)* &    5.85(58) &     7.75(19) &    8.97(29) &    17.14(439)* &   15.25(418)* &  16.45(354)* \\
& MMM2      &    13.69(102) &    10.70(277) &   11.37(115) &    3.96(32) &     6.95(27) &  10.37(195) &      17.78(44) &  \bf 18.49(31) &   18.40(24)* \\
& 3s\_vs\_5z  &  \bf 21.15(41) &       4.42(2) &  19.36(615)* &    5.99(58) &      3.23(5) &    6.68(55) &    18.17(417)* &   19.03(577)* &   16.04(287) \\ \\
\hline \\
\multirow{7}{*}{\rotatebox[origin=c]{90}{LBF}} 
& 8x8-2p-2f-c    &   \bf 1.00(0) &  \bf 1.00(0) &   \bf 1.00(0) &   0.46(2) &  0.61(30) &  \bf 1.00(0) &  \bf 1.00(0) &  \bf 1.00(0) &  0.96(7)* \\
& 8x8-2p-2f-2s-c &   \bf 1.00(0) &  \bf 1.00(0) &   0.78(5) &   0.70(4) &  0.45(15) &  \bf 1.00(0) &   0.85(6) &  \bf 1.00(0) &  \bf 1.00(0) \\
& 10x10-3p-3f       &   0.93(2) &  \bf 1.00(0) &   0.98(1) &  0.24(4) &   0.19(6) &  \bf 1.00(0) &   0.99(1) &  0.84(8) & 0.84(8) \\ 
& 10x10-3p-3f-2s    &   0.86(1) &  0.94(3)* &   0.70(3) &   0.41(3) &  0.29(12) &  \bf 0.96(2) &   0.72(3) &  0.90(3) &  0.90(1) \\
& 15x15-3p-5f       &  0.17(8) &  \bf 0.89(4) &   0.77(8) &   0.10(2) &   0.08(4) &  0.87(6)* &   0.77(2) &  0.15(2) &  0.09(4) \\
&  15x15-4p-3f      &   0.54(18) &  0.99(1)* &   0.98(1) &   0.17(3) &   0.17(4) &  \bf 1.00(0) &   0.96(2) &  0.38(13) &  0.15(6) \\
& 15x15-4p-5f       &  0.22(4) &  0.93(3)* &  0.67(22) &   0.12(6) &   0.12(6) &  \bf 0.95(1) &  0.70(25)* &  0.30(4) &  0.25(9) \\
\\
\hline \\
\multirow{3}{*}{\rotatebox[origin=c]{90}{RWARE}} 
& Tiny 4p &  0.72(37) &  26.34(460) &  31.82(1071) &  0.54(10) &  1.16(15) &  32.50(979) &  \bf49.42(122) &  0.80(28) &  0.30(19) \\
& Small 4p &  0.14(28) &   6.54(115) &   19.78(312) &  0.18(12) &  0.16(16) &  10.30(148) &  \bf27.00(180) &  0.18(27) &   0.06(8) \\
& Tiny 2p  &  0.28(38) &   8.18(125) &  20.22(176)* &  0.44(34) &  0.48(34) &   8.38(259) &  \bf21.16(150) &   0.12(7) &  0.14(19) \\ \\
\bottomrule
\end{tabular}
}

\end{table}

\subsection{Independent Learning}

We find that IL algorithms perform adequately in all tasks despite their simplicity. However, performance of IL is limited in partially observable SMAC and RWARE tasks, compared to their CTDE counterparts, due to IL algorithms' inability to reason over joint information of agents. 

\textbf{IQL:} 
IQL performs significantly worse than the other IL algorithms in the partially-observable Speaker-Listener task and in all RWARE tasks. IQL is particularly effective in all but three LBF tasks, where relatively larger grid-worlds are used. IQL achieves the best performance among all algorithms in the ``3s\_vs\_5z'' task, while it performs competitively in the rest of the SMAC tasks.

\textbf{IA2C:}
The stochastic policy of IA2C appears to be particularly effective on  all environments except in a few SMAC tasks.  In the majority of tasks, it performs similarly to IPPO with the exception of RWARE and some SMAC tasks. However, it achieves higher returns than IQL in all but two SMAC tasks. Despite its simplicity, IA2C performs competitively compared to all CTDE algorithms, and significantly outperforms COMA and MADDPG in the majority of the tasks.

\textbf{IPPO:}
IPPO in general performs competitively in all tasks across the different environments. On average (\Cref{fig:norm_returns}) it achieves higher returns than IA2C in MPE, SMAC and RWARE tasks, but lower returns in the LBF tasks.
IPPO also outperforms MAA2C in the partially-observable RWARE tasks, but in general it performs worse compared to its centralised MAPPO version.

\subsection{Centralised Training Decentralised Execution}
Centralised training aims to learn powerful critics over joint observations and actions to enable reasoning over a larger information space. We find that learning such critics is valuable in tasks which require significant coordination under partial observability, such as the MPE Speaker-Listener and harder SMAC tasks. In contrast, IL is competitive compared to CTDE algorithms in fully-observable tasks of MPE and LBF. 
Our results also indicate that in most RWARE tasks, MAA2C and MAPPO significantly improve the achieved returns compared to their IL (IA2C and IPPO) versions. %
However, training state-action value functions appears challenging in RWARE tasks with sparse rewards, leading to very low performance of the remaining CTDE algorithms (COMA, VDN and QMIX).

\paragraph{Centralised Multi-Agent Policy Gradient} Centralised policy gradient methods vary significantly in performance.

\textbf{MADDPG:} MADDPG performs worse than all the other algorithms except COMA, in the majority of the tasks. It only performs competitively in some MPE tasks.
It also exhibits very low returns in discrete grid-world environments LBF and RWARE. We believe that these results are a direct consequence of the biased categorical reparametarisation using Gumbel-Softmax.

\textbf{COMA:}
In general, COMA exhibits one of the lowest performances in most tasks and only performs competitively in one SMAC task. We found that COMA suffers very high variance in the computation of the counterfactual advantage. In the Speaker-Listener task, it fails to find the sub-optimal local minima solution that correspond to returns around to -17. Additionally, it does not exhibit any learning in the RWARE tasks in contrast to other on-policy algorithms.

\textbf{MAA2C:}
MAA2C in general performs competitively in the majority of the tasks, except a couple of SMAC tasks. Compared to MAPPO, MAA2C achieves slightly higher returns in the MPE and the LBF tasks, but in most cases significantly lower returns in the SMAC and RWARE tasks.

\textbf{MAPPO:} MAPPO achieves high returns in the vast majority of tasks and only performs slightly worse than other algorithms in some MPE and LBF tasks. Its main advantage is the combination of on-policy optimisation with its surrogate objective which significantly improves the sample efficiency compared to MAA2C. Its benefits can be observed in RWARE tasks where its achieved returns exceed the returns of all other algorithms (but not always significantly).

\paragraph{Value Decomposition}
Value decomposition is an effective approach in most environments. In the majority of tasks across all environments except RWARE, VDN and QMIX outperform or at least match the highest returns of any other algorithm. This suggests that VDN and QMIX share the major advantages of centralised training. 
In RWARE, VDN and QMIX do not exhibit any learning, similar to IQL, COMA and MADDPG, indicating that value decomposition methods require sufficiently dense rewards to successfully learn to decompose the value function into the individual agents.

\textbf{VDN:}
While VDN and QMIX perform similarly in most environments, the difference in performance is most noticeable in some MPE tasks. It appears VDN's assumption of linear value function decomposition is mostly violated in this environment. In contrast, VDN and QMIX perform similarly in most SMAC tasks and across all LBF tasks, where the global utility can apparently be represented by a linear function of individual agents' utilities.

\textbf{QMIX:}
Across almost all tasks, QMIX achieves consistently high returns, but does not necessarily achieve the highest returns among all algorithms. Its value function decomposition allows QMIX to achieve slightly higher returns in some of the more complicated tasks where the linear value decomposition of VDN in is not sufficient.

\subsection{Parameter Sharing}

\begin{wrapfigure}{r}{0.35\textwidth}
    \centering %
    \includegraphics[width=0.39\textwidth]{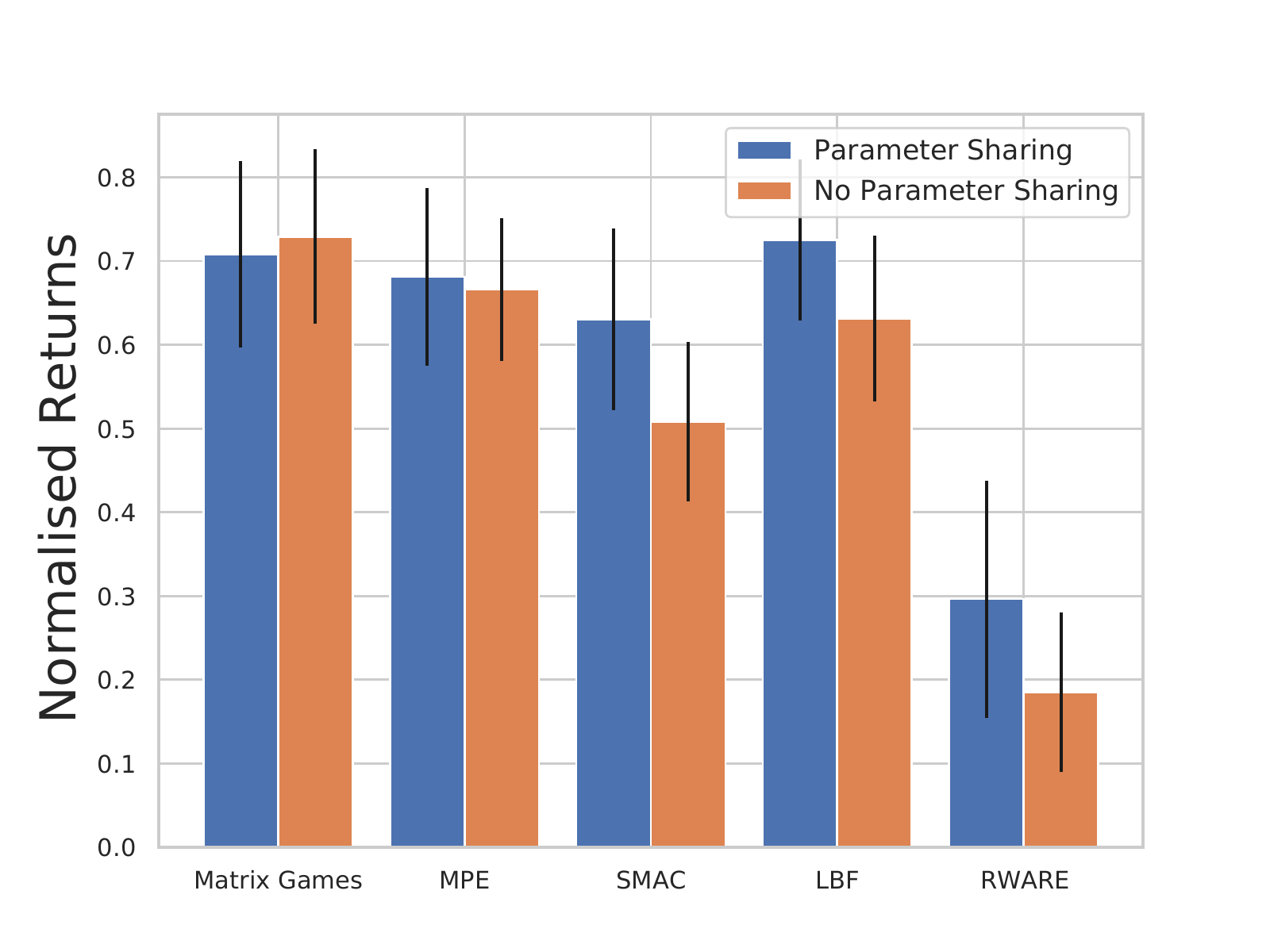}
    \caption{Normalised maximum returns averaged over all algorithms with/without parameter sharing (with standard error).}
    \label{fig:s_ns_comp}
\end{wrapfigure}%

\Cref{fig:s_ns_comp} presents the normalised maximum returns averaged over the nine algorithms and tasks  with and without parameter sharing in all environments. We observe that in all environments except the matrix games, parameter sharing improves the returns over no parameter sharing. 
While the average values presented in \Cref{fig:s_ns_comp} do not seem statistically significant, by looking closer in Tables \ref{tab:max_returns} and \ref{tab:max_returns_ns} we observe that in several cases of algorithm-task pairs the improvement due to parameter sharing seems significant.
Such improvements can be observed for most algorithms in MPE tasks, especially in Speaker-Listener and Tag. For SMAC, we observe that parameter sharing improves the returns in harder tasks. Similar observations can be made for LBF and RWARE. In these environments, the return improvement of parameter sharing appears to correlate with the sparsity of rewards. For tasks with larger grid-worlds or fewer agents, where the reward is more sparse, parameter sharing leads to large increases in returns compared to simpler tasks.
This does not come as a surprise since parameter sharing uses a larger number of trajectories to train the same shared architecture to improve sample efficiency compared to no parameter sharing.

\section{Analysis}\label{sec:discussion}

\textbf{Independent learning can be effective in multi-agent systems. Why and when?} It is often stated that IL is inferior to centralised training methods due to the environment becoming non-stationary from the perspective of any individual agent. This is true in many cases and particularly crucial when IL is paired with off-policy training from an experience replay buffer, as pointed out by \citet{lowe2017multi}. In our experiments, IQL trains agents independently using such a replay buffer and is thereby limited in its performance in tasks that require extensive coordination among the agents. There, agents depend on the information about other agents and their current behaviour to choose well-coordinated actions and hence learning such policies from a replay buffer (where other agents differ in their behaviour) appears infeasible. However, this is not a concern to multi-agent environments in general. In smaller SMAC tasks and most LBF tasks, each agent can independently learn a policy that achieves relatively high returns
by utilising only %
its local observation history, and without requiring extensive coordination with other agents. E.g.\ in LBF, agents "only" have to learn to pick up the food until it is collected. %
Of course, they will have to coordinate such behaviour with other agents, but naively going to food (especially when others are also close) and attempting to pick it up can be a viable local minima policy, and hard to improve upon.
Whenever more complicated coordination is required, such as simultaneously picking up an item with higher level, exploring and learning those joint actions becomes difficult. 
IA2C and IPPO on the other hand learn on-policy, so there is no such training from a replay buffer. These algorithms should be expected to achieve higher returns in the majority of the tasks as they learn given the currently demonstrated behaviour of other agents. As long as effective behaviour is eventually identified through time and exploration, IA2C and IPPO can learn more effective policies than IQL despite also learning independently.

\textbf{Centralised information is required for extensive coordination under partial-observability.} 
We note that the availability of joint information (observations and actions) over all agents serves as a powerful training signal to optimise individual policies whenever the full state is not available to individual agents.
Comparing the performance in \Cref{tab:max_returns} of IA2C and MAA2C, two almost identical algorithms aside from their critics, we notice that MAA2C achieves equal or higher returns in the majority of the tasks. This difference is particularly significant in tasks where agent observations lack important information about other agents or parts of the environment outside of their receptive field due to partial observability. This can be observed in RWARE tasks with 4 agents which require extensive coordination so that agents are not stuck in the narrow passages. However, in RWARE Tiny 2p task, the performance of IA2C and MAA2C is similar as only two agents rarely get stuck in the narrow passages.
Finally, IA2C and MAA2C have access to the same information in fully-observable tasks, such as most LBF and MPE tasks, leading to similar returns. A similar pattern can be observed for IPPO and MAPPO. %
However, we also observe that centralised training algorithms such as COMA, MADDPG, VDN and QMIX are unable to learn effective behaviour in the partially-observable RWARE. We hypothesise that training larger networks over the joint observation- and action-space, as required for these algorithms, demands sufficient training signals. However, rewards are sparse in RWARE and observations are comparably large.

\textbf{Value decomposition -- VDN vs QMIX.}
Lastly, we address the differences observed in value decomposition applied by VDN and QMIX. Such decomposition offers an improvement in comparison to the otherwise similar IQL algorithm across most tasks. %
Both VDN and QMIX are different in their decomposition. QMIX uses a trainable mixing network to compute the joint Q-value compared to VDN which assumes linear decomposition.
With the additional flexibility, QMIX aims to improve learnability, i.e.\ it simplifies the learning objective for each agent to maximise, while ensuring the global objective is maximised by all agents \citep{agogino2008analyzing}.  Such flexibility appears to mostly benefit convergence in harder MPE tasks, such as Speaker-Listener and Tag, but comes at additional expense seen in environments like LBF, where the decomposition did not have to be complex.
It appears that the dependency of rewards with respect to complicated interactions between agents in MPE tasks and the resulting non-stationarity benefits more complex decomposition. Finally, VDN and QMIX perform significantly worse than the policy gradient methods (except COMA) in the sparse-reward RWARE tasks. This does not come as a surprise, since the utility of the agents is rarely greater than 0, which makes it hard to successfully learn the individual utilities.

\section{Conclusion}
\label{sec:conclusion}
We evaluated nine MARL algorithms in a total of 25 cooperative learning tasks, including combinations of partial/full observability, sparse/dense rewards, and a number of agents ranging from two to ten.
We compared algorithm performance in terms of maximum and average returns. Additionally, we further analysed the effectiveness of independent learning, centralised training, and value decomposition and identify types of environments in which each strategy is expected to perform well. 
We created EPyMARL, an open-source codebase for consistent evaluation of MARL algorithms in cooperative tasks. Finally, we implement and open-source LBF and RWARE, two new multi-agent environments which focus on sparse-reward exploration which previous environments do not cover.
Our work is limited to cooperative environments and commonly-used MARL algorithms. Competitive environments as well as solutions to a variety of MARL challenges such as exploration, communication, and opponent modelling require additional studies in the future.
We hope that our work sheds some light on the relative strengths and limitations of existing MARL algorithms, to provide guidance in terms of practical considerations and future research.

\section*{Funding Disclosure}
This research was in part financially supported by the UK EPSRC Centre for Doctoral Training in Robotics and Autonomous Systems (G.P., F.C.), and the Edinburgh University Principal’s Career Development Scholarship (L.S.).

\bibliographystyle{plainnat} 
\bibliography{references}

\begin{thebibliography}{52}
\providecommand{\natexlab}[1]{#1}
\providecommand{\url}[1]{\texttt{#1}}
\expandafter\ifx\csname urlstyle\endcsname\relax
  \providecommand{\doi}[1]{doi: #1}\else
  \providecommand{\doi}{doi: \begingroup \urlstyle{rm}\Url}\fi

\bibitem[Agogino and Tumer(2008)]{agogino2008analyzing}
Adrian~K Agogino and Kagan Tumer.
\newblock Analyzing and visualizing multiagent rewards in dynamic and
  stochastic domains.
\newblock \emph{International Conference on Autonomous Agents and Multi-Agent
  Systems}, 2008.

\bibitem[Albrecht and Ramamoorthy(2012)]{ar2012aamas}
Stefano~V. Albrecht and Subramanian Ramamoorthy.
\newblock Comparative evaluation of {MAL} algorithms in a diverse set of ad hoc
  team problems.
\newblock \emph{International Conference on Autonomous Agents and Multi-Agent
  Systems}, 2012.

\bibitem[Albrecht and Ramamoorthy(2013)]{albrecht2013game}
Stefano~V. Albrecht and Subramanian Ramamoorthy.
\newblock A game-theoretic model and best-response learning method for ad hoc
  coordination in multiagent systems.
\newblock \emph{International Conference on Autonomous Agents and Multi-Agent
  Systems}, 2013.

\bibitem[Albrecht and Stone(2017)]{as2017aamas}
Stefano~V. Albrecht and Peter Stone.
\newblock Reasoning about hypothetical agent behaviours and their parameters.
\newblock \emph{International Conference on Autonomous Agents and Multi-Agent
  Systems}, 2017.

\bibitem[Andrychowicz et~al.(2021)Andrychowicz, Raichuk, Sta{\'n}czyk, Orsini,
  Girgin, Marinier, Hussenot, Geist, Pietquin, Michalski,
  et~al.]{andrychowicz2020matters}
Marcin Andrychowicz, Anton Raichuk, Piotr Sta{\'n}czyk, Manu Orsini, Sertan
  Girgin, Raphael Marinier, L{\'e}onard Hussenot, Matthieu Geist, Olivier
  Pietquin, Marcin Michalski, et~al.
\newblock What matters in on-policy reinforcement learning? a large-scale
  empirical study.
\newblock \emph{International Conference on Learning Representations}, 2021.

\bibitem[Bard et~al.(2020)Bard, Foerster, Chandar, Burch, Lanctot, Song,
  Parisotto, Dumoulin, Moitra, Hughes, et~al.]{bard2020hanabi}
Nolan Bard, Jakob~N. Foerster, Sarath Chandar, Neil Burch, Marc Lanctot,
  H.~Francis Song, Emilio Parisotto, Vincent Dumoulin, Subhodeep Moitra, Edward
  Hughes, et~al.
\newblock The hanabi challenge: A new frontier for ai research.
\newblock \emph{Artificial Intelligence}, page 103216, 2020.

\bibitem[Beattie et~al.(2016)Beattie, Leibo, Teplyashin, Ward, Wainwright,
  K{\"u}ttler, Lefrancq, Green, Vald{\'e}s, Sadik, et~al.]{beattie2016deepmind}
Charles Beattie, Joel~Z. Leibo, Denis Teplyashin, Tom Ward, Marcus Wainwright,
  Heinrich K{\"u}ttler, Andrew Lefrancq, Simon Green, V{\'\i}ctor Vald{\'e}s,
  Amir Sadik, et~al.
\newblock Deepmind lab.
\newblock \emph{arXiv preprint arXiv:1612.03801}, 2016.

\bibitem[{Bellemare} et~al.(2013){Bellemare}, {Naddaf}, {Veness}, and
  {Bowling}]{bellemare13arcade}
Marc~G. {Bellemare}, Yavar {Naddaf}, Joel {Veness}, and Michael {Bowling}.
\newblock The arcade learning environment: An evaluation platform for general
  agents.
\newblock \emph{Journal of Artificial Intelligence Research}, pages 253--279,
  2013.

\bibitem[Christianos et~al.(2020)Christianos, Sch\"afer, and
  Albrecht]{christianos2020shared}
Filippos Christianos, Lukas Sch\"afer, and Stefano~V. Albrecht.
\newblock Shared experience actor-critic for multi-agent reinforcement
  learning.
\newblock In \emph{Advances in Neural Information Processing Systems}, 2020.

\bibitem[Christianos et~al.(2021)Christianos, Papoudakis, Rahman, and
  Albrecht]{christianos2021scaling}
Filippos Christianos, Georgios Papoudakis, Arrasy Rahman, and Stefano~V.
  Albrecht.
\newblock Scaling multi-agent reinforcement learning with selective parameter
  sharing.
\newblock In \emph{International Conference on Machine Learning}, 2021.

\bibitem[Claus and Boutilier(1998)]{claus1998dynamics}
Caroline Claus and Craig Boutilier.
\newblock The dynamics of reinforcement learning in cooperative multiagent
  systems.
\newblock \emph{AAAI Conference on Artificial Intelligence}, 1998.

\bibitem[Dhariwal et~al.(2017)Dhariwal, Hesse, Klimov, Nichol, Plappert,
  Radford, Schulman, Sidor, Wu, and Zhokhov]{baselines}
Prafulla Dhariwal, Christopher Hesse, Oleg Klimov, Alex Nichol, Matthias
  Plappert, Alec Radford, John Schulman, Szymon Sidor, Yuhuai Wu, and Peter
  Zhokhov.
\newblock {OpenAI} baselines, 2017.

\bibitem[Du et~al.(2019)Du, Han, Fang, Liu, Dai, and Tao]{du2019liir}
Yali Du, Lei Han, Meng Fang, Ji~Liu, Tianhong Dai, and Dacheng Tao.
\newblock {LIIR}: Learning individual intrinsic reward in multi-agent
  reinforcement learning.
\newblock \emph{Advances in Neural Information Processing Systems}, 2019.

\bibitem[Duan et~al.(2016)Duan, Chen, Houthooft, Schulman, and
  Abbeel]{duan2016benchmarking}
Yan Duan, Xi~Chen, Rein Houthooft, John Schulman, and Pieter Abbeel.
\newblock Benchmarking deep reinforcement learning for continuous control.
\newblock \emph{International Conference on Machine Learning}, 2016.

\bibitem[Engstrom et~al.(2019)Engstrom, Ilyas, Santurkar, Tsipras, Janoos,
  Rudolph, and Madry]{engstrom2019implementation}
Logan Engstrom, Andrew Ilyas, Shibani Santurkar, Dimitris Tsipras, Firdaus
  Janoos, Larry Rudolph, and Aleksander Madry.
\newblock Implementation matters in deep {RL}: A case study on {PPO} and
  {TRPO}.
\newblock \emph{International Conference on Learning Representations}, 2019.

\bibitem[Foerster et~al.(2016)Foerster, Assael, De~Freitas, and
  Whiteson]{foerster2016learning}
Jakob~N. Foerster, Ioannis~Alexandros Assael, Nando De~Freitas, and Shimon
  Whiteson.
\newblock Learning to communicate with deep multi-agent reinforcement learning.
\newblock \emph{Advances in Neural Information Processing Systems}, 2016.

\bibitem[Foerster et~al.(2018)Foerster, Farquhar, Afouras, Nardelli, and
  Whiteson]{foerster2018counterfactual}
Jakob~N. Foerster, Gregory Farquhar, Triantafyllos Afouras, Nantas Nardelli,
  and Shimon Whiteson.
\newblock Counterfactual multi-agent policy gradients.
\newblock \emph{AAAI Conference on Artificial Intelligence}, 2018.

\bibitem[Hasselt(2010)]{hasselt2010double}
Hado Hasselt.
\newblock Double q-learning.
\newblock \emph{Advances in Neural Information Processing Systems}, 2010.

\bibitem[He et~al.(2016)He, Boyd-Graber, Kwok, and
  Daum{\'e}~III]{he2016opponent}
He~He, Jordan Boyd-Graber, Kevin Kwok, and Hal Daum{\'e}~III.
\newblock Opponent modeling in deep reinforcement learning.
\newblock \emph{International Conference on Machine Learning}, 2016.

\bibitem[Henderson et~al.(2018)Henderson, Islam, Bachman, Pineau, Precup, and
  Meger]{henderson2018deep}
Peter Henderson, Riashat Islam, Philip Bachman, Joelle Pineau, Doina Precup,
  and David Meger.
\newblock Deep reinforcement learning that matters.
\newblock \emph{AAAI Conference on Artificial Intelligence}, 2018.

\bibitem[Hernandez-Leal et~al.(2019)Hernandez-Leal, Kartal, and
  Taylor]{hernandez2019survey}
Pablo Hernandez-Leal, Bilal Kartal, and Matthew~E. Taylor.
\newblock A survey and critique of multiagent deep reinforcement learning.
\newblock \emph{International Conference on Autonomous Agents and Multi-Agent
  Systems}, 2019.

\bibitem[Iqbal and Sha(2019)]{iqbal2018actor}
Shariq Iqbal and Fei Sha.
\newblock Actor-attention-critic for multi-agent reinforcement learning.
\newblock \emph{International Conference on Machine Learning}, 2019.

\bibitem[Jang et~al.(2017)Jang, Gu, and Poole]{jang2016categorical}
Eric Jang, Shixiang Gu, and Ben Poole.
\newblock Categorical reparameterization with gumbel-softmax.
\newblock \emph{International Conference on Learning Representations}, 2017.

\bibitem[Jaques et~al.(2019)Jaques, Lazaridou, Hughes, Gulcehre, Ortega,
  Strouse, Leibo, and De~Freitas]{jaques2018social}
Natasha Jaques, Angeliki Lazaridou, Edward Hughes, Caglar Gulcehre, Pedro~A.
  Ortega, DJ~Strouse, Joel~Z. Leibo, and Nando De~Freitas.
\newblock Social influence as intrinsic motivation for multi-agent deep
  reinforcement learning.
\newblock \emph{International Conference on Machine Learning}, 2019.

\bibitem[Johnson et~al.(2016)Johnson, Hofmann, Hutton, and
  Bignell]{johnson2016malmo}
Matthew Johnson, Katja Hofmann, Tim Hutton, and David Bignell.
\newblock The malmo platform for artificial intelligence experimentation.
\newblock In \emph{International Joint Conference on Artificial Intelligence},
  2016.

\bibitem[Lillicrap et~al.(2015)Lillicrap, Hunt, Pritzel, Heess, Erez, Tassa,
  Silver, and Wierstra]{lillicrap2015continuous}
Timothy~P. Lillicrap, Jonathan~J. Hunt, Alexander Pritzel, Nicolas Heess, Tom
  Erez, Yuval Tassa, David Silver, and Daan Wierstra.
\newblock Continuous control with deep reinforcement learning.
\newblock \emph{arXiv preprint arXiv:1509.02971}, 2015.

\bibitem[Lin(1992)]{lin1992self}
Long-Ji Lin.
\newblock Self-improving reactive agents based on reinforcement learning,
  planning and teaching.
\newblock \emph{Machine Learning}, page 293–321, 1992.

\bibitem[Lowe et~al.(2017)Lowe, Wu, Tamar, Harb, Abbeel, and
  Mordatch]{lowe2017multi}
Ryan Lowe, Yi~Wu, Aviv Tamar, Jean Harb, OpenAI~Pieter Abbeel, and Igor
  Mordatch.
\newblock Multi-agent actor-critic for mixed cooperative-competitive
  environments.
\newblock \emph{Advances in Neural Information Processing Systems}, 2017.

\bibitem[Maddison et~al.(2017)Maddison, Mnih, and Teh]{maddison2016concrete}
Chris~J. Maddison, Andriy Mnih, and Yee~Whye Teh.
\newblock The concrete distribution: A continuous relaxation of discrete random
  variables.
\newblock \emph{International Conference on Learning Representations}, 2017.

\bibitem[Mnih et~al.(2015)Mnih, Kavukcuoglu, Silver, Rusu, Veness, Bellemare,
  Graves, Riedmiller, Fidjeland, Ostrovski, et~al.]{mnih2015human}
Volodymyr Mnih, Koray Kavukcuoglu, David Silver, Andrei~A. Rusu, Joel Veness,
  Marc~G. Bellemare, Alex Graves, Martin Riedmiller, Andreas~K. Fidjeland,
  Georg Ostrovski, et~al.
\newblock Human-level control through deep reinforcement learning.
\newblock \emph{Nature}, pages 529--533, 2015.

\bibitem[Mnih et~al.(2016)Mnih, Badia, Mirza, Graves, Lillicrap, Harley,
  Silver, and Kavukcuoglu]{mnih2016asynchronous}
Volodymyr Mnih, Adria~Puigdomenech Badia, Mehdi Mirza, Alex Graves, Timothy~P.
  Lillicrap, Tim Harley, David Silver, and Koray Kavukcuoglu.
\newblock Asynchronous methods for deep reinforcement learning.
\newblock \emph{International Conference on Machine Learning}, 2016.

\bibitem[Mordatch and Abbeel(2017)]{mordatch2017emergence}
Igor Mordatch and Pieter Abbeel.
\newblock Emergence of grounded compositional language in multi-agent
  populations.
\newblock \emph{arXiv preprint arXiv:1703.04908}, 2017.

\bibitem[Nash(1951)]{nash1951non}
John Nash.
\newblock Non-cooperative games.
\newblock \emph{Annals of mathematics}, pages 286--295, 1951.

\bibitem[Papoudakis et~al.(2019)Papoudakis, Christianos, Rahman, and
  Albrecht]{papoudakis2019dealing}
Georgios Papoudakis, Filippos Christianos, Arrasy Rahman, and Stefano~V.
  Albrecht.
\newblock Dealing with non-stationarity in multi-agent deep reinforcement
  learning.
\newblock \emph{arXiv preprint arXiv:1906.04737}, 2019.

\bibitem[Papoudakis et~al.(2021)Papoudakis, Christianos, and
  Albrecht]{papoudakis2021local}
Georgios Papoudakis, Filippos Christianos, and Stefano~V. Albrecht.
\newblock Local information agent modelling in partially-observable
  environments.
\newblock In \emph{Advances in Neural Information Processing Systems}, 2021.

\bibitem[Paszke et~al.(2019)Paszke, Gross, Massa, Lerer, Bradbury, Chanan,
  Killeen, Lin, Gimelshein, Antiga, Desmaison, Kopf, Yang, DeVito, Raison,
  Tejani, Chilamkurthy, Steiner, Fang, Bai, and Chintala]{NEURIPS2019_bdbca288}
Adam Paszke, Sam Gross, Francisco Massa, Adam Lerer, James Bradbury, Gregory
  Chanan, Trevor Killeen, Zeming Lin, Natalia Gimelshein, Luca Antiga, Alban
  Desmaison, Andreas Kopf, Edward Yang, Zachary DeVito, Martin Raison, Alykhan
  Tejani, Sasank Chilamkurthy, Benoit Steiner, Lu~Fang, Junjie Bai, and Soumith
  Chintala.
\newblock Pytorch: An imperative style, high-performance deep learning library.
\newblock In \emph{Advances in Neural Information Processing Systems}, 2019.

\bibitem[Rahman et~al.(2021)Rahman, H\"opner, Christianos, and
  Albrecht]{rahman2021open}
Arrasy Rahman, Niklas H\"opner, Filippos Christianos, and Stefano~V. Albrecht.
\newblock Towards open ad hoc teamwork using graph-based policy learning.
\newblock In \emph{International Conference on Machine Learning}, 2021.

\bibitem[Raileanu et~al.(2018)Raileanu, Denton, Szlam, and
  Fergus]{raileanu2018modeling}
Roberta Raileanu, Emily Denton, Arthur Szlam, and Rob Fergus.
\newblock Modeling others using oneself in multi-agent reinforcement learning.
\newblock \emph{International Conference on Machine Learning}, 2018.

\bibitem[Rashid et~al.(2018)Rashid, Samvelyan, De~Witt, Farquhar, Foerster, and
  Whiteson]{rashid2018qmix}
Tabish Rashid, Mikayel Samvelyan, Christian~Schroeder De~Witt, Gregory
  Farquhar, Jakob Foerster, and Shimon Whiteson.
\newblock {QMIX}: monotonic value function factorisation for deep multi-agent
  reinforcement learning.
\newblock \emph{International Conference on Machine Learning}, 2018.

\bibitem[Resnick et~al.(2018)Resnick, Eldridge, Ha, Britz, Foerster, Togelius,
  Cho, and Bruna]{resnick2018pommerman}
Cinjon Resnick, Wes Eldridge, David Ha, Denny Britz, Jakob Foerster, Julian
  Togelius, Kyunghyun Cho, and Joan Bruna.
\newblock Pommerman: A multi-agent playground.
\newblock \emph{arXiv preprint arXiv:1809.07124}, 2018.

\bibitem[Ryu et~al.(2020)Ryu, Shin, and Park]{ryu2019multi}
Heechang Ryu, Hayong Shin, and Jinkyoo Park.
\newblock Multi-agent actor-critic with hierarchical graph attention network.
\newblock \emph{AAAI Conference on Artificial Intelligence}, 2020.

\bibitem[Samvelyan et~al.(2019)Samvelyan, Rashid, Schroeder~de Witt, Farquhar,
  Nardelli, Rudner, Hung, Torr, Foerster, and Whiteson]{samvelyan19smac}
Mikayel Samvelyan, Tabish Rashid, Christian Schroeder~de Witt, Gregory
  Farquhar, Nantas Nardelli, Tim~GJ Rudner, Chia-Man Hung, Philip~HS Torr,
  Jakob Foerster, and Shimon Whiteson.
\newblock The {StarCraft} multi-agent challenge.
\newblock \emph{International Conference on Autonomous Agents and Multi-Agent
  Systems}, 2019.

\bibitem[Schulman et~al.(2017)Schulman, Wolski, Dhariwal, Radford, and
  Klimov]{schulman2017proximal}
John Schulman, Filip Wolski, Prafulla Dhariwal, Alec Radford, and Oleg Klimov.
\newblock Proximal policy optimization algorithms.
\newblock \emph{arXiv preprint arXiv:1707.06347}, 2017.

\bibitem[Silver et~al.(2014)Silver, Lever, Heess, Degris, Wierstra, and
  Riedmiller]{silver2014deterministic}
David Silver, Guy Lever, Nicolas Heess, Thomas Degris, Daan Wierstra, and
  Martin Riedmiller.
\newblock Deterministic policy gradient algorithms.
\newblock \emph{International Conference on Machine Learning}, 2014.

\bibitem[Sukhbaatar et~al.(2016)Sukhbaatar, Fergus,
  et~al.]{sukhbaatar2016learning}
Sainbayar Sukhbaatar, Rob Fergus, et~al.
\newblock Learning multiagent communication with backpropagation.
\newblock \emph{Advances in Neural Information Processing Systems}, 2016.

\bibitem[Sunehag et~al.(2018)Sunehag, Lever, Gruslys, Czarnecki, Zambaldi,
  Jaderberg, Lanctot, Sonnerat, Leibo, Tuyls, et~al.]{sunehag2017value}
Peter Sunehag, Guy Lever, Audrunas Gruslys, Wojciech~Marian Czarnecki, Vinicius
  Zambaldi, Max Jaderberg, Marc Lanctot, Nicolas Sonnerat, Joel~Z. Leibo, Karl
  Tuyls, et~al.
\newblock Value-decomposition networks for cooperative multi-agent learning.
\newblock \emph{International Conference on Autonomous Agents and Multi-Agent
  Systems}, 2018.

\bibitem[Sutton(1988)]{sutton1988learning}
Richard~S. Sutton.
\newblock Learning to predict by the methods of temporal differences.
\newblock \emph{Machine learning}, 1988.

\bibitem[Tan(1993)]{tan1993multi}
Ming Tan.
\newblock Multi-agent reinforcement learning: Independent vs. cooperative
  agents.
\newblock \emph{International Conference on Machine Learning}, 1993.

\bibitem[Vinyals et~al.(2017)Vinyals, Ewalds, Bartunov, Georgiev, Vezhnevets,
  Yeo, Makhzani, K{\"u}ttler, Agapiou, Schrittwieser,
  et~al.]{vinyals2017starcraft}
Oriol Vinyals, Timo Ewalds, Sergey Bartunov, Petko Georgiev, Alexander~Sasha
  Vezhnevets, Michelle Yeo, Alireza Makhzani, Heinrich K{\"u}ttler, John
  Agapiou, Julian Schrittwieser, et~al.
\newblock {StarCraft} {II}: A new challenge for reinforcement learning.
\newblock \emph{arXiv preprint arXiv:1708.04782}, 2017.

\bibitem[Wang et~al.(2019)Wang, Bao, Clavera, Hoang, Wen, Langlois, Zhang,
  Zhang, Abbeel, and Ba]{wang2019benchmarking}
Tingwu Wang, Xuchan Bao, Ignasi Clavera, Jerrick Hoang, Yeming Wen, Eric
  Langlois, Shunshi Zhang, Guodong Zhang, Pieter Abbeel, and Jimmy Ba.
\newblock Benchmarking model-based reinforcement learning.
\newblock \emph{arXiv preprint arXiv:1907.02057}, 2019.

\bibitem[Watkins and Dayan(1992)]{watkins1992q}
Christopher J. C.~H. Watkins and Peter Dayan.
\newblock Q-learning.
\newblock \emph{Machine Learning}, 1992.

\bibitem[Yu et~al.(2021)Yu, Velu, Vinitsky, Wang, Bayen, and
  Wu]{yu2021surprising}
Chao Yu, Akash Velu, Eugene Vinitsky, Yu~Wang, Alexandre Bayen, and Yi~Wu.
\newblock The surprising effectiveness of mappo in cooperative, multi-agent
  games.
\newblock \emph{arXiv preprint arXiv:2103.01955}, 2021.

\end{thebibliography}

\clearpage

\appendix

\section*{Responsibility Statement}

The authors bear all responsibility in case of violation of rights in the proposed environments and the included benchmark. All the proposed environments use an MIT license for their source code.

\section{New Multi-Agent Reinforcement Learning Environments}
\label{app:new_envs}
As part of this work, we developed and open-sourced two novel MARL environments focusing on cooperation and sparse-rewards. In this supplementary section, we will provide details about both added environments including our motivation for creating them, accessibility and licensing, installation information, a description of their interface and code snippets to get started as well as further details on their observations, reward functions and dynamics. For high-level descriptions, see \Cref{sec:lbf,sec:rware} and see \Cref{appendix:tasks} for more information on the specific tasks used for the benchmark.

\subsection{Motivation}
Environments for MARL evaluation are scattered and few environments have been established as standard benchmarking problems. Most notably, the Starcraft Multi-Agent Challenge (SMAC)~\citep{samvelyan19smac} and Multi-Agent Particle Environment (MPE)~\citep{mordatch2017emergence} are prominent examples for MARL environments. While more environments exist, they often represent different types of games such as turn-based board games~\citep{bard2020hanabi} or contain image frames as observations~\citep{johnson2016malmo,beattie2016deepmind,resnick2018pommerman}. Environments with these properties often require further solutions not specific to MARL. Additionally, we found that the core challenge of exploration, for which a multitude of environments exist for single-agent RL research (e.g. Atari games such as Montezuma's Revenge~\citep{bellemare13arcade}), is underrepresented in MARL environments. The Level-Based Foraging and Multi-Robot Warehouse environments aim to represent sparse-reward hard exploration problems which require significant cooperation across agents. Both environments are flexible in their configurations to enable partial- or full-observability, fully-cooperative or mixed reward settings and allow faster training than a multitude of other environments (see \Cref{app:env_speed_comparison} for a comparison of simulation speeds across all environments used in the benchmark).

\subsection{Accessibility and Licensing}
Both environments are publicly available on GitHub under the following links.

\begin{table}[h]
\centering
\caption{\label{tab:env_links}GitHub repositories for Level-Based Foraging and Multi-Robot Warehouse environments.}
    \begin{tabular}{@{}llllll@{}}
        \toprule
        Level-Based Foraging & \url{https://github.com/uoe-agents/lb-foraging}\\ \midrule
        Multi-Robot Warehouse & \url{https://github.com/uoe-agents/robotic-warehouse}\\ \bottomrule
    \end{tabular}
\end{table}

The environments are licensed under the MIT license which can be found in the \texttt{LICENSE} file within respective repositories. Both environments will be supported and maintained by the authors as needed.

\subsection{Installation}
Our novel environments can be installed as Python packages. We recommend users to setup a virtual environment to manage packages and dependencies for individual projects, e.g.\ using \href{https://docs.python.org/3/tutorial/venv.html}{\texttt{venv}} or \href{https://www.anaconda.com/products/individual}{Anaconda}. Then the code repository can be cloned using git and installed as a package using the Python package manager \href{https://docs.python.org/3/installing/index.html}{\texttt{pip}} as follows at the example of the Multi-Robot Warehouse environment:

\begin{minted}[formatcom=\BashFancyFormatLine]{bash}
git clone git@github.com:uoe-agents/robotic-warehouse.git
cd robotic-warehouse
pip install -e .
\end{minted}

In order to install the Level-Based Foraging environment, execute the following, similar code:
\begin{minted}[formatcom=\BashFancyFormatLine]{bash}
git clone git@github.com:uoe-agents/lb-foraging.git
cd lb-foraging
pip install -e .
\end{minted}

\subsection{Environment Interface}
Both environments follow the interface framework of \href{https://gym.openai.com/}{OpenAI's Gym}. Below, we will piece-by-piece explain the interface and commands needed to interact with the installed environment within Python.

\paragraph{Package import} First, the installed package and \texttt{gym} (installed above as dependency) need to be imported (shown at the example of the Multi-Robot Warehouse):
\begin{minted}{python}
import gym
import robotic_warehouse
\end{minted}

\paragraph{Environment creation} Following the import of the required packages, the environment can be instantiated. A large selection of configurations for both environments are already registered to \texttt{gym} upon importing the package. These can simply be created:
\begin{minted}{python}
env = gym.make("rware-tiny-2ag-v1")
\end{minted}

For an overview over the naming of these environment configuration names, see \Cref{app:env_names}.

\paragraph{Start an episode} A new episode within the environment can be started with the following command:
\begin{minted}{python}
obs = env.reset()
\end{minted}
This function returns the initial state or observation of the environment, indicating $s_0$.

\paragraph{Environment steps} In order to interact with the environment, an action for each agent should be provided. In the case of the \pyth{"rware-tiny-2ag-v1"}, there are two agents in the environment. Therefore, the environment expects to receive an action for all agents. In order to gain insight into the shape of expected actions or received observations, the respective spaces of the environment can be inspected:

\begin{minted}{python}
env.action_space  # Tuple(Discrete(5), Discrete(5))
env.observation_space  # Tuple(Box(XX,), Box(XX,))
\end{minted}

Using the action space of the environment, we can sample random valid actions and take a step in the environment:

\begin{minted}{python}
actions = env.action_space.sample()  # the action space can be sampled
print(actions)                       # e.g. (1, 0)
next_obs, reward, done, info = env.step(actions)

print(done)    # [False, False]
print(reward)  # [0.0, 0.0]
\end{minted}
Note, that for these multi-agent environments, actions are a \pyth{list} or \pyth{tuple} containing individual actions for all agents. Similarly, the received values are also lists of observations at the next timestep (\pyth{next_obs}), rewards for the completed transition (\pyth{reward}), flags indicating whether the episode has terminated (\pyth{done}) and an additional dictionary (\pyth{info}) which may contain meta-information on the transition or episode.

\paragraph{Rendering}
Both novel environments support rendering to visualise states and inspect agents' behaviour. See \Cref{fig:env_observations} for exemplary visualisations of these environments.

\begin{minted}{python}
env.render()
\end{minted}

\paragraph{Single episode} We can put all these steps together for a script which executes a single episode using random actions and rendering the environment:

\begin{minted}{python}
import gym
import robotic_warehouse

env = gym.make("rware-tiny-2ag-v1")

obs = env.reset()
done = [False] * env.n_agents

while not all(done):
    actions = env.action_space.sample()
    next_obs, reward, done, info = env.step(actions)
    env.render()

env.close()
\end{minted}

\subsection{Environment Naming}
\label{app:env_names}
In this section, we will briefly outline the configuration possibilities for Level-Based Foraging and Multi-Robot Warehouse environments.

\paragraph{Level-Based Foraging} For the Level-Based Foraging environment, we can create an environment as follows
\begin{minted}[fontsize=\footnotesize]{python}
env = gym.make("Foraging<obs>-<x_size>x<y_size>-<n_agents>p-<food>f<force_c>-v1")
\end{minted}
with the following options for each field:
\begin{itemize}
    \item \mintinline{python}{<obs>}: This optional field can either be empty (\mintinline{python}{""}) or indicate a partially observable task with visibility radius of two fields (\mintinline{python}{"-2s}).
    \item \mintinline{python}{<x_size>}: This field indicates the horizontal size of the environment map and can by default take any values between $5$ and $20$.
    \item \mintinline{python}{<y_size>}: This field indicates the vertical size of the environment map and can by default take any values between $5$ and $20$. It should be noted, that upon import only environments with square dimensions (\mintinline{python}{<x_size> = <y_size>}) are registered and ready for creation.
    \item \mintinline{python}{<n_agents>}: This field indicates the number of agents within the environment. By default, any values between $2$ and $5$ are automatically registered.
    \item \mintinline{python}{<food>}: This field indicates the number of food items scattered within the environment. It can take any values between $1$ and $10$ by default.
    \item \mintinline{python}{<force_c>}: This optional field can either be empty (\mintinline{python}{""}) or indicate a task with only "cooperative food" (\mintinline{python}{"-coop"}. In the latter case, the environment will only contain food of a level such that all agents have to cooperate in order to pick the food up. This mode should only be used with up to four agents.
\end{itemize}

In order to register environments of more different configurations, see the \href{https://github.com/uoe-agents/lb-foraging/blob/master/lbforaging/__init__.py}{current registration}.

\paragraph{Multi-Robot Warehouse} For the Multi-Robot Warehouse environment, we can create an environment as follows
\begin{minted}[fontsize=\footnotesize]{python}
env = gym.make("rware-<size>-<num_agents>ag<diff>-v1")
\end{minted}
with the following options for each field:
\begin{itemize}
    \item \mintinline{python}{<size>}: This field represents the size of the warehouse. By default the size can take on four values: \mintinline{python}{"tiny"}, \mintinline{python}{"small"}, \mintinline{python}{"medium"} and \mintinline{python}{"large"}. These size identifiers define the number of rows and columns of groups of shelves within the warehouse and set these to be (1, 3), (2, 3), (2, 5) and (3, 5) respectively. By default, each group of shelves consists of 16 shelves organised in a $8 \times 2$ grid.
    \item \mintinline{python}{<num_agents>}: This field indicates the number of agents and can by default take any values between $1$ and $20$.
    \item \mintinline{python}{<diff>}: This optional field can indicate changes in the difficulty of the environment given by the total number of requests at a time. Agents have to collect and deliver specific requested shelves. By default, there are $N$ requests at each point in time with $N$ being the number of agents. With this field, the number of requests can be set to half (\mintinline{python}{"-hard"}) or double (\mintinline{python}{"-easy"}) the number of agents.
\end{itemize}
Additionally, by default agents observe only fields within immediate grids next to their location and episodes terminate after $500$ steps. For a more extensive set of configurations, including variations of visibility, see the \href{https://github.com/uoe-agents/robotic-warehouse/blob/master/robotic_warehouse/__init__.py#L36}{full registration}.

\subsection{Environment Details - Level-Based Foraging}
Below, we will provide additional details to observations, actions and dynamics for the Level-Based Foraging environment.

\begin{figure}[t]
    \centering
    \begin{subfigure}{.49\linewidth}
         \centering
         \includegraphics[height=5cm]{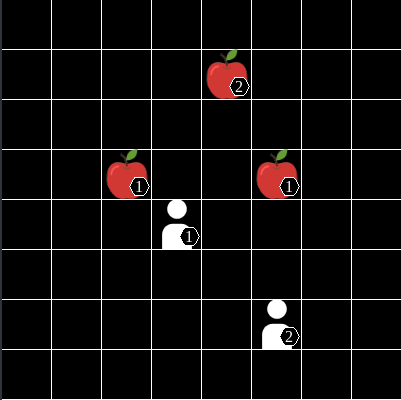}
         \caption{\texttt{Foraging-8x8-2p-3f-v1}}
         \label{fig:lbf_obs_full}
    \end{subfigure}
    \hfill
    \begin{subfigure}{.49\linewidth}
        \centering
        \includegraphics[height=5cm]{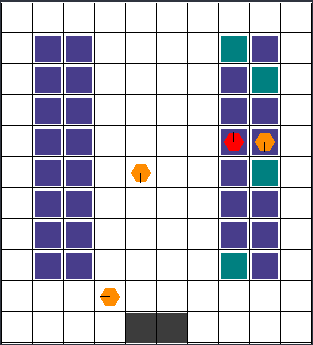}
        \caption{\texttt{rware-tiny-4ag-v1}}
        \label{fig:rware_observations}
    \end{subfigure}
    \caption{Environment renderings matching observations for (a) Level-Based Foraging and (b) Multi-Robot Warehouse.}
    \label{fig:env_observations}
\end{figure}

\paragraph{Observations} As seen above, agents receive observations at each timestep which correspond to the full state of the environment or a partial view in the case of partial observability. Below, we will outline a realistic observation at the example of the \texttt{Foraging-8x8-2p-3f-v1} task visualised in \Cref{fig:lbf_obs_full}:

\begin{minted}[fontsize=\footnotesize]{python}
(
    array([1., 4., 2., 3., 2., 1., 3., 5., 1., 6., 6., 2., 4., 4., 1.],
          dtype=float32),
    array([1., 4., 2., 3., 2., 1., 3., 5., 1., 4., 4., 1., 6., 6., 2.],
          dtype=float32)
)
\end{minted}
The observation consists of two arrays, each corresponding to the observation of one of the two agents within the environment. Within that vector, triplets of the form (x, y, level) are written. Specifically, the first three (number of food items in the environment) triplets for a total of 9 elements contain the x and y coordinates and level of each food item, and the following two (number of agents) triplets have the respective values for each agent. The coordinates always start from the bottom left square in the observability radius of the agent. When food items or agents are not visible, either because they are outside of the observable radius or the food has been picked up, then the respective values are replaced with (-1, -1, 0).

\paragraph{Actions} In Level-Based Foraging environments, each agent has six possible discrete actions to choose at each timestep:
\begin{equation*}
    A^i = \{\text{Noop, Move North, Move South, Move West, Move East, Pickup}\}
\end{equation*}
While the first action corresponds to the action simply staying within its grid and the last action being used to pickup nearby food, the remaining four actions encode discrete 2D navigation. Upon choosing these actions, the agent moves a single cell in the chosen direction within the grid.

\paragraph{Rewards} Agents only receive non-zero reward for picking up food within the Level-Based Foraging environment. The reward for picking up food depends on the level of the collected food as well as the level of each contributing agent. The reward of agent $i$ for picking up food is defined as
\begin{equation*}
    r^i = \frac{FoodLevel * AgentLevel}{\sum{FoodLevels}\sum{LoadingAgentsLevel}}
\end{equation*}
with normalisation. Rewards are normalised to ensure that the sum of all agents' returns on a solved episode equals to one.

\paragraph{Dynamics} The transition function of the Level-Based Foraging domain is fairly straightforward. Agents transition to cells following their movement based on chosen actions. Furthermore, agents successfully collect food as long as the sum of the levels of all loading agents is greater or equal to the level of the food.

\subsection{Environment Details - Multi-Robot Warehouse}
Now, we will provide additional details to observations, actions and dynamics for the Multi-Robot Warehouse environment.

\paragraph{Observations}
Agent observations for the Multi-Robot Warehouse environment are defined to be partially observable and contain all information about cells in the immediate proximity of an agent. By default, each agent observes information within a $3 \times 3$ square centred on the agent, but the visibility range can be modified as an environment parameter. Agents observe their own location, rotation and load, the location and rotation of other observed agents as well as nearby shelves with information whether those are requested or not. For the precise details, see the example below matching the state visualised in \Cref{fig:rware_observations}:

\begin{minted}[fontsize=\footnotesize]{python}
(
    array([8., 3., 0., 0., 1., 0., 0., 0., 0., 1., 0., 0., 0., 1., 0., 0., 1.,
           0., 0., 0., 1., 1., 0., 1., 0., 0., 0., 0., 0., 0., 1., 0., 0., 0.,
           1., 0., 1., 0., 1., 0., 0., 1., 0., 0., 1., 0., 0., 0., 0., 0., 1.,
           1., 0., 0., 0., 1., 0., 0., 1., 0., 0., 0., 1., 0., 0., 1., 0., 0.,
           0., 0., 0.], dtype=float32),
    array([4., 9., 0., 0., 0., 1., 0., 1., 0., 1., 0., 0., 0., 0., 0., 0., 1.,
           0., 0., 0., 0., 0., 0., 1., 0., 0., 0., 0., 0., 0., 1., 0., 0., 0.,
           0., 0., 1., 0., 0., 1., 0., 0., 0., 0., 1., 0., 0., 0., 0., 0., 0.,
           1., 0., 0., 0., 0., 0., 0., 1., 0., 0., 0., 0., 0., 0., 1., 0., 0.,
           0., 0., 0.], dtype=float32),
    ...
)
\end{minted}
Again, each element of the list is the observation that corresponds to each of the agents. The first three values of the vectors correspond to the agent itself with the x and y coordinates, and a value of ``1'' or ``0'' depending on whether the agent is currently carrying a shelf. The next four values are a one-hot encoding of the direction the current agent is facing (up/down/left/right respectively for each item in the one-hot encoding). Then, a single value of ``1'' or ``0'' represents whether the agent is currently in a location that acts as a path and therefore not allowed to place shelves on that location. The rest of the values can be split into 9 groups of 7 elements, each group corresponding to a square in the observation radius (3x3 centred around the agent - can be increased from the default for more visibility). In this group, the elements are only ones and zeros and correspond to: agent exists in the square, one-hot encoding (4 elements) of direction of agent (if exists), shelf exists in the square, shelf (if exists) is currently requested to be delivered to the goal location.

\paragraph{Actions} The action space within the Multi-Robot Warehouse environment is very similar to the Level-Based Foraging domain with four discrete actions to choose from:
\begin{equation*}
    A^i = \{\text{Turn Left, Turn Right, Move Forward, Load/ Unload Shelf}\}
\end{equation*}
Agents also have to navigate a 2D grid-world, but they are only able to move forward or rotate to change their orientation in $90~\degree$ steps in either direction. Agents are unable to move upon cells which are already occupied. Besides movement, agents are only able to load and unload shelves. Loading shelves is only possible when the agent is located at the location of an unloaded shelf. Similarly, agents are only able to unload a currently loaded shelf on a location where no shelf is located but is within a group where initially a shelf has been stored.

\paragraph{Rewards}
At each time, a set number of shelves $R$ is requested. Agents are rewarded with a reward of $1$ for successfully delivering a requested shelf to a goal location at the bottom of the warehouse. A significant challenge in this environment is for agents to successfully deliver requested shelves but also finding an empty location to return the previously delivered shelf. Without unloading the previously delivered shelf, the agent is unable to collect new requested shelves. Having multiple steps between deliveries leads to a very sparse reward signal.

\paragraph{Dynamics}
Agents move through the grid-world as expected given their chosen actions. Whenever multiple agents collide, i.e.\ they attempt to move to the same location, movement is resolved in a way to maximise mobility. When two or more agents attempt to move to the same location, we prioritise agents to move which also blocks other agents. Otherwise, the selection is done arbitrarily.
Additionally, it is worth noting that a shelf is uniformly sampled and added to the list of currently requested shelves whenever a previously requested shelf is delivered to a goal location. Therefore, $R$ requested shelves are constantly ensured.
Note that $R$ directly affects the difficulty of the environment. A small $R$, especially on a larger grid, dramatically affects the sparsity of the reward and thus makes exploration more difficult as randomly delivering a requested shelf becomes increasingly improbable.

\subsection{Environments Simulation Speed Comparison}
\label{app:env_speed_comparison}
\begin{figure}[t]
    \centering
    \includegraphics[width=\textwidth]{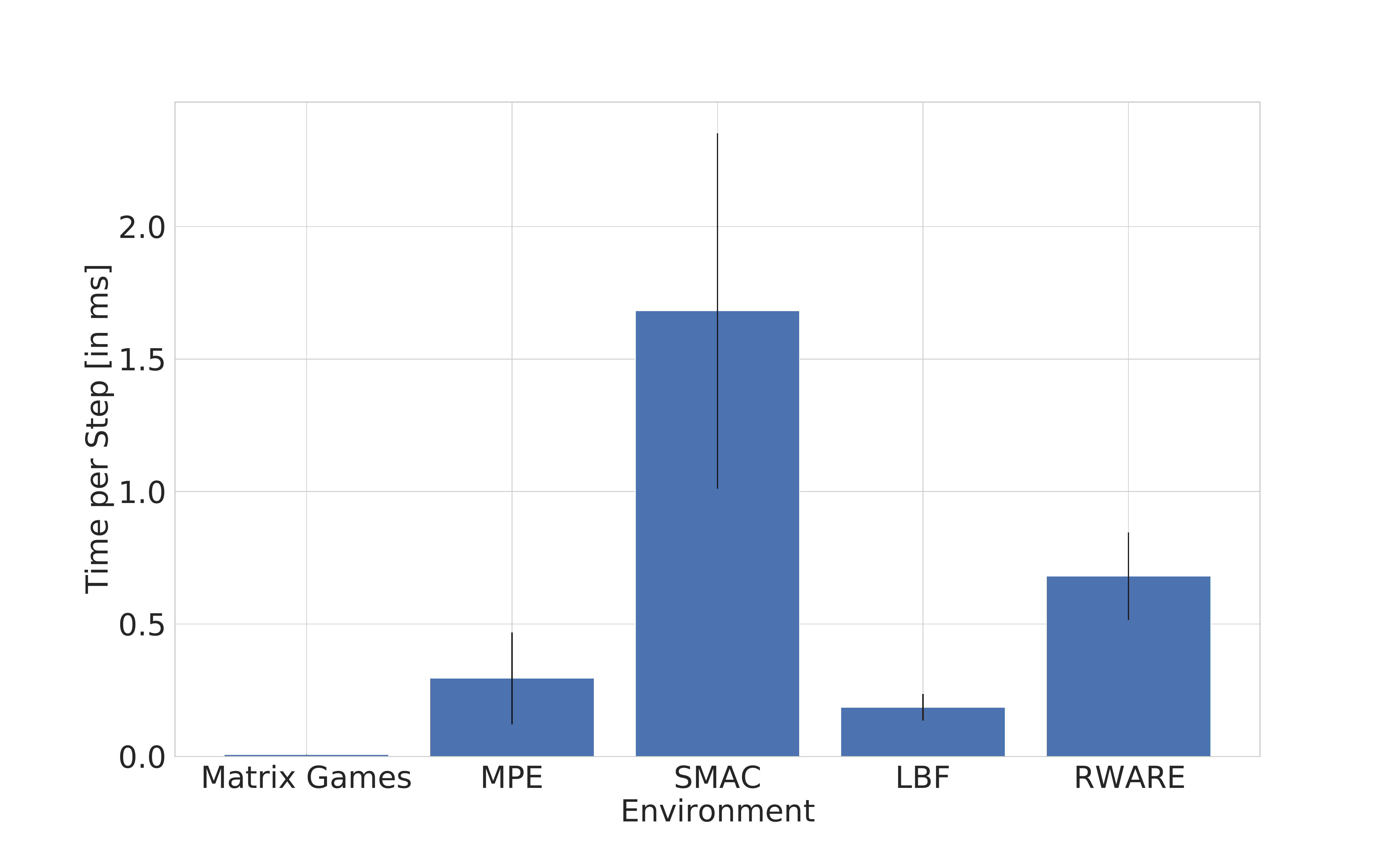}
    \caption{Mean simulation time per step for all environments. Bars indicate standard deviations of simulation speed across all tasks within the environments.}
    \label{fig:env_speedtest_barchart}
\end{figure}

In order to demonstrate and compare simulation speed of all considered environments within the conducted benchmark, we simulate 10,000 environments steps using random action selection. Environments were simulated without rendering to represent conditions at training or evaluation time. Total time and time per step are reported in \Cref{tab:env_speed_comparison}. Additionally, mean simulation time per step with standard deviations across for all environments are reported in \Cref{fig:env_speedtest_barchart}.

While matrix games are unsurprisingly the fastest environments to simulate due to their simplicity, Level-Based Foraging tasks are faster to simulate compared to all other environments aside simplest Multi-Agent Particle environment tasks. Despite their arguably more complex environments, the Multi-Robot Warehouse environment is only marginally more expensive to simulate compared to the Multi-Agent Particle environment. It is also worth noting that the cost of simulation of Level-Based Foraging and Multi-Robot Warehouse tasks mostly depends on the number of agents. This allows to simulate large warehouses without additional cost. Unsurprisingly, the Starcraft Multi-Agent Challenge is by far the most expensive environment to simulate as it requires to run the complex game of StarCraft II.

Speed comparisons are conducted on a personal computer running Ubuntu 20.04.2 with a Intel Core i7-10750H CPU (six cores at 2.60GHz) and 16GB of RAM. Simulation was executed on a single CPU thread.

\begin{table}[t]
\centering
\caption{\label{tab:env_speed_comparison} Simulation time for 10,000 steps and time per step in all 25 tasks.}
\small
\robustify\bf
\begin{tabular}{p{1em} l S S S}
\toprule
& \textbf{Tasks}  & \mbox{\specialcell{Number of \\ agents}} & \mbox{\specialcell{Total time \\ {[in s]}}} & \mbox{\specialcell{Time per step \\ {[in ms]}}} \\ \\
\hline \\
\multirow{6}{*}{\rotatebox[origin=c]{90}{Matrix Games}} & Climbing          & 2     & 0.079     & 0.008 \\
                                                        & Penalty k=0       & 2     & 0.079     & 0.008 \\
                                                        & Penalty k=-25     & 2     & 0.078     & 0.008 \\
                                                        & Penalty k=-50     & 2     & 0.080     & 0.008 \\
                                                        & Penalty k=-25     & 2     & 0.081     & 0.008 \\
                                                        & Penalty k=-25     & 2     & 0.078     & 0.008 \\ \\
\hline \\
\multirow{4}{*}{\rotatebox[origin=c]{90}{MPE}}          & Speaker-Listener  & 2     & 0.852     & 0.085 \\
                                                        & Spread            & 3     & 4.395     & 0.439 \\
                                                        & Adversary         & 3     & 1.651     & 0.165 \\
                                                        & Tag               & 4     & 4.911     & 0.491 \\ \\
\hline \\
\multirow{5}{*}{\rotatebox[origin=c]{90}{SMAC}}         & 2s\_vs\_1sc       & 2     & 7.798     & 0.780 \\
                                                        & 3s5z              & 8     & 18.656    & 1.866 \\
                                                        & MMM2              & 10    & 22.480    & 2.248 \\
                                                        & corridor          & 6     & 24.890    & 2.489 \\
                                                        & 3s\_vs\_5z        & 3     & 10.266    & 1.027 \\ \\
\hline \\
\multirow{7}{*}{\rotatebox[origin=c]{90}{LBF}}          &  15x15-4p-3f      & 4     & 2.516     & 0.252 \\
                                                        & 8x8-2p-2f-2s-c    & 2     & 1.210     & 0.121 \\
                                                        & 10x10-3p-3f-2s    & 3     & 1.662     & 0.166 \\
                                                        & 8x8-2p-2f-c       & 2     & 1.253     & 0.125 \\
                                                        & 15x15-4p-5f       & 4     & 2.559     & 0.256 \\
                                                        & 15x15-3p-5f       & 3     & 1.939     & 0.194 \\
                                                        & 10x10-3p-3f       & 3     & 1.849     & 0.185 \\ \\
\hline \\
\multirow{3}{*}{\rotatebox[origin=c]{90}{RWARE}}        & Tiny 2p           & 2     & 4.469     & 0.447 \\
                                                        & Tiny 4p           & 4     & 7.976     & 0.798 \\
                                                        & Small 4p          & 4     & 7.974     & 0.797 \\ \\
\bottomrule
\end{tabular}
\end{table}

\section{The EPyMARL Codebase}

As part of this work we extended the well-known PyMARL codebase \citep{samvelyan19smac} to include more algorithms, support more environments as well as allow for more flexible tuning of the implementation details. 

\subsection{Motivation}

Implementation details in Deep RL algorithms significantly affect their achieved returns~\citep{engstrom2019implementation}. This problem becomes even more apparent in deep MARL research, where the existence of multiple agents significantly increases the amount of implementation details that affect the performance. Therefore, it is often difficult to measure the benefits of newly proposed algorithms compared to existing ones. We believe that a unified codebase, which implements the majority of MARL algorithms used as building blocks in research, would significantly benefit the community. Finally, EPyMARL allows for tuning of additional implementation details compared to PyMARL, including parameter sharing, reward standardisation and entropy regularisation.

\subsection{Accessibility and Licensing}

All code for EPyMARL is publicly available open-source on GitHub under the following link:
\url{https://github.com/uoe-agents/epymarl}.  

All source code that has been taken from the PyMARL repository was licensed (and remains so) under the Apache License v2.0 (included in LICENSE file). Any new code is also licensed under the Apache License v2.0. The NOTICE file in the GitHub repository contains information about the files that have been added or modified compared to the original PyMARL codebase. %

\subsection{Installation}
In the GitHub repository of EPyMARL, the file \emph{requirements.txt} contains all Python packages that have to be install as dependencies in order to use the codebase. We recommend using a Python virtual environment for training MARL algorithm using EPyMARL. After activating the virtual environment, the following commands will install the required packages

\begin{minted}[formatcom=\BashFancyFormatLine, breaklines]{bash}
git clone git@github.com:uoe-agents/epymarl.git
cd epymarl
pip install -r requirements.txt
\end{minted}

\subsection{Execution}

EPyMARL is written in Python 3. The neural networks and their operations are implemented using the Pytorch framework \citep{NEURIPS2019_bdbca288}.
To train an algorithm (QMIX in this example) in a Gym-based environment, execute the following command from the home folder of EPyMARL:

\begin{minted}[formatcom=\BashFancyFormatLine, breaklines]{bash}
python3 src/main.py --config=qmix --env-config=gymma with env_args.time_limit=50 env_args.key="lbforaging:Foraging-8x8-2p-3f-v0"
\end{minted}

where \texttt{config} is the configuration of the algorithm, \texttt{gymma} is a Gym compatible wrapper, \texttt{env\_args.time\_limit} is the time limit of the task (number of steps before the episode ends), and \texttt{env\_args.key} is the name of the task. Default configuration for all algorithms can be found in the \texttt{src/config/algs} folder of the codebase.

\section{Task Specifications}
\label{appendix:tasks}
Below, we provide details and descriptions of the environments and tasks used for the evaluation.

\begin{figure}[t]
  \begin{subfigure}{0.24\linewidth}
    \centering
    \fbox{\includegraphics[width=0.9\linewidth]{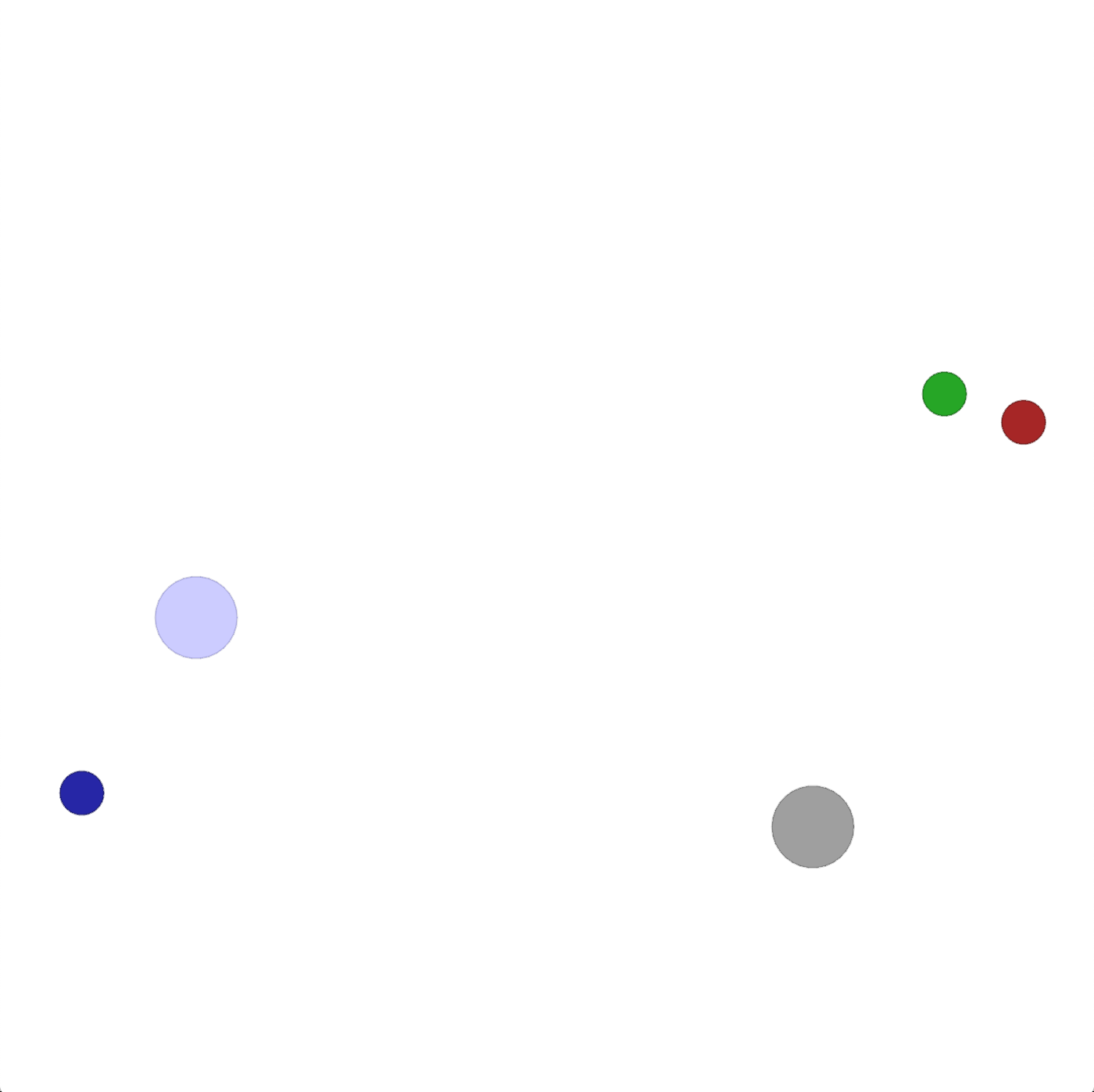}}
    \label{fig:mpe_speaker_listener}
    \caption{}
  \end{subfigure}
  \hfill
  \begin{subfigure}{0.24\linewidth}
    \centering
    \fbox{\includegraphics[width=0.9\linewidth]{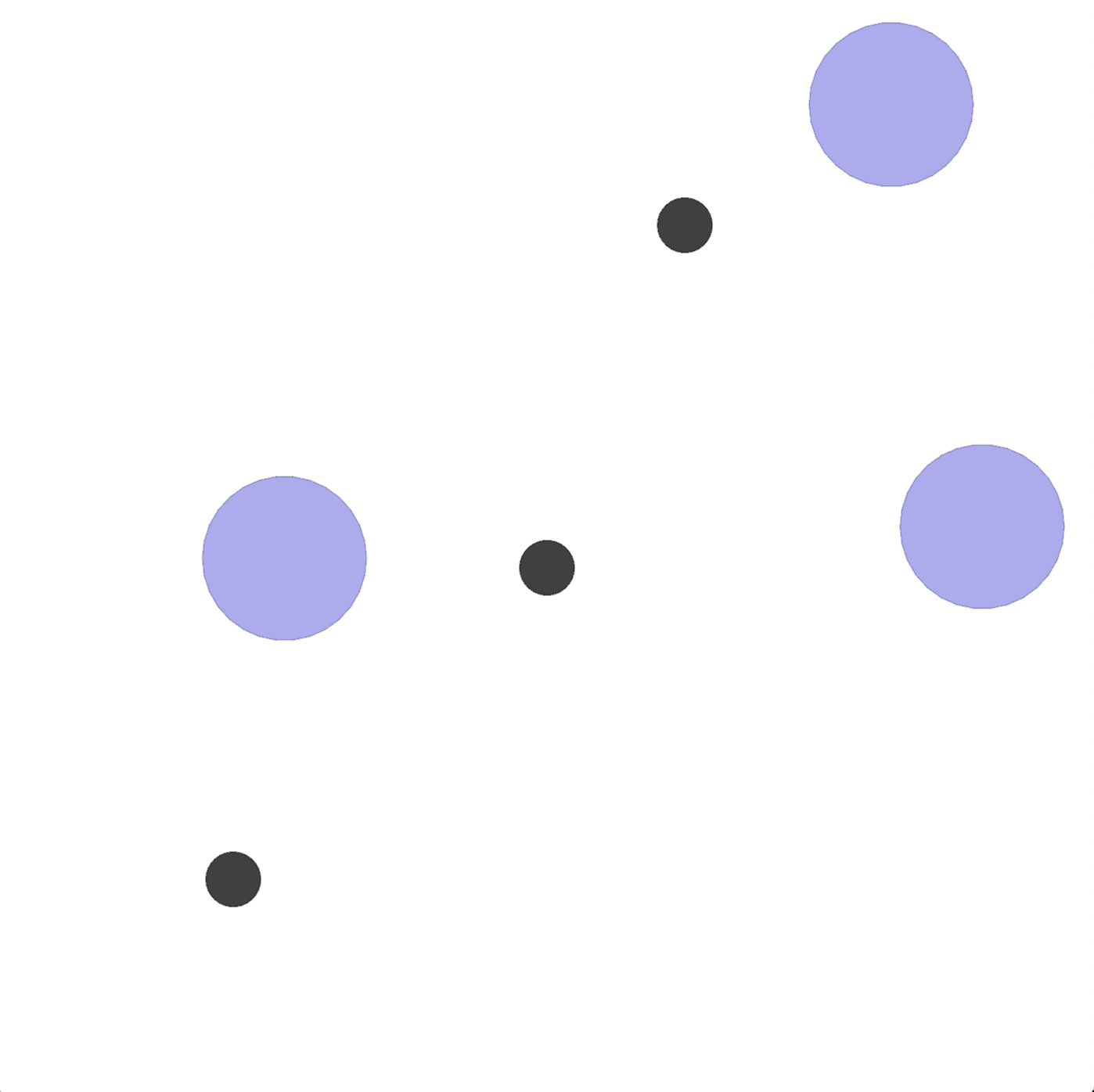}}
    \label{fig:mpe_spread}
    \caption{}
  \end{subfigure}
    \hfill
  \begin{subfigure}{0.24\linewidth}
    \centering
    \fbox{\includegraphics[width=0.9\linewidth]{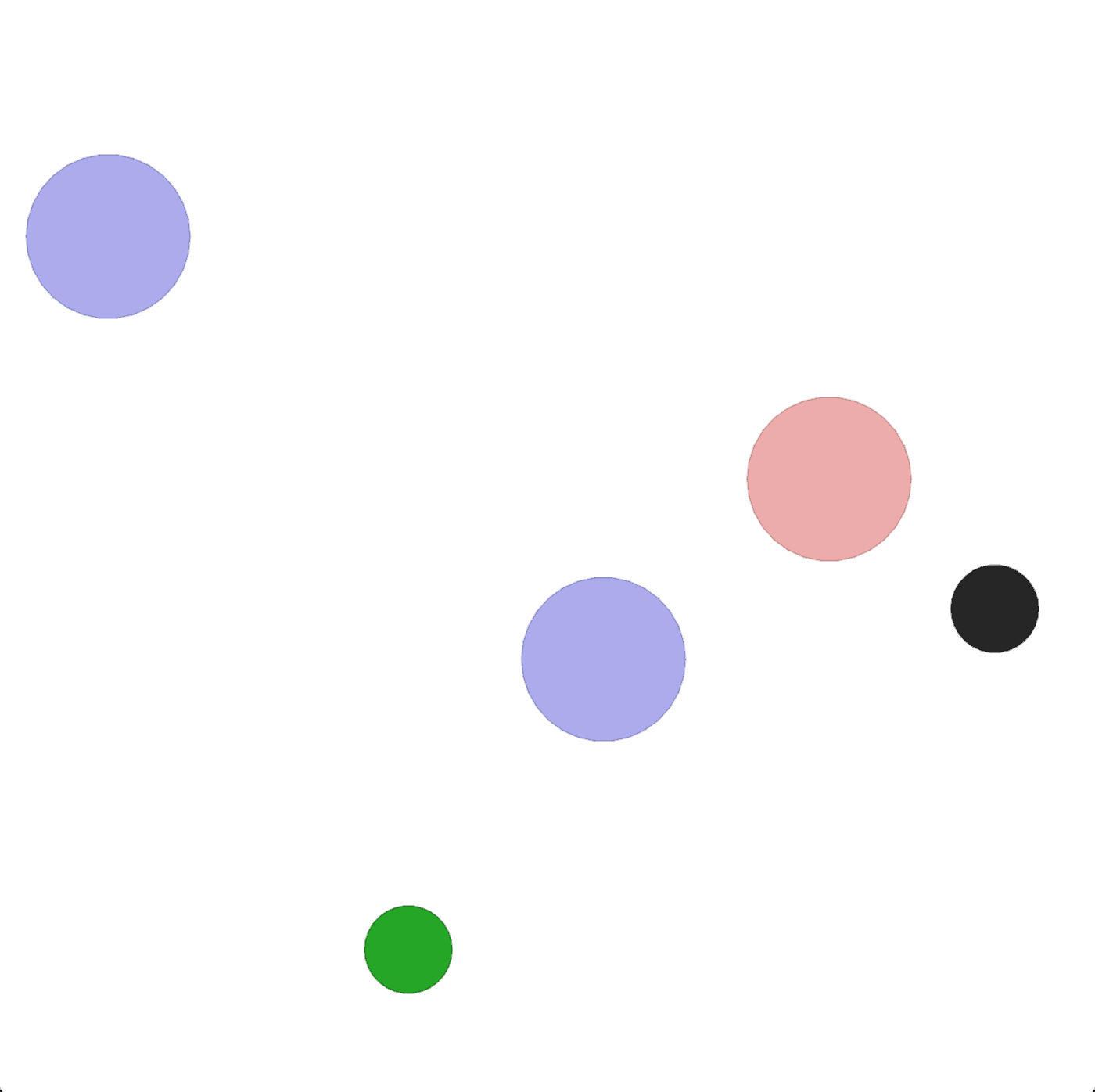}}
    \label{fig:mpe_adversary}
    \caption{}
  \end{subfigure}
  \hfill
  \begin{subfigure}{0.24\linewidth}
    \centering
    \fbox{\includegraphics[width=0.9\linewidth]{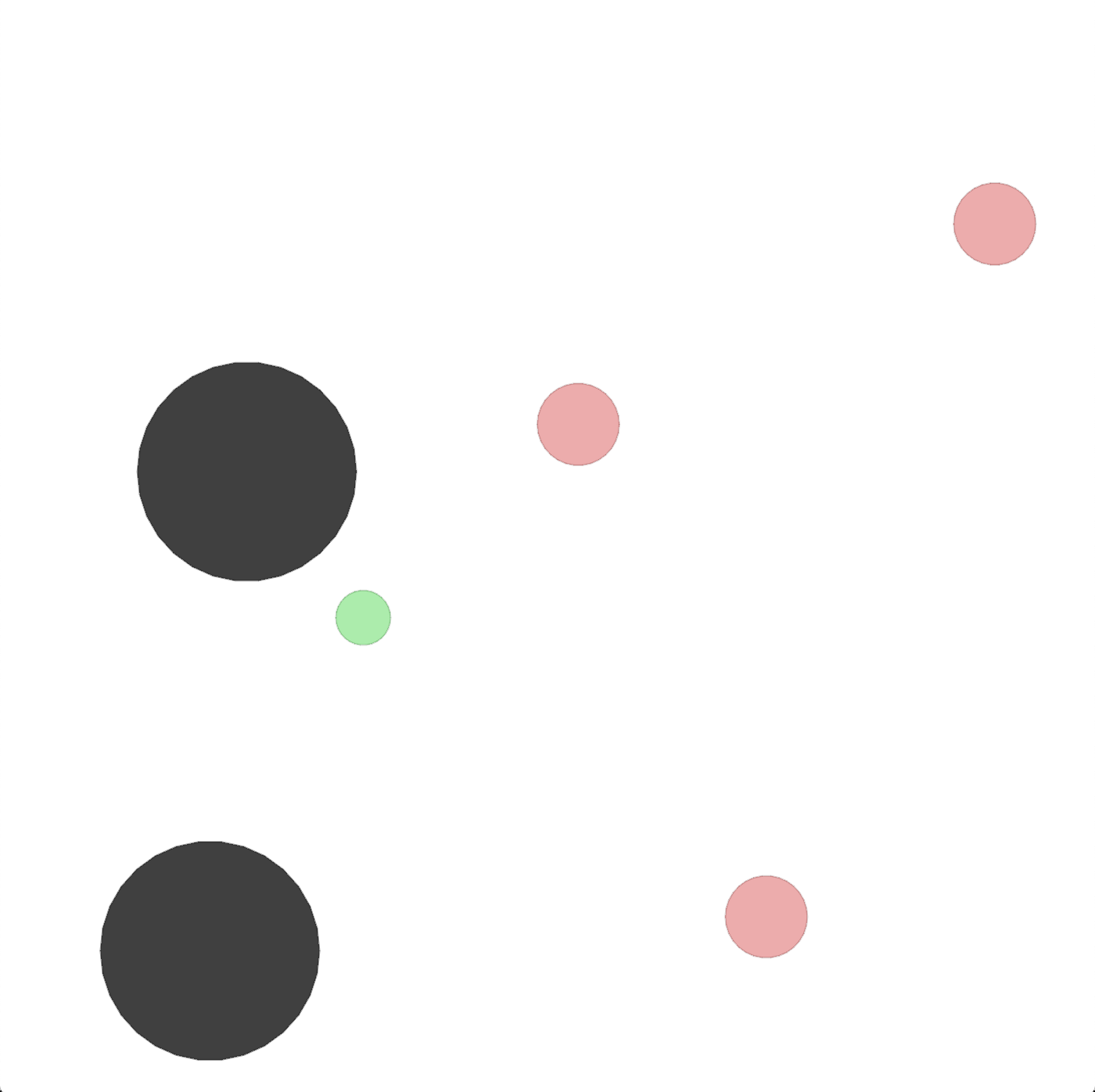}}
    \label{fig:mpe_tag}
    \caption{}
  \end{subfigure}

  \caption{Illustration of MPE tasks (a) Speaker-Listener, (b) Spread, (c) Adversary and (d) Predator-Prey.}
  \label{fig:mpe}
\end{figure}

\subsection{Multi-Agent Particle Environment \citep{mordatch2017emergence}}

This environment consists of multiple tasks involving the cooperation and competition between agents. All tasks involve particles and landmarks in a continuous two-dimensional environment. Observations consist of high-level feature vectors and agents are receiving dense reward signals. The action space among all tasks and agents is discrete and usually includes five possible actions corresponding to no movement, move right, move left, move up or move down. All experiments in this environment are executed with a maximum episode length of 25, i.e.\ episodes are terminated after 25 steps and a new episode is started. All considered tasks are visualised in \Cref{fig:mpe}.

\textbf{MPE Speaker-Listener:}
In this task, one static speaker agent has to communicate a goal landmark to a listening agent capable of moving. There are a total of three landmarks in the environment and both agents are rewarded with the negative Euclidean distance of the listener agent towards the goal landmark. The speaker agent only observes the colour of the goal landmark. Meanwhile, the listener agent receives its velocity, relative position to each landmark and the communication of the speaker agent as its observation. As actions, the speaker agent has three possible options which have to be trained to encode the goal landmark while the listener agent follows the typical five discrete movement actions of MPE tasks.

\textbf{MPE Spread:}
In this task, three agents are trained to move to three landmarks while avoiding collisions with each other. All agents receive their velocity, position, relative position to all other agents and landmarks. The action space of each agent contains five discrete movement actions. Agents are rewarded with the sum of negative minimum distances from each landmark to any agent and a additional term is added to punish collisions among agents.

\textbf{MPE Adversary:}
In this task, two cooperating agents compete with a third adversary agent. There are two landmarks out of which one is randomly selected to be the goal landmark. Cooperative agents receive their relative position to the goal as well as relative position to all other agents and landmarks as observations. However, the adversary agent observes all relative positions without receiving information about the goal landmark. All agents have five discrete movement actions. Agents are rewarded with the negative minimum distance to the goal while the cooperative agents are additionally rewarded for the distance of the adversary agent to the goal landmark. Therefore, the cooperative agents have to move to both landmarks to avoid the adversary from identifying which landmark is the goal and reaching it as well. For this competitive scenario, we use a fully cooperative version where the adversary agent is controlled by a pretrained model obtained by training all agents using the MADDPG algorithm for 25,000 episodes.

\textbf{MPE Predator-Prey:}
In this task, three cooperating predators hunt a forth agent controlling a faster prey. Two landmarks are placed in the environment as obstacles. All agents receive their own velocity and position as well as relative positions to all other landmarks and agents as observations. Predator agents also observe the velocity of the prey. All agents choose among five movement actions. The agent controlling the prey is punished for any collisions with predators as well as for leaving the observable environment area (to prevent it from simply running away without needing to learn to evade). Predator agents are collectively rewarded for collisions with the prey. We employ a fully cooperative version of this task with a pretrained prey agent. Just as for the Adversary task, the model for the prey is obtained by training all agents using the MADDPG algorithm for 25,000 episodes.

\subsection{StarCraft Multi-Agent Challenge \citep{samvelyan19smac}}

\begin{figure}[t]
    \begin{subfigure}{0.26\linewidth}
        \centering
        \includegraphics[width=0.9\textwidth]{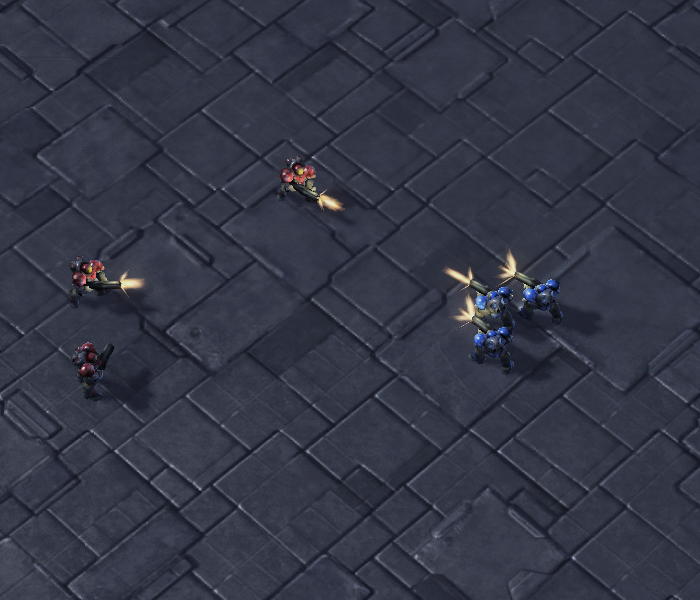}
        \label{fig:smac_3m}
        \caption{}
    \end{subfigure}
    \hfill
    \begin{subfigure}{0.38\linewidth}
        \centering
        \includegraphics[width=0.9\textwidth]{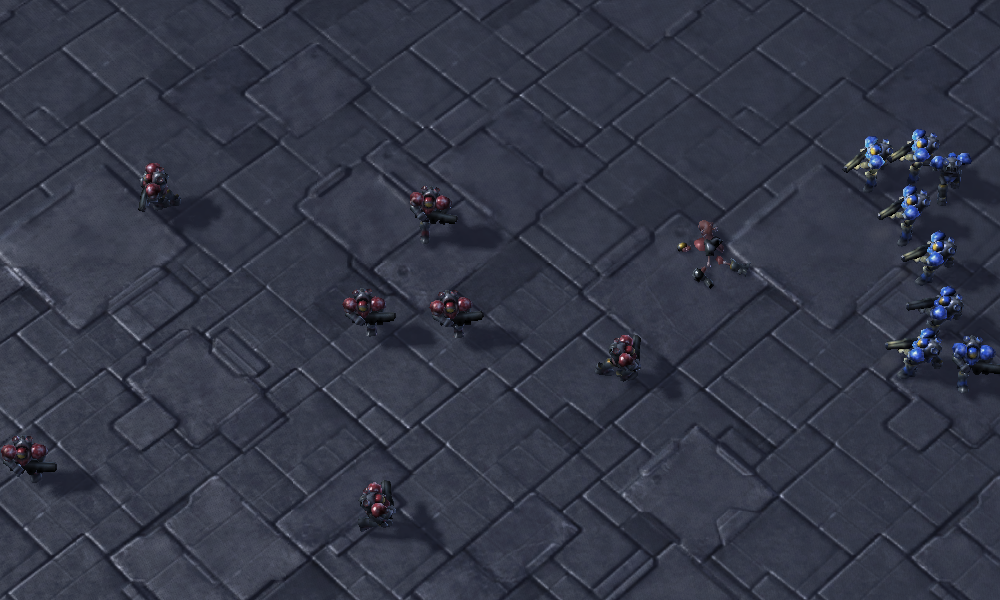}
        \label{fig:smac_8m}
        \caption{}
    \end{subfigure}
    \hfill
    \begin{subfigure}{0.32\linewidth}
        \centering
        \includegraphics[width=0.9\textwidth]{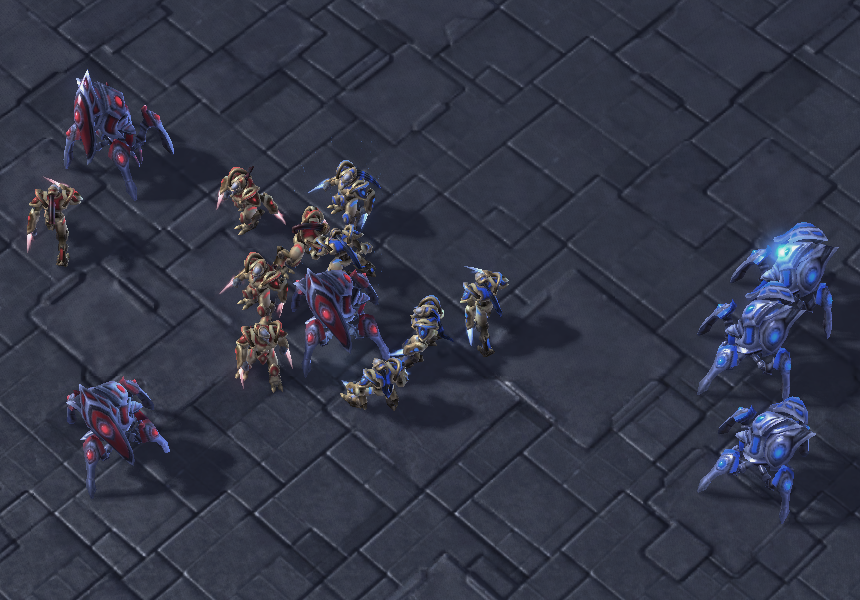}
        \label{fig:smac_3s5z}
        \caption{}
    \end{subfigure}
    
    \begin{subfigure}{0.473\linewidth}
        \centering
        \includegraphics[width=0.9\textwidth]{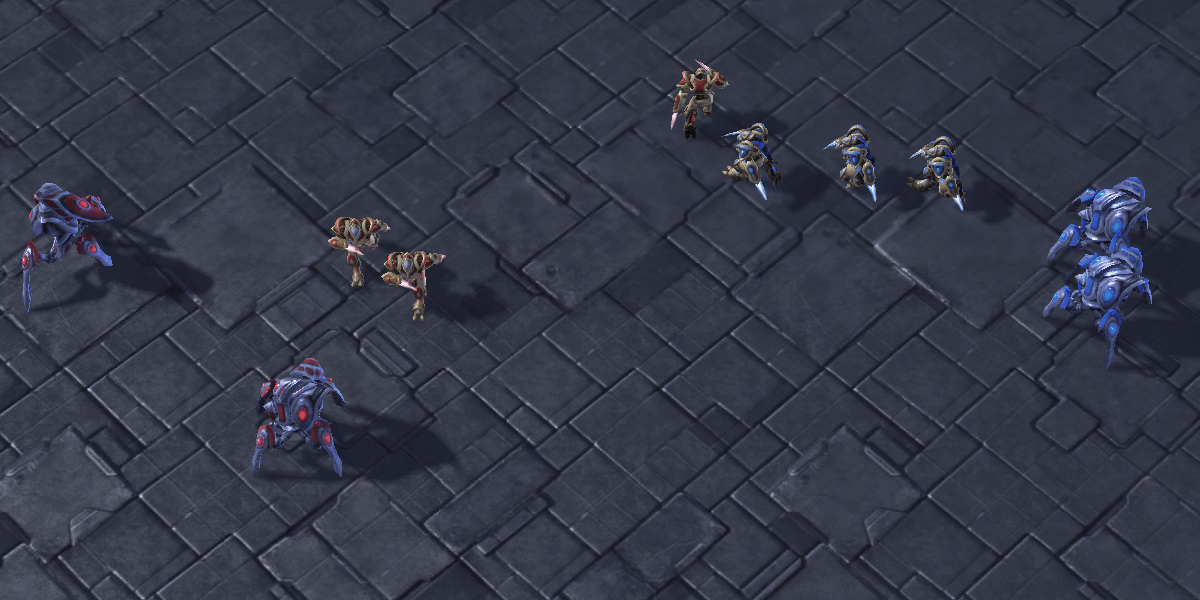}
        \label{fig:smac_2s3z}
        \caption{}
    \end{subfigure}
    \hfill
    \begin{subfigure}{0.505\linewidth}
        \centering
        \includegraphics[width=0.9\textwidth]{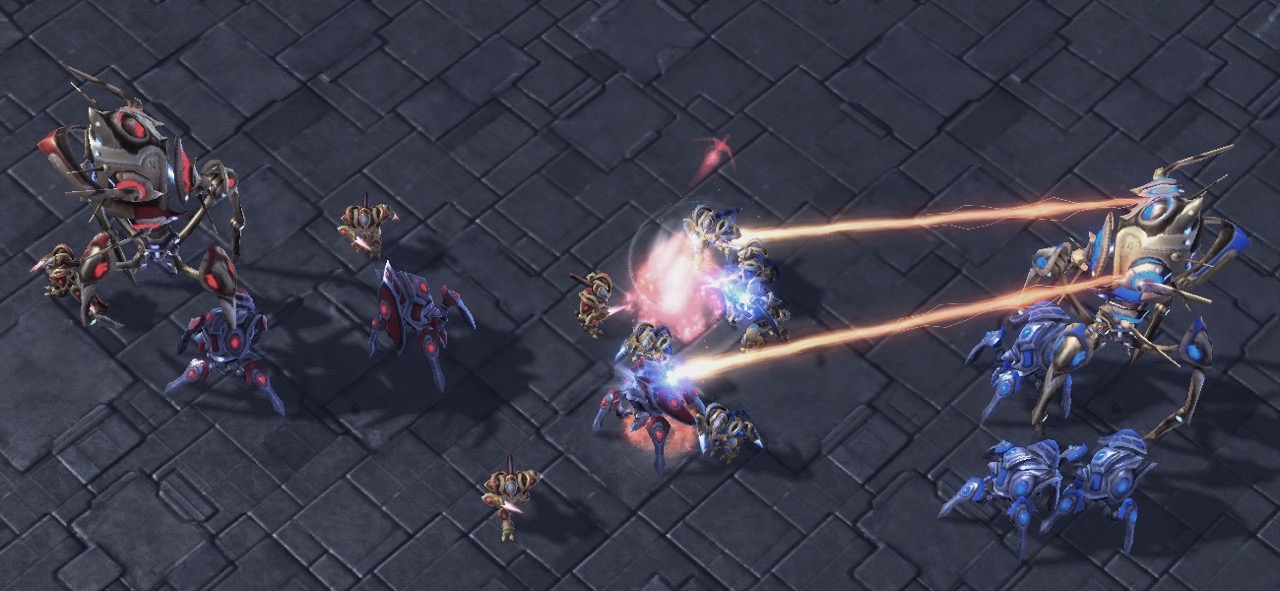}
        \label{fig:smac_1c3s5z}
        \caption{}
    \end{subfigure}
    \caption{Examples of SMAC tasks with various team configurations and unit types.}
    \label{fig:smac}
\end{figure}

The StarCraft Multi-Agent Challenge is a set of fully cooperative, partially observable multi-agent tasks. This environment implements a variety of micromanagement tasks based on the popular real-time strategy game StarCraft II\footnote{StarCraft II is a trademark of Blizzard Entertainment\textsuperscript{TM}.} and makes use of the StarCraft II Learning Environment (SC2LE) \citep{vinyals2017starcraft}. Each task is a specific combat scenario in which a team of agents, each agent controlling an individual unit, battles against an army controlled by the centralised built-in AI of the StarCraft game. These tasks require agents to learn precise sequences of actions to enable skills like \textit{kiting} as well as coordinate their actions to focus their attention on specific opposing units. All considered tasks are symmetric in their structure, i.e.\ both armies consist of the same units. \Cref{fig:smac} visualises each considered task in this environment.

\textbf{SMAC 2s\_vs\_1sc:}
In this scenario, agents control two stalker units and defeat the enemy team consisting of a single, game-controlled spine crawler.

\textbf{SMAC 3s5z:}
In this symmetric scenario, each team controls three stalkers and five zerglings for a total of eight agents.

\textbf{SMAC MMM2:}
In this symmetric scenario, each team controls seven marines, two marauders, and one medivac unit. The medivac unit assists other team members by healing them instead of inflicting damage to the enemy team.

\textbf{SMAC corridor:}
In this asymmetric scenario, agents control six zealots fighting an enemy team of 24 zerglings controlled by the game. This tasks requires agents to make effective use of terrain features and employ certain game-specific tricks to win.

\textbf{SMAC 3s\_vs\_5z:}
Finally, in this scenario a team of three stalkers is controlled by agents to fight against a team of five game-controlled zerglings.

\subsection{Level-Based Foraging \citep{albrecht2013game}}

\begin{figure}
    \begin{subfigure}{0.24\linewidth}
        \centering
        \includegraphics[width=0.9\linewidth]{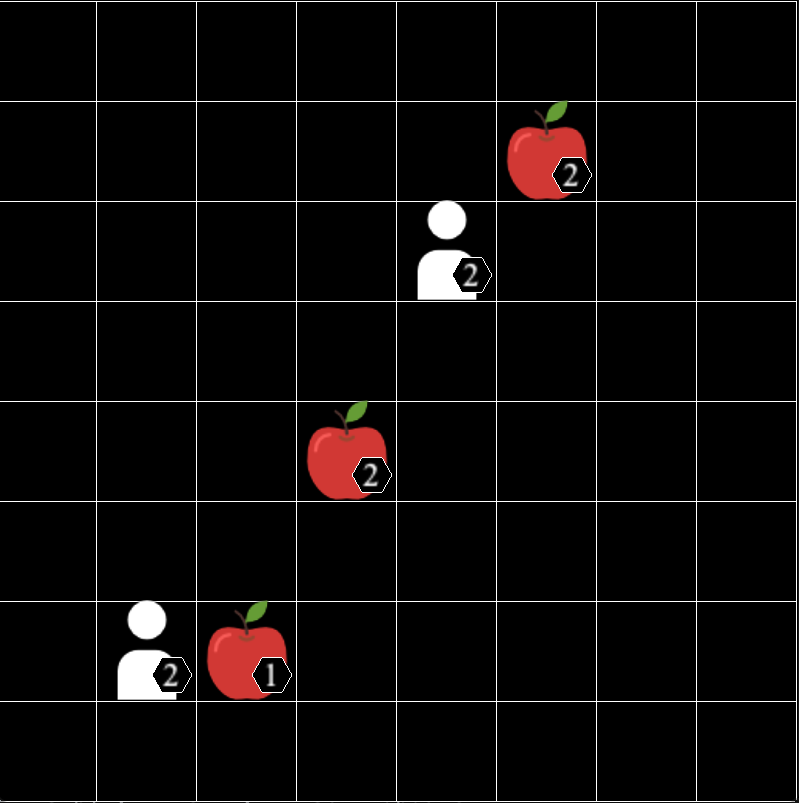}
        \caption{}
        \label{fig:lbf_coop1}
    \end{subfigure}
    \hfill
    \begin{subfigure}{0.24\linewidth}
        \centering
        \includegraphics[width=0.9\linewidth]{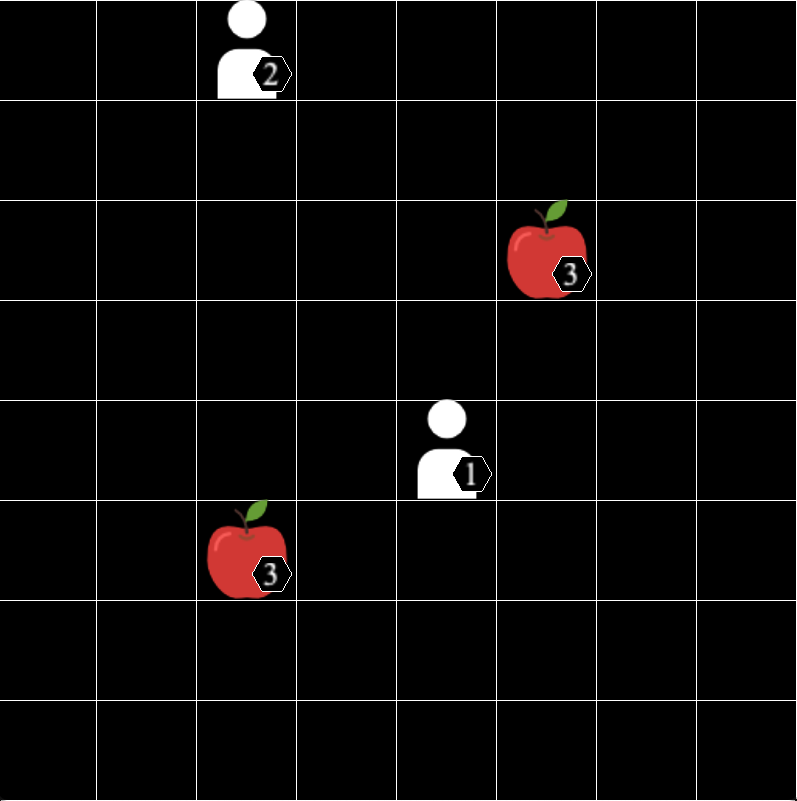}
        \caption{}
        \label{fig:lbf_coop2}
    \end{subfigure}
    \hfill
    \begin{subfigure}{0.24\linewidth}
        \centering
        \includegraphics[width=0.9\linewidth]{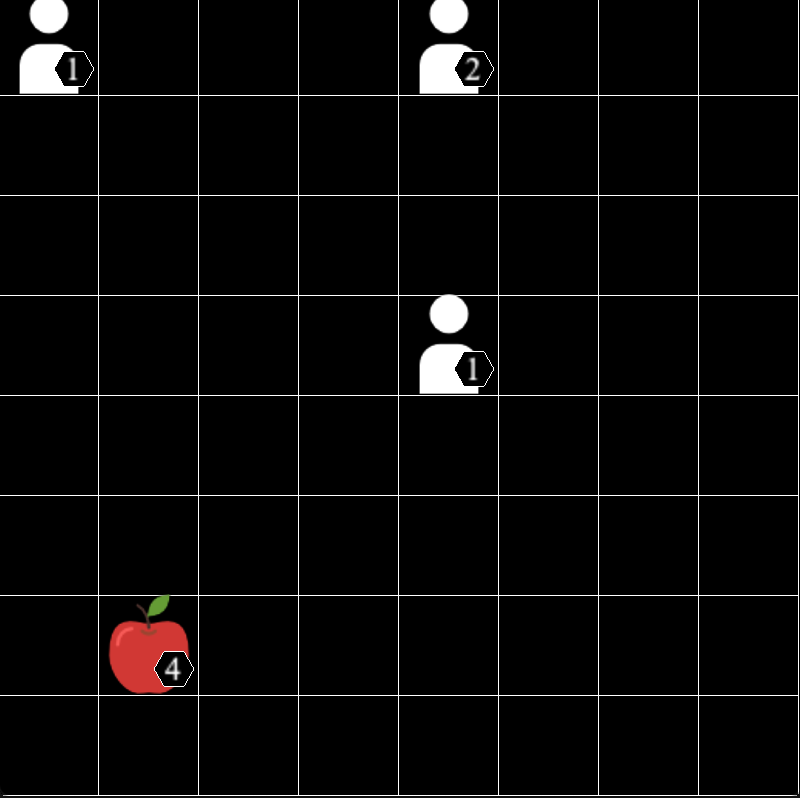}
        \caption{}
        \label{fig:lbf_mixed1}
    \end{subfigure}
    \hfill
    \begin{subfigure}{0.24\linewidth}
        \centering
        \includegraphics[width=0.9\linewidth]{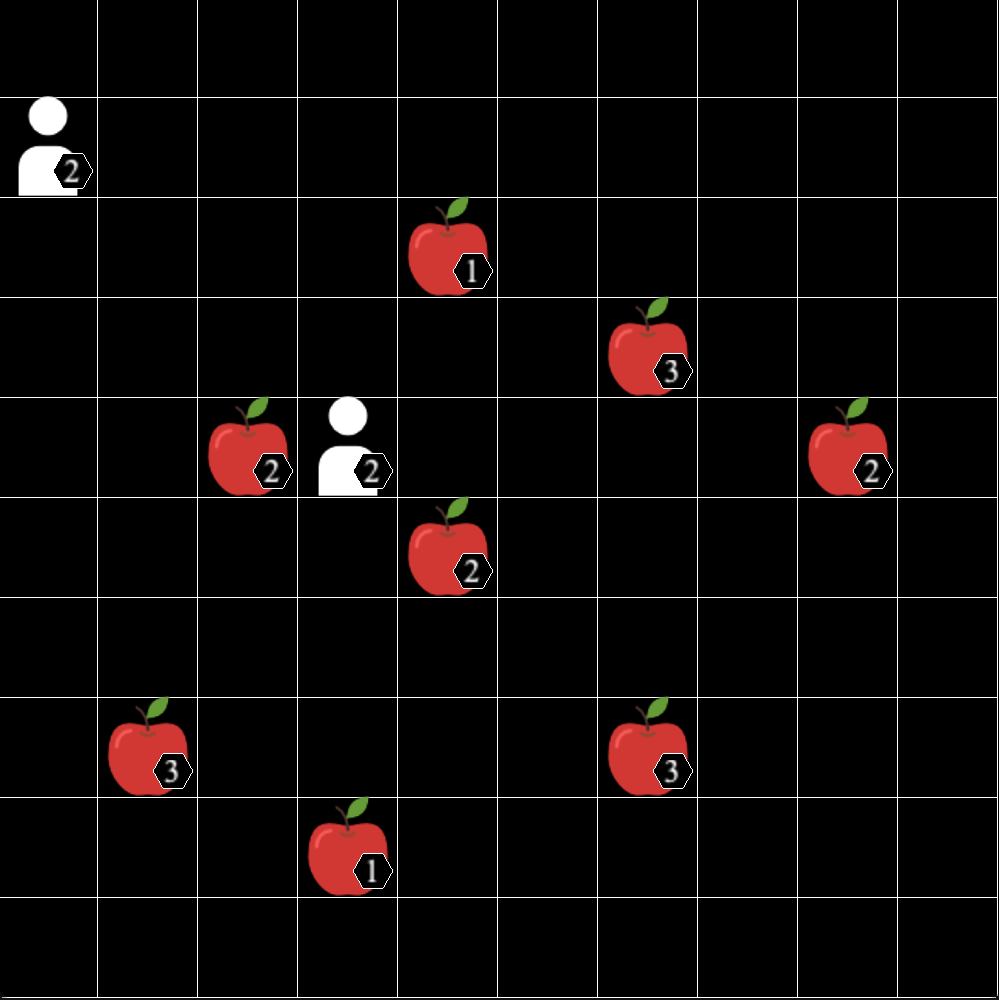}
        \caption{}
        \label{fig:lbf_mixed2}
    \end{subfigure}
    \caption{Examples of LBF tasks with variable agents, grid-sizes and food items.}
    \label{fig:lbf}
\end{figure}

The Level-Based Foraging environment consists of tasks focusing on the coordination of involved agents. The task for each agent is to navigate the grid-world map and collect items. Each agent and item is assigned a level and items are randomly scattered in the environment. In order to collect an item, agents have to choose a certain action next to the item. However, such collection is only successful if the sum of involved agents' levels is equal or greater than the item level. Agents receive reward equal to the level of the collected item. \Cref{fig:lbf} shows the tasks used for our experiments. Unless otherwise specified, every agent can observe the whole map, including the positions and levels of all the entities and can choose to act by moving in one of four directions or attempt to pick up an item.

The tasks that were selected are denoted first by the grid-world size (e.g. $15\times15$ means a 15 by 15 grid-world). Then, the number of agents is shown (e.g. ``4p'' means four agents/players), and the number of items scattered in the grid (e.g. ``3f'' means three food items). We also have some special flags, the ``2s'' which denotes partial observability with a range of two squares. In those tasks, agents can only observe items or other agents as long as they are located in a square of size 5x5 centred around them. Finally, the flag ``c'' means a cooperative-only variant were all items can only be picked up if all agents in the level attempt to load it simultaneously. 

There are many aspects of LBF that are interesting. An increased number of agents and grid-world size naturally makes the problem harder, requiring agents to coordinate to cover a larger area. In addition, partial observability tasks could benefit from agents modelling other agents since it can improve coordination. A tasks that we were unable to get any algorithm to learn is the cooperative-only variant with three or more agents. In that case, many agents are required to coordinate to be able to gather any reward, making the task very hard to solve given the sparsity of the rewards. This last task could on its own provide an interesting challenge for research on multi-agent intrinsic exploration.  

Notably, LBF is not only designed for a cooperative reward. In other settings, it can work as a mixed competitive/cooperative environment, where agents must switch between cooperating (for gathering items that they can only pick with others) and competing for items that they can load on their own (without sharing the item reward).

\subsection{Multi-Robot Warehouse}

\begin{figure}
    \begin{subfigure}{0.5\linewidth}
        \centering
        \includegraphics[width=0.5\linewidth]{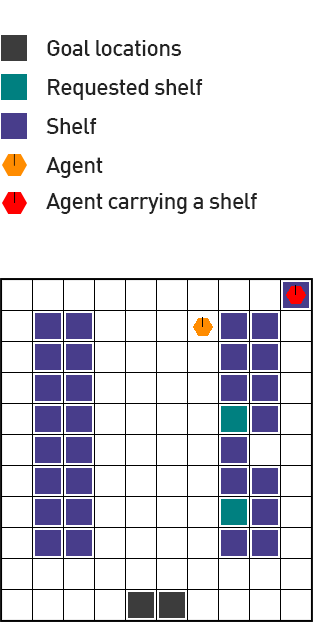}
        \caption{}
        \label{fig:rware_tiny}
    \end{subfigure}
    \begin{subfigure}{0.5\linewidth}
        \centering
        \includegraphics[width=0.5\linewidth]{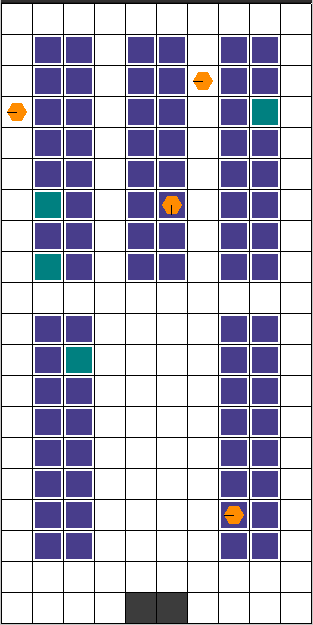}
        \caption{}
        \label{fig:rware_small}
    \end{subfigure}
    \caption{Illustrations of (a) Tiny 2p and (b) Small 4p.}
    \label{fig:rware}
\end{figure}

The multi-robot warehouse environment is a set of collaborative, partially observable multi-agent tasks simulating a warehouse operated by robots. Each agent controls a single robot aiming to collect requested shelves. At all times, $N$ shelves are requested and each timestep a request is delivered to the goal location, a new (currently unrequested) shelf is uniformly sampled and added to the list of requests. Agents observe a $3\times3$ grid including information about potentially close agents, given by their location and rotation, as well as information on surrounding shelves and a list of requests. The action space is discrete and contains of four actions corresponding to turning left or right, moving forward and loading or unloading a shelf. %
Agents are only rewarded whenever an agent is delivering a requested shelf to a goal position. Therefore, a very specific and long sequence of actions is required to receive any non-zero rewards, making this environment very sparsely rewarded.
We use multi-robot warehouse tasks with warehouses of varying size and number of agents $N$ (which is also equal to the number of requested shelves). \Cref{fig:rware} illustrates the tiny and small warehouses with $2$ agents.

For this environment, we have defined three tasks. The ``tiny'' map is a grid-world of 11 by 11 squares and the ``'small'' is 11 by 20 squares. The ``2p'' and ``4p'' signify two and four robots (agents) respectively. The larger map is considerably harder given the increased sparsity of the rewards.

There are many challenges in the multi-robot warehouse domain as well. The tasks we test in the benchmark paper are limited because of the algorithms' inability to learn and decompose the reward under such sparse reward settings. However, if this challenge is surpassed, either by using a non-cooperative setting or a new algorithm, then the domain offers additional challenges by requiring algorithms to scale to a large number of agents, communicate intentions to increase efficiency and more. This environment is based on a real-world problem, and scaling to scenarios with large maps and hundreds of agents still requires a considerable research effort.

\section{Computational Cost}

\begin{figure}[H]
    \centering
    \begin{subfigure}{0.19\textwidth}
        \includegraphics[width=1.0\textwidth]{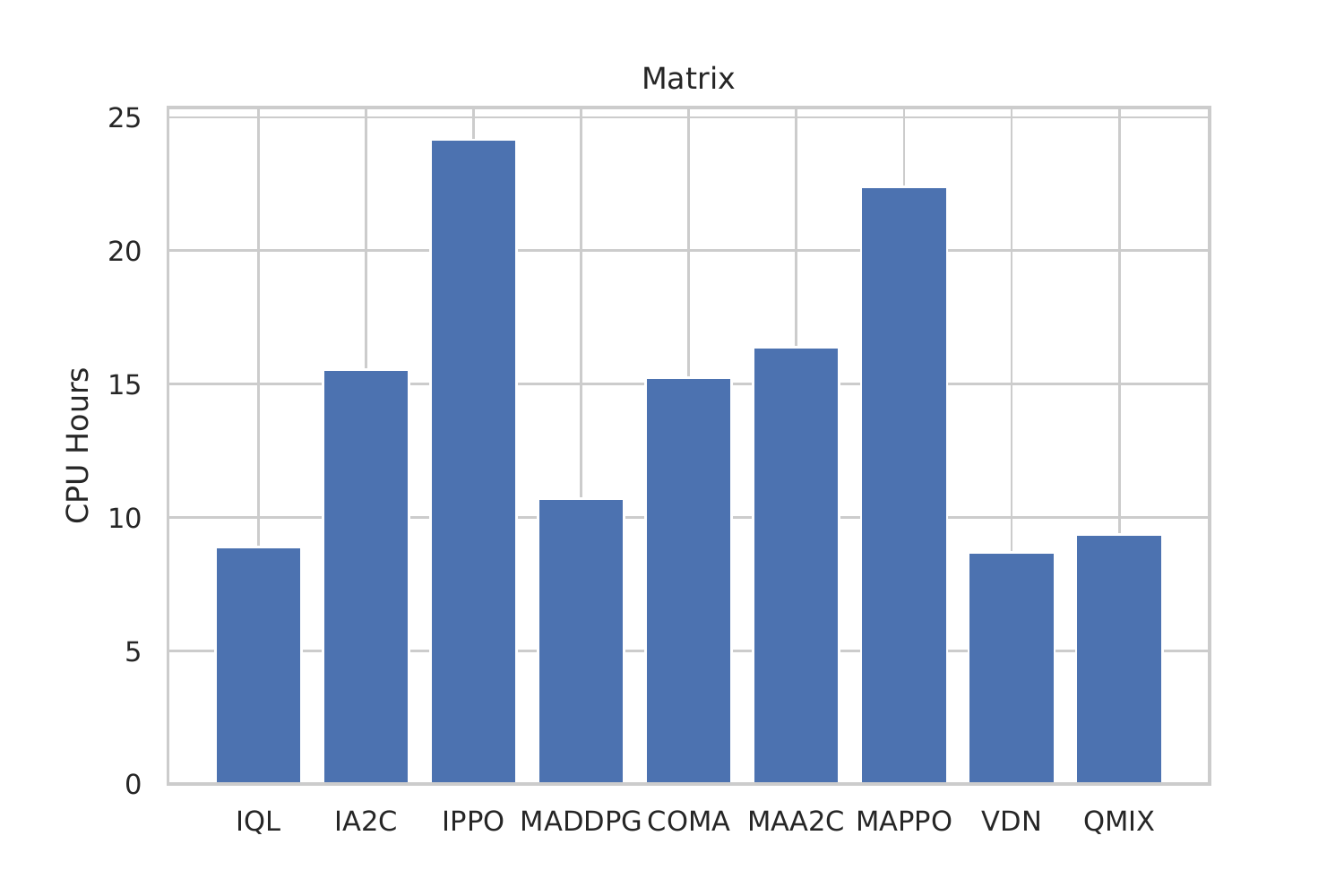}
    \end{subfigure}
     \hfill
    \begin{subfigure}{0.19\textwidth}
        \includegraphics[width=1.0\textwidth]{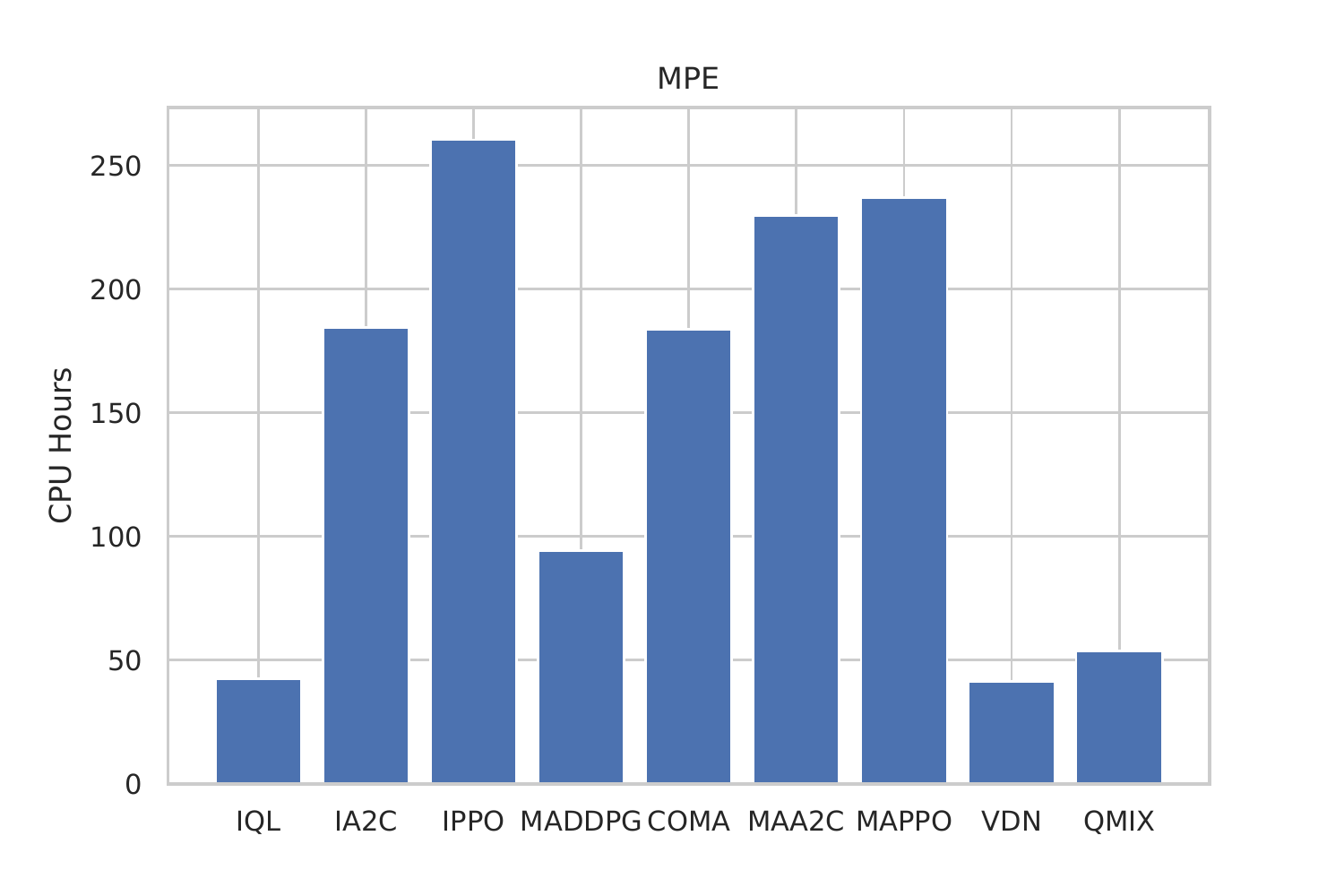}
    \end{subfigure}
     \hfill
    \begin{subfigure}{0.19\textwidth}
        \includegraphics[width=1.0\textwidth]{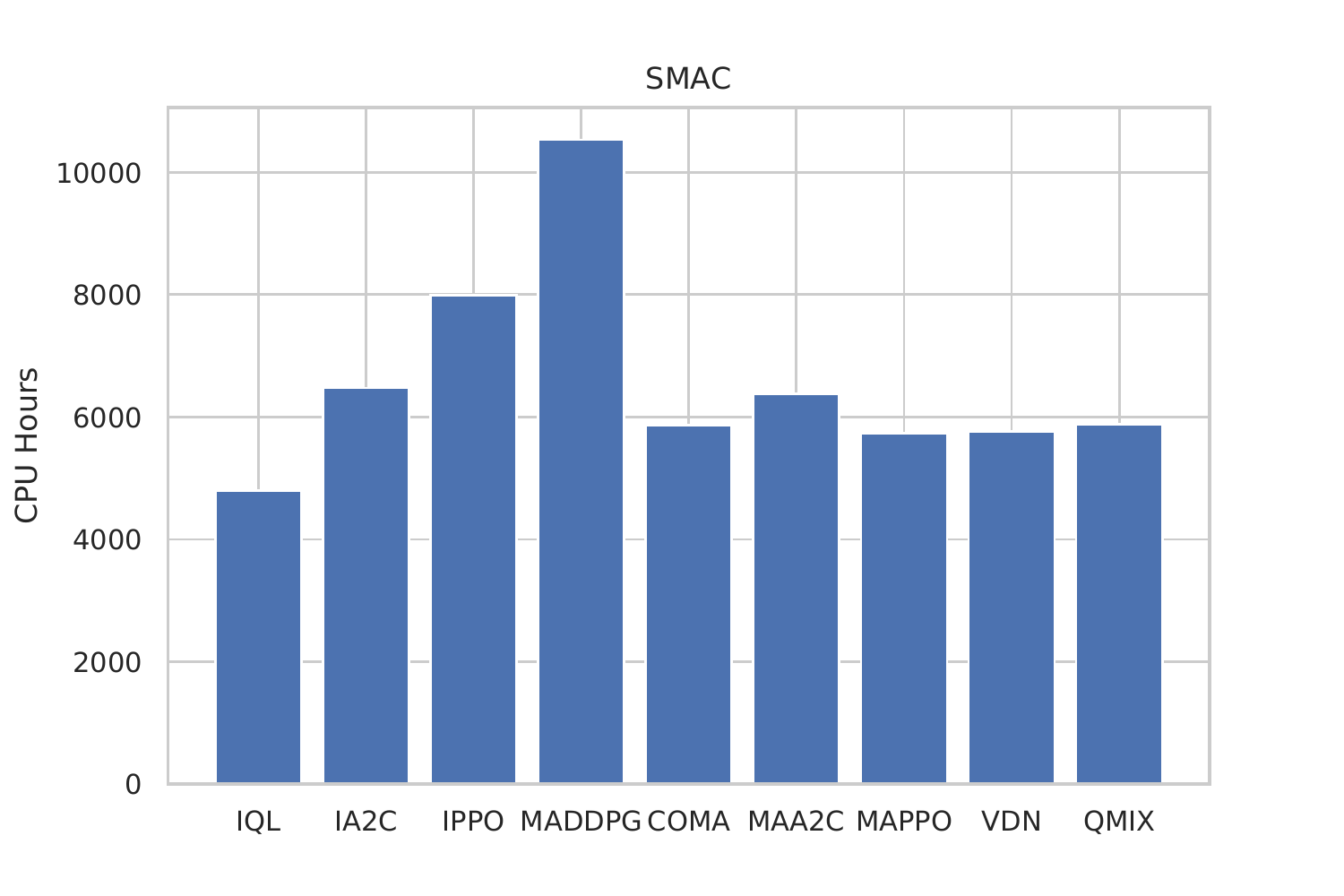}
    \end{subfigure}
     \hfill
    \begin{subfigure}{0.19\textwidth}
          \includegraphics[width=1.0\textwidth]{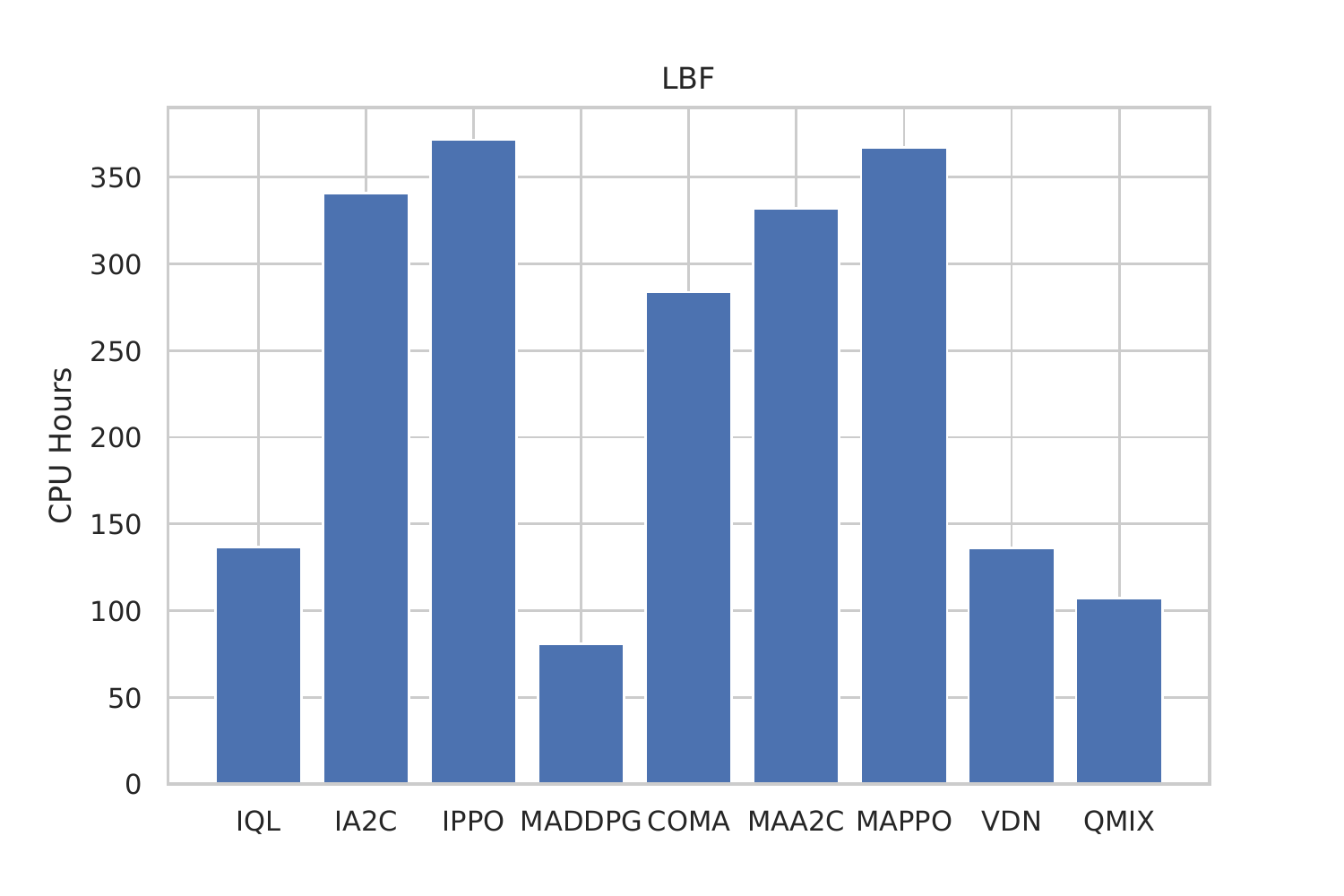}
    \end{subfigure}
    \hfill
    \begin{subfigure}{0.19\textwidth}
          \includegraphics[width=1.0\textwidth]{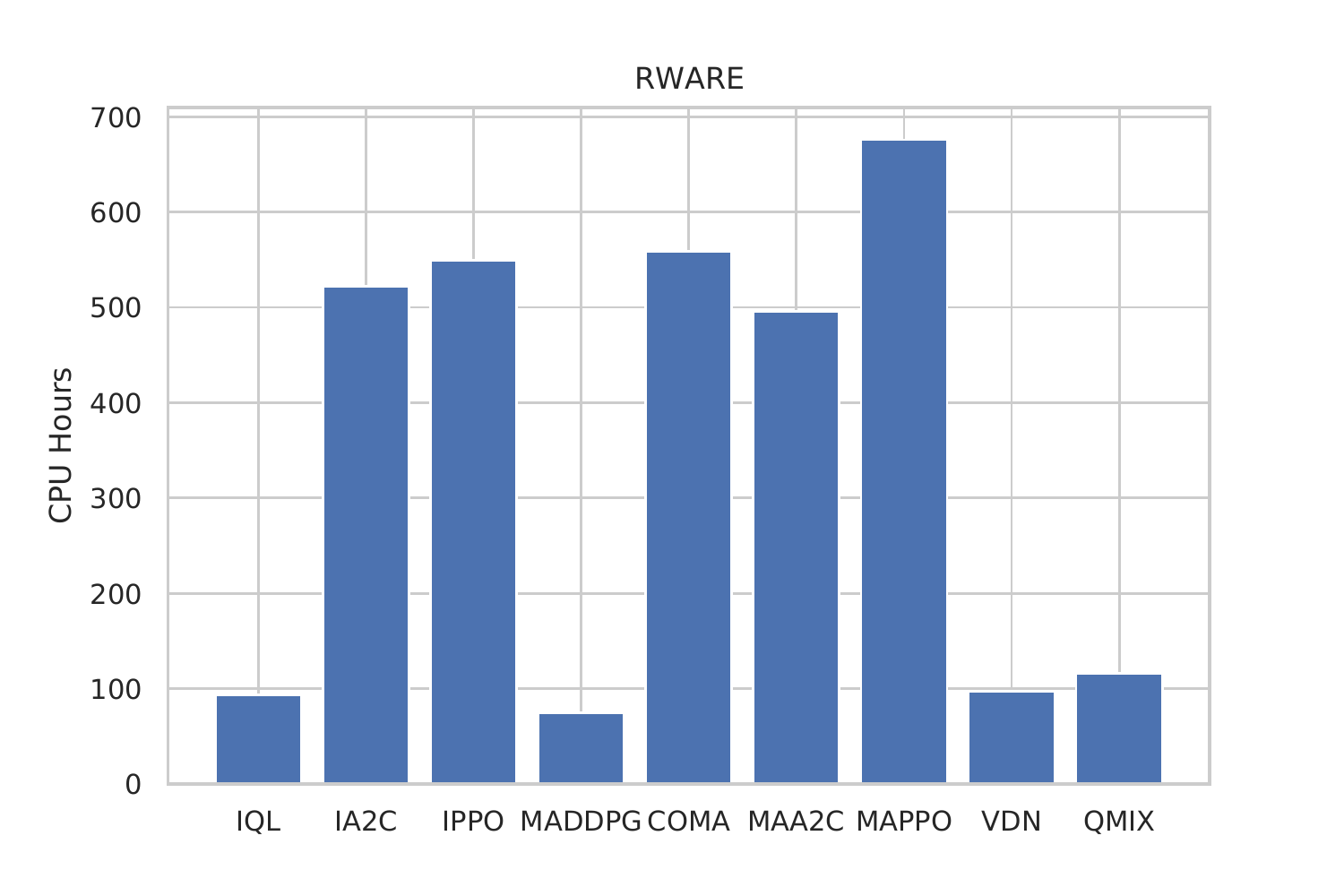}
    \end{subfigure}
    
    \begin{subfigure}{0.19\textwidth}
        \includegraphics[width=1.0\textwidth]{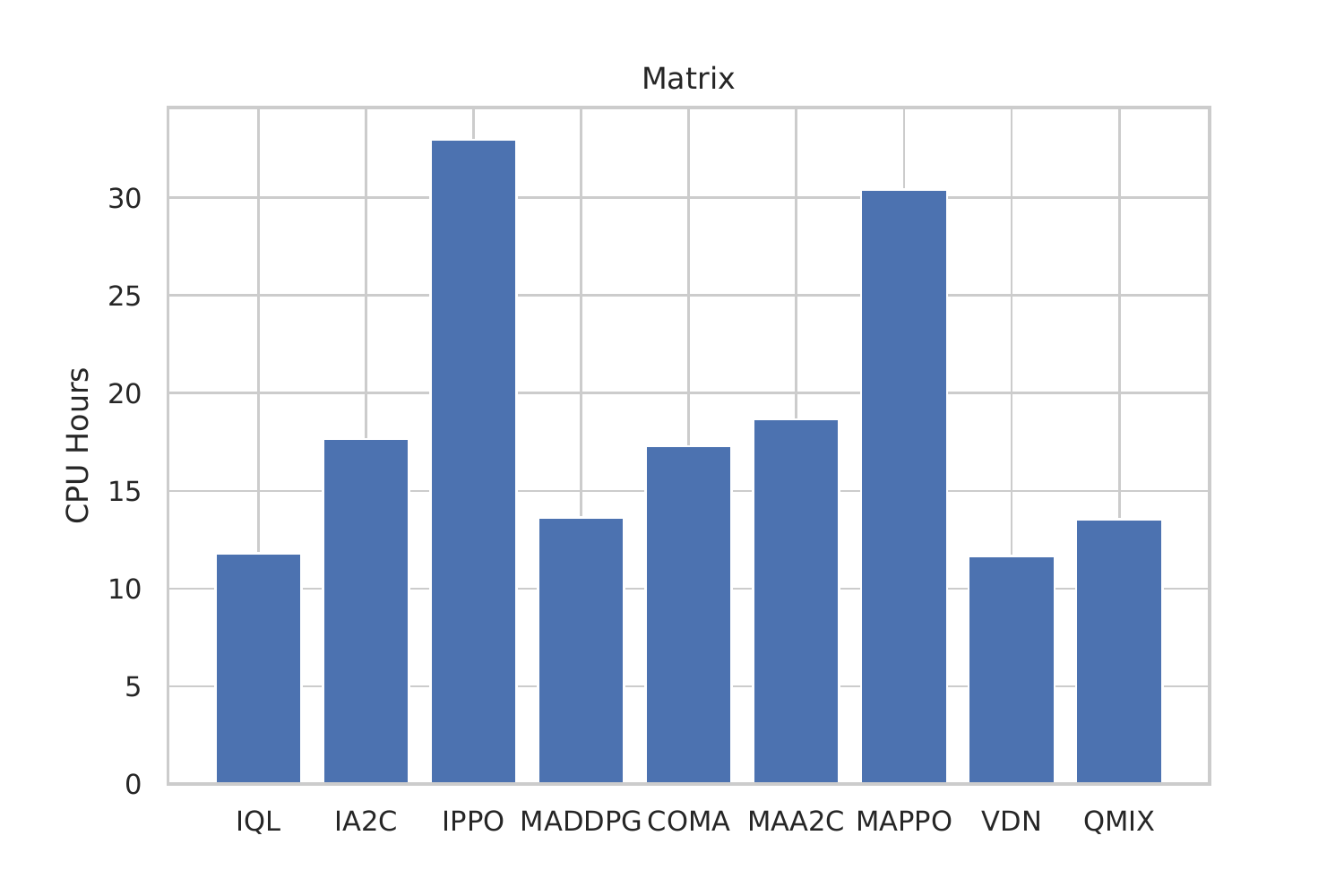}
    \end{subfigure}
     \hfill
    \begin{subfigure}{0.19\textwidth}
        \includegraphics[width=1.0\textwidth]{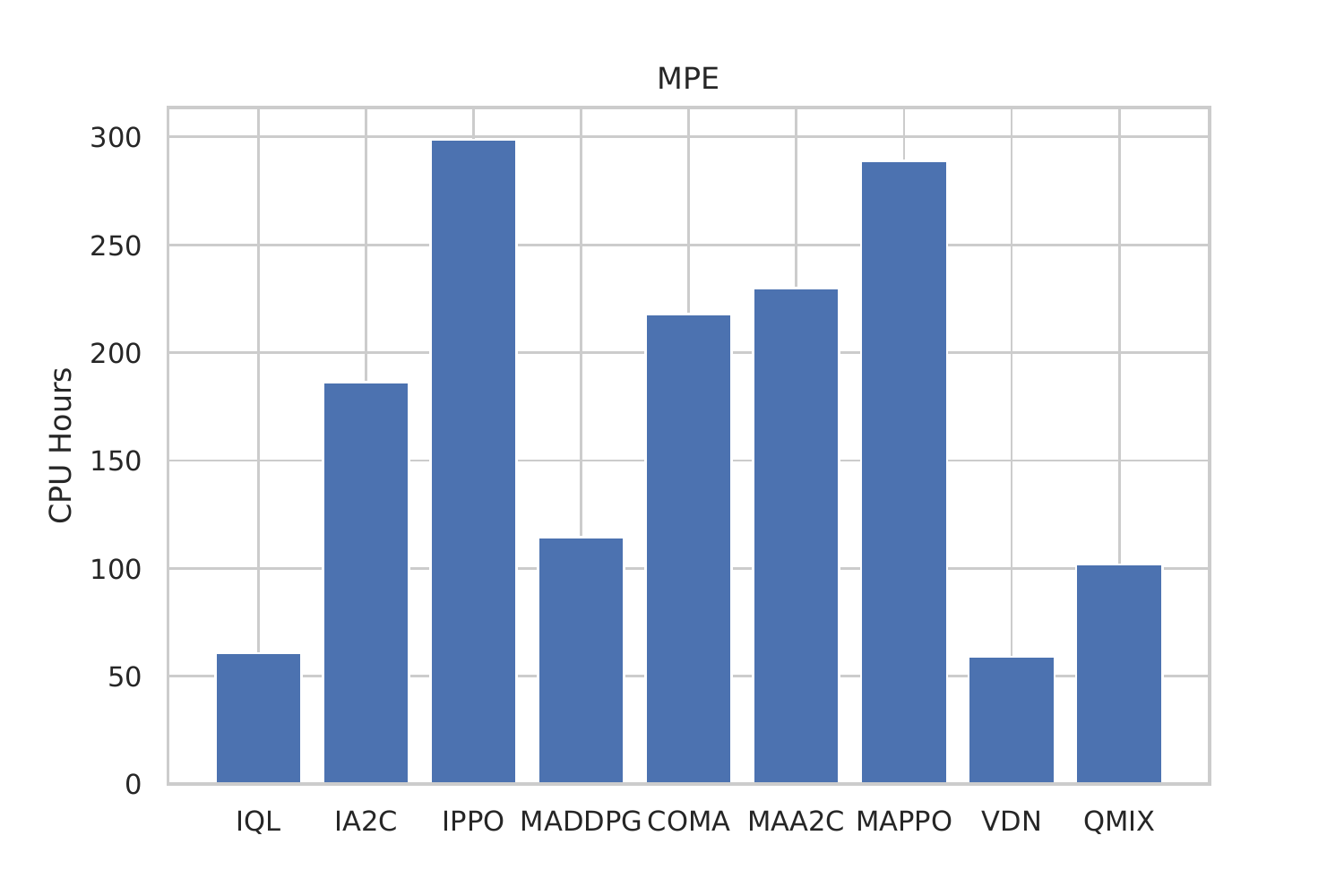}
    \end{subfigure}
     \hfill
    \begin{subfigure}{0.19\textwidth}
        \includegraphics[width=1.0\textwidth]{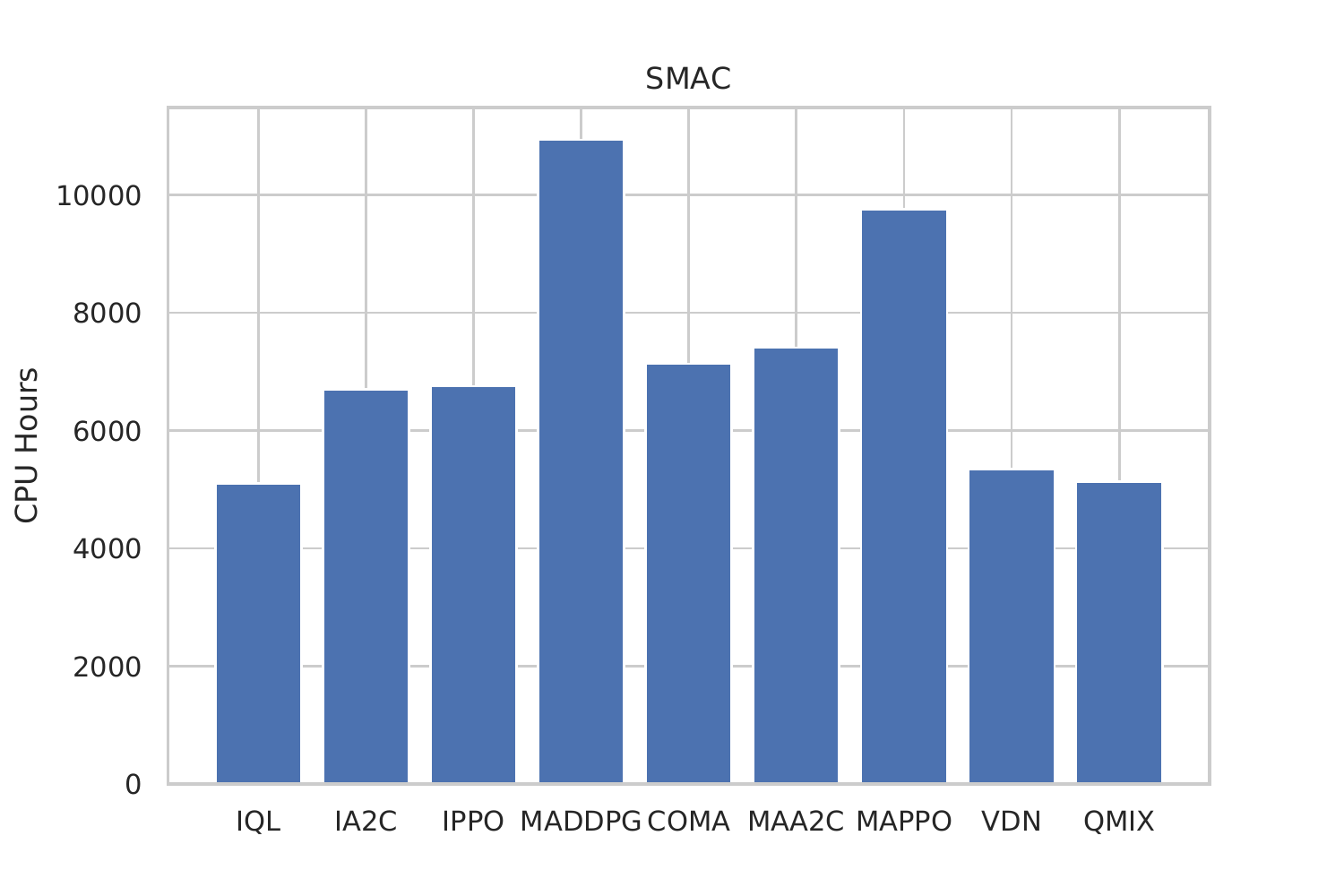}
    \end{subfigure}
     \hfill
    \begin{subfigure}{0.19\textwidth}
          \includegraphics[width=1.0\textwidth]{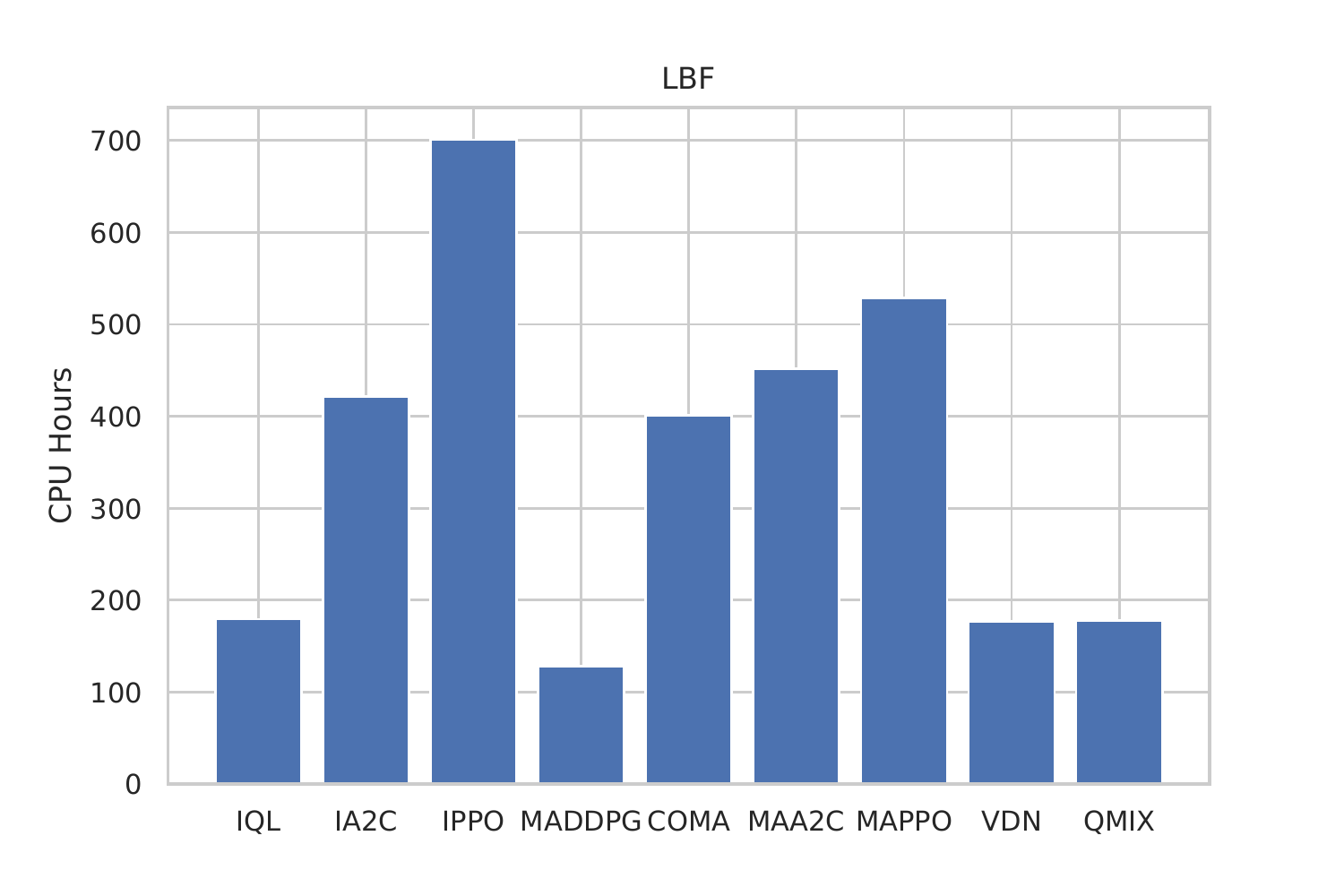}
    \end{subfigure}
    \hfill
    \begin{subfigure}{0.19\textwidth}
          \includegraphics[width=1.0\textwidth]{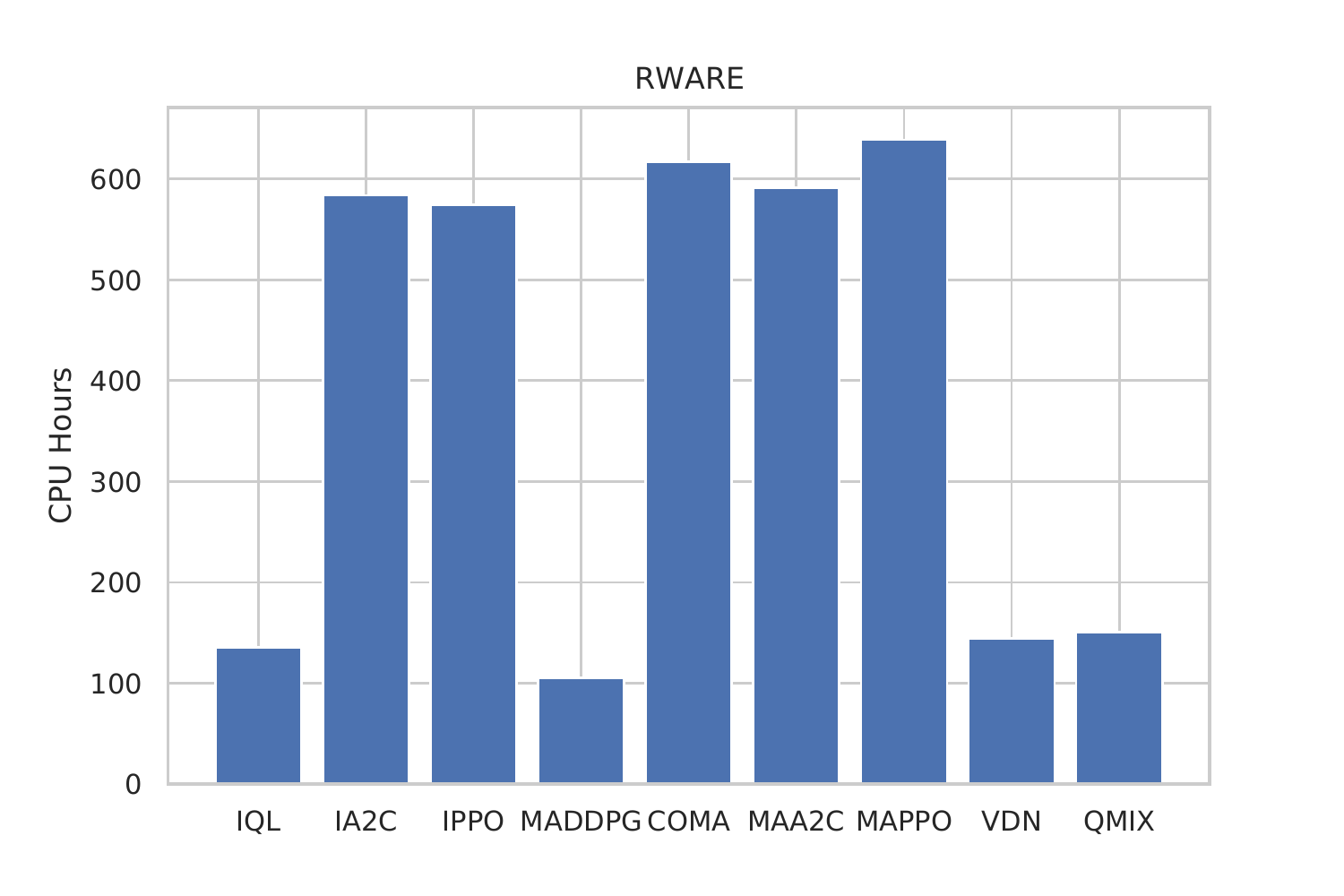}
    \end{subfigure}
\caption{\label{fig:cpu_hours} CPU hours required to execute the experiments for each algorithm and environment with  (top row) and without (bottom row) parameter sharing.}

\end{figure}

Approximately 138,916 CPU hours were spent for executing the experiments presented in the paper without considering the CPU hours required for the hyperparameter search. \Cref{fig:cpu_hours} presents the cumulative CPU hours required to train each algorithm in each environments (summed over the different tasks and seeds) with and without parameter sharing using the best identified hyperparameter configurations reported in \Cref{sec:hyper}. We observe that the computational cost of running experiments in SMAC is significantly higher compared to any other environment. Finally, the CPU hours required for training the algorithms without sharing is slightly higher compared to training with parameter sharing.

\newpage
\section{Additional Results}
\label{sec:add_results}
\Cref{tab:mean_returns} presents the average returns over training of the nine algorithms in the 25 different tasks with parameter sharing. Tables \ref{tab:max_returns_ns} and \ref{tab:mean_returns_ns} present the maximum and the average returns respectively of the nine algorithms in the 25 different tasks without  parameter sharing.
\begin{table}[H]
\centering
\caption{\label{tab:mean_returns} Average returns and 95\% confidence interval over five seeds for all nine algorithms with parameter sharing in all 25 tasks.}
\resizebox{\linewidth}{!}{
\robustify\bf
\begin{tabular}{p{1em} l S S  S  S  S  S  S S S S}
\toprule
& \textbf{Tasks \textbackslash Algs. }  &             {IQL} &           {IA2C} &             {IPPO} &              {MADDPG} &              {COMA} &              {MAA2C} &               {MAPPO} &             {VDN} &           {QMIX} \\ \\
\hline \\
\multirow{6}{*}{\rotatebox[origin=c]{90}{Matrix Games}}
& Climbing    &   134.65(63) &  169.34(109) &  170.70(177) &  156.45(809) &  177.31(4952) &  167.89(436) &  170.76(179) &   125.50(54) &   125.50(54) \\
& Penalty k=0   &  245.64(170) &  244.27(113) &   247.44(59) &   246.39(45) &    245.63(64) &   244.78(87) &    247.69(9) &  239.80(293) &  243.76(236) \\
& Penalty k=-25  &   44.65(267) &    48.29(30) &    48.44(21) &     48.59(5) &     48.33(20) &    48.45(12) &    48.46(22) &   43.32(598) &   45.99(327) \\
& Penalty k=-50  &   39.70(515) &    46.56(57) &    46.79(48) &    47.22(17) &     46.61(44) &    46.81(28) &    46.82(49) &   39.70(963) &   42.28(631) \\
& Penalty k=-75  &   34.75(762) &    44.82(84) &    45.15(75) &    45.83(28) &     44.88(68) &    45.16(43) &    45.19(76) &  34.75(1426) &   38.56(934) \\
& Penalty k=-100 &  29.80(1010) &   43.08(113) &   43.50(102) &    44.42(36) &     43.15(93) &    43.51(59) &   43.55(104) &  29.80(1889) &  34.85(1237) \\ \\
\hline \\
\multirow{4}{*}{\rotatebox[origin=c]{90}{MPE}} 
& Speaker-Listener&   -27.64(390) &  -17.61(299) &   -17.42(323) &    -18.46(68) &    -38.20(629) &    -15.17(44) &   -15.01(64) &   -27.41(311) &   -21.29(279) \\
& Spread       &  -155.81(150) &  -152.72(96) &  -149.89(291) &  -157.10(230) &  -245.22(8446) &  -144.73(409) &  -149.26(94) &  -148.57(167) &  -154.70(490) \\
& Adversary    &      7.58(14) &     10.18(5) &     10.21(16) &     7.80(143) &       6.12(35) &     10.11(14) &      9.61(7) &      7.64(21) &      8.11(37) \\
& Tag            &    13.70(197) &   12.43(105) &    13.60(295) &     6.65(390) &       5.11(58) &    11.93(209) &   13.78(440) &    15.24(159) &    15.00(273) \\ \\
\hline \\
\multirow{5}{*}{\rotatebox[origin=c]{90}{SMAC}} 
& 2s\_vs\_1sc &  14.76(45) &    19.74(2) &   19.44(29) &  10.15(132) &   9.04(83) &  17.89(85) &    19.67(9) &   16.11(23) &   15.98(77) \\
& 3s5z      &  14.09(28) &  14.84(129) &  11.80(151) &   8.60(235) &  15.51(98) &  18.82(14) &   19.09(38) &   17.85(25) &    18.36(7) \\
& corridor  &  10.91(82) &  13.14(124) &  14.60(343) &    5.15(25) &   7.00(15) &   7.89(28) &  13.20(298) &  11.14(166) &  11.67(188) \\
& MMM2      &  10.11(32) &   7.31(189) &   9.97(133) &     3.42(5) &   6.50(17) &  9.07(135) &   15.39(16) &   15.93(23) &   15.63(32) \\
& 3s\_vs\_5z  &  17.35(23) &     4.32(4) &  13.38(436) &    5.34(47) &  1.15(135) &   6.17(39) &  13.09(263) &  14.72(401) &   9.68(187) \\
\\
\hline \\
\multirow{7}{*}{\rotatebox[origin=c]{90}{LBF}} 
& 8x8-2p-2f-c    &   0.75(4) &  0.97(0) &   0.94(2) &  0.32(2) &  0.32(12) &  0.97(0) &   0.95(1) &  0.64(9) &   0.39(10) \\
& 8x8-2p-2f-2s-c &   0.86(1) &  0.97(0) &   0.50(1) &  0.54(5) &   0.24(8) &  0.97(0) &   0.77(2) &  0.83(1) &   0.77(3) \\
& 10x10-3p-3f      &   0.54(2) &  0.95(1) &   0.90(2) &  0.20(6) &   0.15(5) &  0.95(1) &   0.91(1) &  0.40(5) &   0.32(7) \\ 
& 10x10-3p-3f-2s   &   0.69(2) &  0.84(1) &   0.62(1) &  0.27(2) &   0.23(6) &  0.85(2) &   0.66(1) &  0.64(2) &   0.67(1) \\
& 15x15-3p-5f      &  0.09(2) &  0.61(6) &   0.41(9) &  0.08(0) &   0.06(3) &  0.59(9) &   0.43(9) &  0.08(1) &  0.04(1) \\
& 15x15-4p-3f      &   0.24(5) &  0.89(3) &   0.82(6) &  0.13(1) &   0.12(3) &  0.92(1) &   0.79(3) &  0.16(3) &   0.08(1) \\
& 15x15-4p-5f     &  0.15(3) &  0.59(6) &  0.40(13) &  0.13(1) &   0.07(2) &  0.73(2) &  0.39(14) &  0.15(2) &   0.09(2) \\
\\
\hline \\
\multirow{3}{*}{\rotatebox[origin=c]{90}{RWARE}} 
& Tiny 2p  &   0.04(3) &   2.91(45) &  12.63(138) &  0.11(7) &  0.13(5) &    3.20(41) &  15.42(120) &   0.03(1) &  0.03(3) \\
& Tiny 4p  &  0.33(13) &  10.30(93) &  22.68(740) &  0.28(3) &  0.39(6) &  14.39(401) &  40.17(142) &  0.29(13) &  0.10(9) \\
& Small 4p &   0.03(4) &   2.45(18) &   9.19(236) &  0.06(2) &  0.08(1) &    3.48(42) &  18.12(111) &   0.02(3) &  0.01(1) \\
\\

\bottomrule
\end{tabular}
}
\end{table}

\begin{table}[H]
\centering
\caption{\label{tab:max_returns_ns} Maximum returns and 95\% confidence interval over five seeds for all nine algorithms without parameter sharing in all 25 tasks.}
\resizebox{\linewidth}{!}{
\robustify\bf
\begin{tabular}{p{1em} l S S  S  S  S  S  S S S S}
\toprule
& \textbf{Tasks \textbackslash Algs. }  &             {IQL} &           {IA2C} &             {IPPO} &              {MADDPG} &              {COMA} &              {MAA2C} &               {MAPPO} &             {VDN} &           {QMIX} \\ \\
\hline \\
\multirow{6}{*}{\rotatebox[origin=c]{90}{Matrix Games}}
& Climbing    &  150.00(0) &  175.00(0) &    175.00(0) &  170.00(1000) &  195.00(4000) &     175.00(0) &    175.00(0) &     150.00(0) &  155.00(1000) \\
& Penalty k=0   &  250.00(0) &  250.00(0) &    250.00(0) &     249.94(8) &     250.00(0) &     250.00(0) &    250.00(0) &     250.00(0) &     250.00(0) \\
& Penalty k=-25 &   50.00(0) &   50.00(0) &  90.00(8000) &   89.99(8001) &      50.00(0) &  130.00(9798) &  90.00(8000) &  50.00(10000) &  50.00(10000) \\
& Penalty k=-50 &   50.00(0) &   50.00(0) &     50.00(0) &      49.98(1) &      50.00(0) &      50.00(0) &     50.00(0) &  50.00(10000) &  50.00(10000) \\
& Penalty k=-75  &   50.00(0) &   50.00(0) &     50.00(0) &      49.98(1) &      50.00(0) &      50.00(0) &     50.00(0) &  50.00(10000) &  50.00(10000) \\
& Penalty k=-100 &   50.00(0) &   50.00(0) &     50.00(0) &      49.98(1) &      50.00(0) &      50.00(0) &     50.00(0) &  50.00(10000) &  50.00(10000) \\ \\
\hline \\
\multirow{4}{*}{\rotatebox[origin=c]{90}{MPE}} & Speaker-Listener   &   -18.61(565) &   -17.08(345) &   -15.56(440)* &   -12.73(73)* &    -26.50(50) &   -13.66(367)* &   -14.35(356)* &   -15.47(126) &   \bf -11.59(67) \\
& Spread          &  -141.87(168) &  -131.74(433)* &  -132.46(354)* &  -136.73(83) &  -169.04(272) &  -130.88(244)* & \bf  -128.64(283) &  -142.13(186) &  -130.97(251)* \\
& Adversary      &      9.09(52) &    10.80(197)* &     11.17(85)* &     8.81(61) &      9.18(43) &    10.88(243)* &     \bf 12.04(53) &      9.34(57) &     11.32(78)* \\
& Tag            &    19.18(230) &    16.04(808)* &    18.46(519)* &    2.82(356) &    19.14(750)* &    26.50(142)* &    17.96(882)* &    18.44(251) &   \bf 26.88(561) \\ \\
\hline \\
\multirow{5}{*}{\rotatebox[origin=c]{90}{SMAC}}  & 
2s\_vs\_1sc &   15.73(108) &     20.23(1) &   20.15(10)* &  10.35(220) &  18.48(328)* &  19.88(38)* &   \bf 20.25(1) &           17.22(90) &            18.83(47) \\
& 3s5z      &   16.85(143) &   14.44(208) &   14.77(251) &   15.05(81) &   19.55(45)* &   19.38(30) &  \bf 19.77(11) &           19.08(29) &            18.40(70) \\
& MMM2      &   10.86(104) &    8.38(290) &     7.35(48) &    4.92(10) &     4.98(42) &   10.79(34) &           10.02(19) &  \bf 16.20(44)&           12.27(228) \\
& corridor  &  11.06(136) &  13.11(427)* &  10.29(460) &    6.57(53) &     8.34(81) &   9.26(308) &            8.51(76) &         11.42(218)* &  \bf 15.12(342) \\
& 3s\_vs\_5z  &  11.80(105)* &      4.41(2) &     4.26(24) &   6.21(197) &      4.13(5) &    5.39(74) &            4.62(11) &         14.42(222)* &  \bf 15.13(386)\\\\
\hline \\
\multirow{7}{*}{\rotatebox[origin=c]{90}{LBF}} 
& 8x8-2p-2f-c    &    \bf 1.00(0) &   \bf 1.00(0) &    \bf 1.00(0) &   0.43(2) &  0.95(7)* &     \bf 1.00(0) &   \bf 1.00(0) &    0.99(1)* & 0.57(15) \\
& 8x8-2p-2f-2s-c &    0.99(1)* &  \bf 1.00(0) &    0.83(3) &   0.66(4) &  0.92(2) &     \bf 1.00(0) &   0.92(5) &    0.99(1)* &    0.98(1) \\
& 10x10-3p-3f   &  0.54(12) &   \bf 1.00(1) &    0.96(1) &   0.20(3) &  0.31(8) &     0.99(1)* &   0.98(1) &  0.39(6) &        0.22(3) \\ 
& 10x10-3p-3f-2s    &    0.77(6) &   0.89(2) &    0.72(2) &   0.29(2) &  0.27(9) &     \bf 0.96(1) &   0.70(4) &    0.76(4) & 0.78(5) \\
& 15x15-3p-5f   &    0.11(2) &  \bf 0.74(12) &   0.62(12)* &   0.10(1) &  0.09(2) &     0.72(5)* &  0.49(19)* &    0.11(1) &    0.06(1) \\
& 15x15-4p-3f      &   0.16(3) &   \bf 0.99(0) &  0.90(1) &   0.17(1) &  0.21(7) &     0.94(5)* &   0.89(2) &  0.13(2) &    0.10(1) \\
& 15x15-4p-5f       &    0.17(0) &  \bf 0.77(10) &    0.69(5)* &   0.15(1) &  0.14(3) &     0.76(4)* &  0.45(17) &    0.14(1) &    0.08(1) \\ \\
\hline \\
\multirow{3}{*}{\rotatebox[origin=c]{90}{RWARE}} 
& Tiny 2p  &   0.06(8) &   5.56(168) &   8.70(404) &  0.24(21) &  1.46(34) &   6.16(194) &  \bf 17.48(452) &   0.06(8) &   0.06(8) \\
& Tiny 4p &  0.44(10) &  18.02(587) &  14.10(520) &  0.44(24) &  1.48(48) &  31.56(429) &  \bf 38.74(299) &  0.34(17) &  0.16(15) \\
& Small 4p &   0.04(5) &    3.10(68) &    5.78(42) &  0.16(12) &  0.16(10) &    5.00(68) &  \bf 13.78(163) &   0.04(5) &   0.06(8) \\ \\

\bottomrule
\end{tabular}
}
\end{table}

\newpage

\begin{table}[H]
\centering
\caption{\label{tab:mean_returns_ns} Average returns and 95\% confidence interval over five seeds for all nine algorithms without parameter sharing in all 25 tasks.}
\resizebox{\linewidth}{!}{
\robustify\bf
\begin{tabular}{p{1em} l S S  S  S  S  S  S S S S}
\toprule
& \textbf{Tasks \textbackslash Algs. }  &             {IQL} &           {IA2C} &             {IPPO} &              {MADDPG} &              {COMA} &              {MAA2C} &               {MAPPO} &             {VDN} &           {QMIX} \\ \\
\hline \\
\multirow{6}{*}{\rotatebox[origin=c]{90}{Matrix Games}}
& Climbing   &  130.20(437) &  164.72(69) &   171.68(41) &  150.05(375) &  187.50(4119) &   169.43(101) &   171.62(39) &  132.52(320) &  132.43(337) \\
& Penalty k=0   &  246.73(139) &  243.65(93) &    247.72(4) &    247.88(5) &    247.10(38) &    246.61(52) &    247.73(3) &  247.03(185) &   247.03(99) \\
& Penalty k=-25  &    49.70(24) &   46.95(16) &  88.05(7901) &  86.14(7519) &     48.32(10) &  125.90(9514) &  88.06(7903) &    50.00(99) &    50.00(99) \\
& Penalty k=-50  &    49.70(24) &   44.31(34) &    46.99(23) &     47.12(7) &     46.54(20) &     46.56(33) &    46.99(24) &    50.00(99) &    50.00(99) \\
& Penalty k=-76 &    49.70(24) &   41.64(50) &    45.44(36) &    45.74(12) &     44.77(31) &     44.88(47) &    45.44(37) &    50.00(99) &    50.00(99) \\
& Penalty k=-100 &    49.70(24) &   38.94(66) &    43.89(48) &    44.33(16) &     42.99(41) &     43.21(62) &    43.90(52) &    50.00(99) &    50.00(99) \\\\
\hline \\
\multirow{4}{*}{\rotatebox[origin=c]{90}{MPE}} & Speaker-Listener &     -30.49(467) &   -23.33(267) &   -22.78(310) &    -17.79(97) &   -33.88(338) &   -19.48(292) &   -20.51(285) &   -29.49(236) &   -20.31(184) \\
& Spread       &  -160.10(159) &  -141.31(386) &  -142.86(449) &  -149.53(241) &  -184.72(299) &  -139.54(478) &  -139.20(458) &  -158.60(206) &  -157.04(138) \\
& Adversary     &      7.82(19) &     9.18(152) &     9.46(102) &     7.24(213) &      7.28(76) &     9.43(193) &     10.20(24) &      8.06(27) &      8.81(52) \\
& Tag        &    12.59(160) &     9.59(430) &    11.90(331) &     1.91(209) &    14.28(543) &    11.79(540) &    10.90(547) &     10.71(69) &    14.27(281) \\ \\
\hline \\
\multirow{5}{*}{\rotatebox[origin=c]{90}{SMAC}} 
& 2s\_vs\_1sc &  13.37(35) &    19.69(5) &  8.59(190) &  7.91(16) &  15.32(321) &  16.22(375) &   19.68(8) &   15.12(46) &   15.66(82) \\
& 3s5z      &  14.23(84) &  12.65(100) &  12.26(98) &  7.65(54) &  17.13(111) &   17.94(28) &  19.03(19) &   17.06(28) &   15.46(59) \\ 
& MMM2      &   8.27(64) &   6.98(172) &   6.23(88) &       3.11(12) &    3.82(36) &    9.85(19) &   9.41(19) &   12.72(56) &   8.05(205) \\
& corridor  &   8.38(82) &   9.81(198) &   7.80(68) &       4.99 (17) &    7.85(46) &    7.78(73) &   8.18(43) &    9.25(85) &  10.76(179) \\
&3s\_vs\_5z  &   9.15(46) &     4.26(2) &   4.20(22) &  3.53(39) &    3.20(89) &    4.91(41) &    4.55(2) &  10.85(161) &  10.09(110) \\ \\
\hline \\
\multirow{7}{*}{\rotatebox[origin=c]{90}{LBF}} 
& 8x8-2p-2f-c   &    0.95(2) &   0.96(1) &    0.91(2) &   0.40(5) &  0.63(13) &     0.97(0) &  0.93(2) &    0.93(0) &    0.90(0) \\
& 8x8-2p-2f-2s-c &    0.97(0) &   0.96(0) &    0.50(0) &   0.52(2) &   0.63(8) &     0.97(0) &  0.79(3) &    0.96(0) &    0.96(0) \\
& 10x10-3p-3f   &  0.77(5) &   0.92(1) &    0.85(1) &   0.15(0) &   0.24(3) &     0.92(1) &  0.85(2) &  0.80(3) &        0.81(4) \\ 
& 10x10-3p-3f-2s   &    0.78(2) &   0.76(1) &    0.61(1) &   0.22(3) &   0.23(5) &     0.80(0) &  0.62(0) &    0.86(2) &  0.86(2) \\
& 15x15-3p-5f       &    0.11(2) &   0.39(9) &   0.33(13) &   0.07(0) &   0.06(0) &     0.40(3) &  0.24(9) &    0.18(1) &    0.08(3) \\
& 15x15-4p-3f   &    0.29(4) &   0.81(1) &  0.60(1) &   0.10(1) &   0.16(3) &     0.74(3) &  0.65(6) &  0.55(6) &    0.12(4) \\
& 15x15-4p-5f       &    0.17(3) &  0.49(10) &   0.40(11) &   0.11(0) &   0.09(1) &     0.44(3) &  0.25(4) &    0.25(1) &    0.11(2) \\
\\
\hline \\
\multirow{3}{*}{\rotatebox[origin=c]{90}{RWARE}}
& Tiny 2p  &  0.01(1) &   2.02(68) &  5.36(305) &   0.07(5) &   0.37(4) &    2.23(45) &  10.39(304) &  0.01(1) &  0.01(1) \\
& Tiny 4p  &  0.13(4) &  6.10(138) &  9.24(420) &   0.25(4) &  0.50(12) &  10.73(131) &  25.14(144) &  0.10(2) &  0.05(2) \\
& Small 4p &  0.01(0) &   1.02(10) &   3.64(38) &   0.03(2) &   0.07(2) &     1.27(9) &    7.06(63) &  0.01(1) &  0.01(1) \\\\

\bottomrule
\end{tabular}
}
\end{table}

\section{Visualisation of the Evaluation Returns During Training}

\begin{figure}[H]
    \centering
        \begin{subfigure}{0.19\textwidth}
        \includegraphics[width=1.0\textwidth]{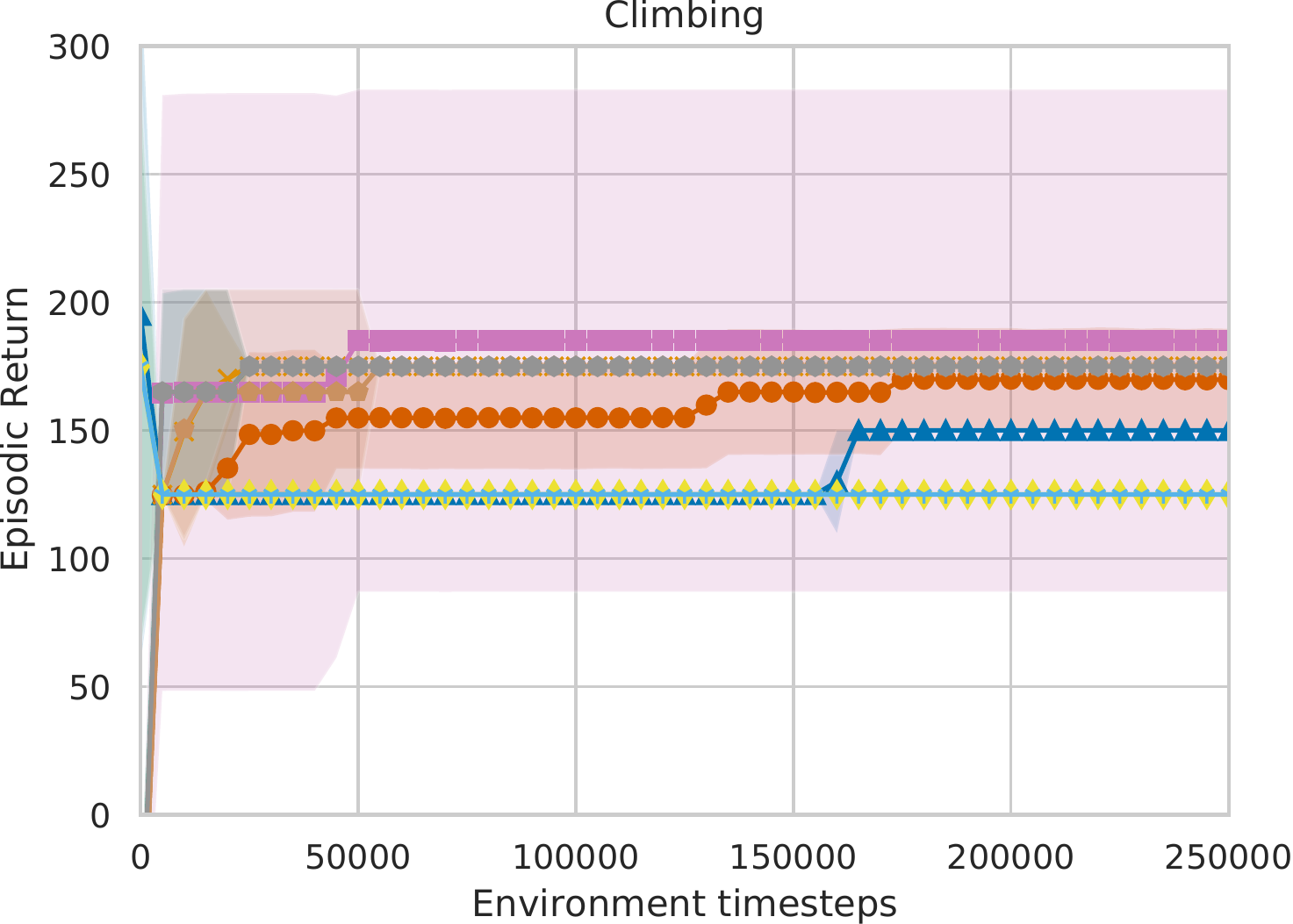}
    \end{subfigure}
    \hfill
    \begin{subfigure}{0.19\textwidth}
        \includegraphics[width=1.0\textwidth]{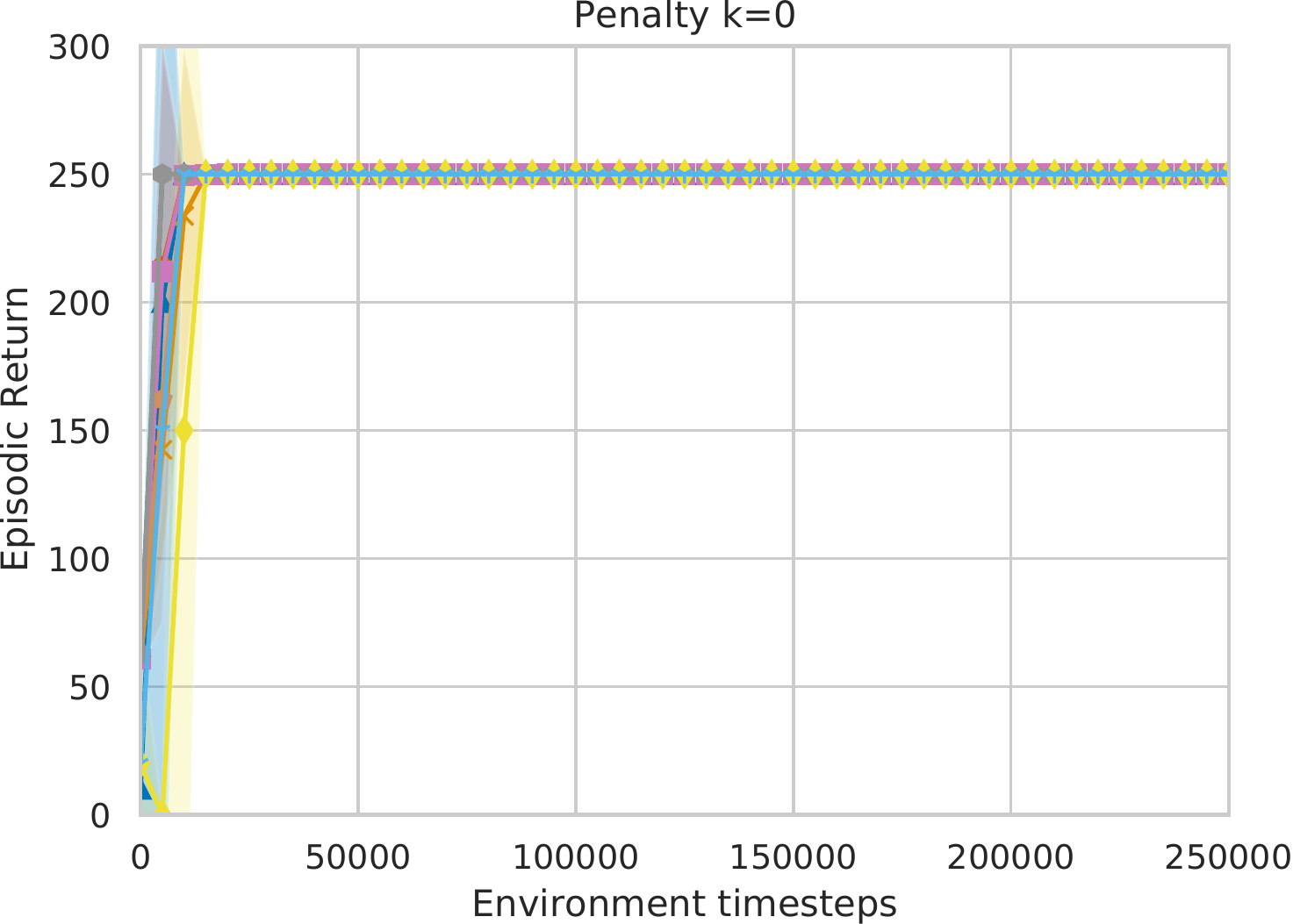}
    \end{subfigure}
     \hfill
    \begin{subfigure}{0.19\textwidth}
        \includegraphics[width=1.0\textwidth]{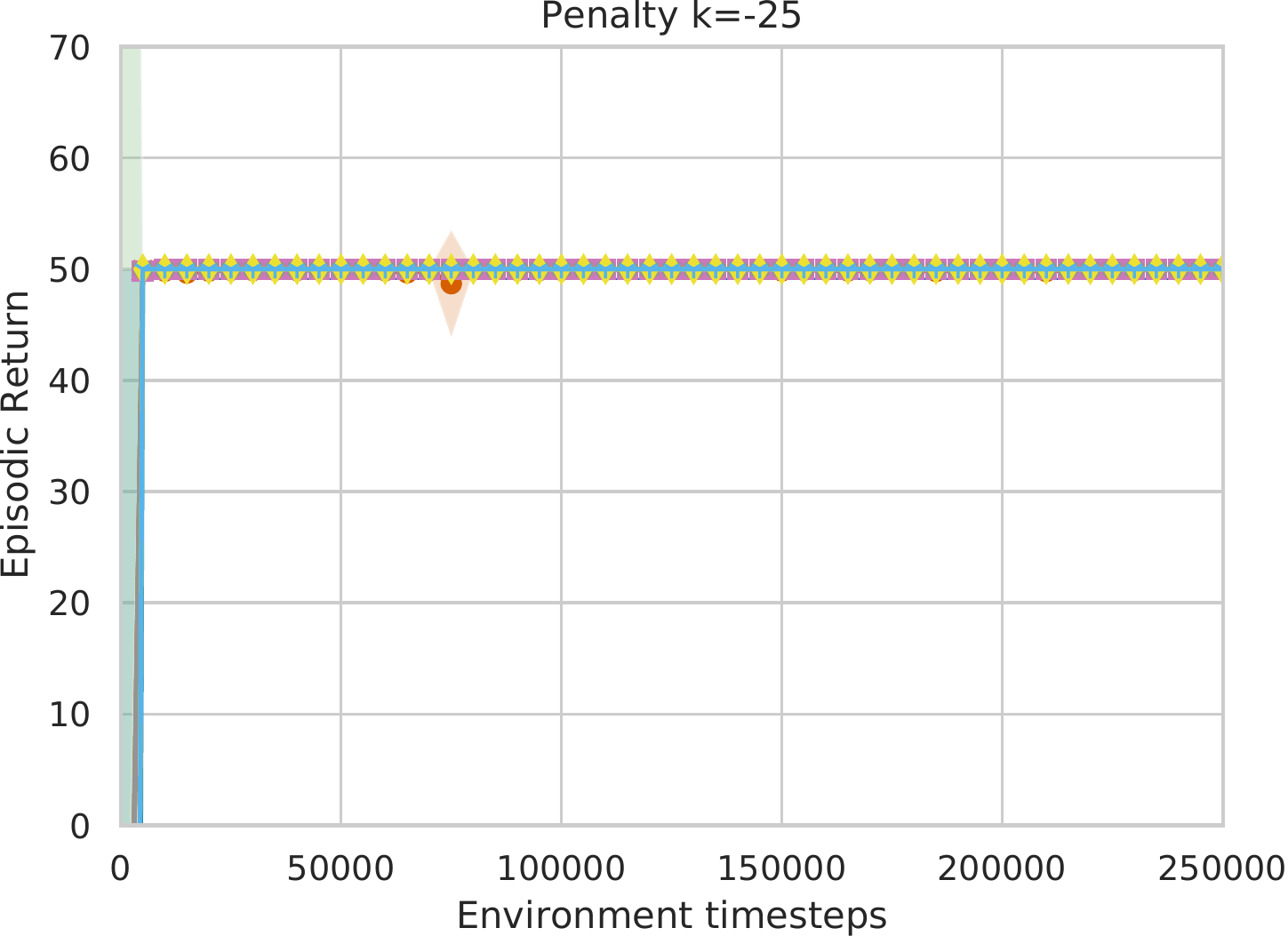}
    \end{subfigure}
     \hfill
    \begin{subfigure}{0.19\textwidth}
        \includegraphics[width=1.0\textwidth]{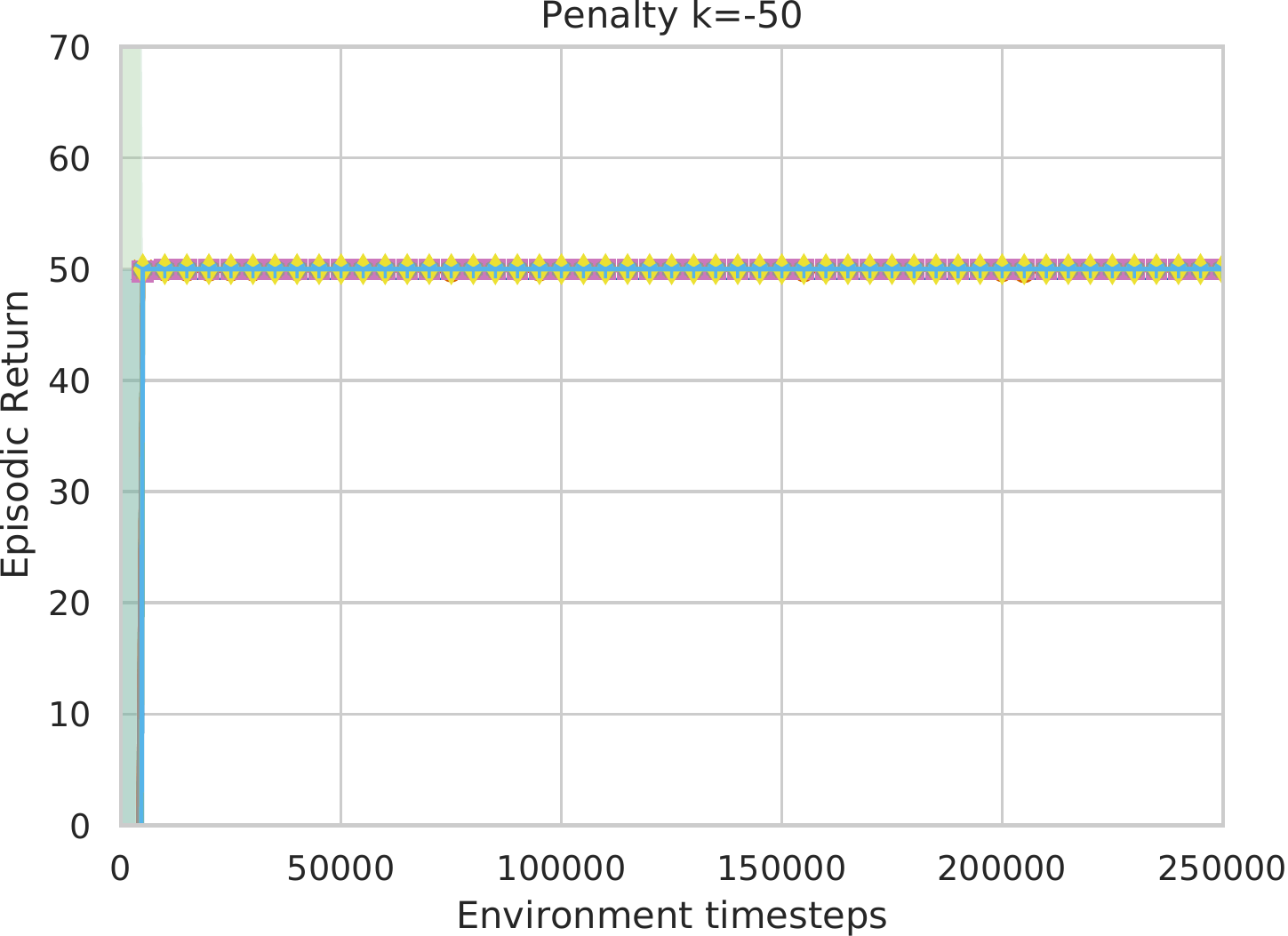}
    \end{subfigure}
      \hfill
    \begin{subfigure}{0.19\textwidth}
        \includegraphics[width=1.0\textwidth]{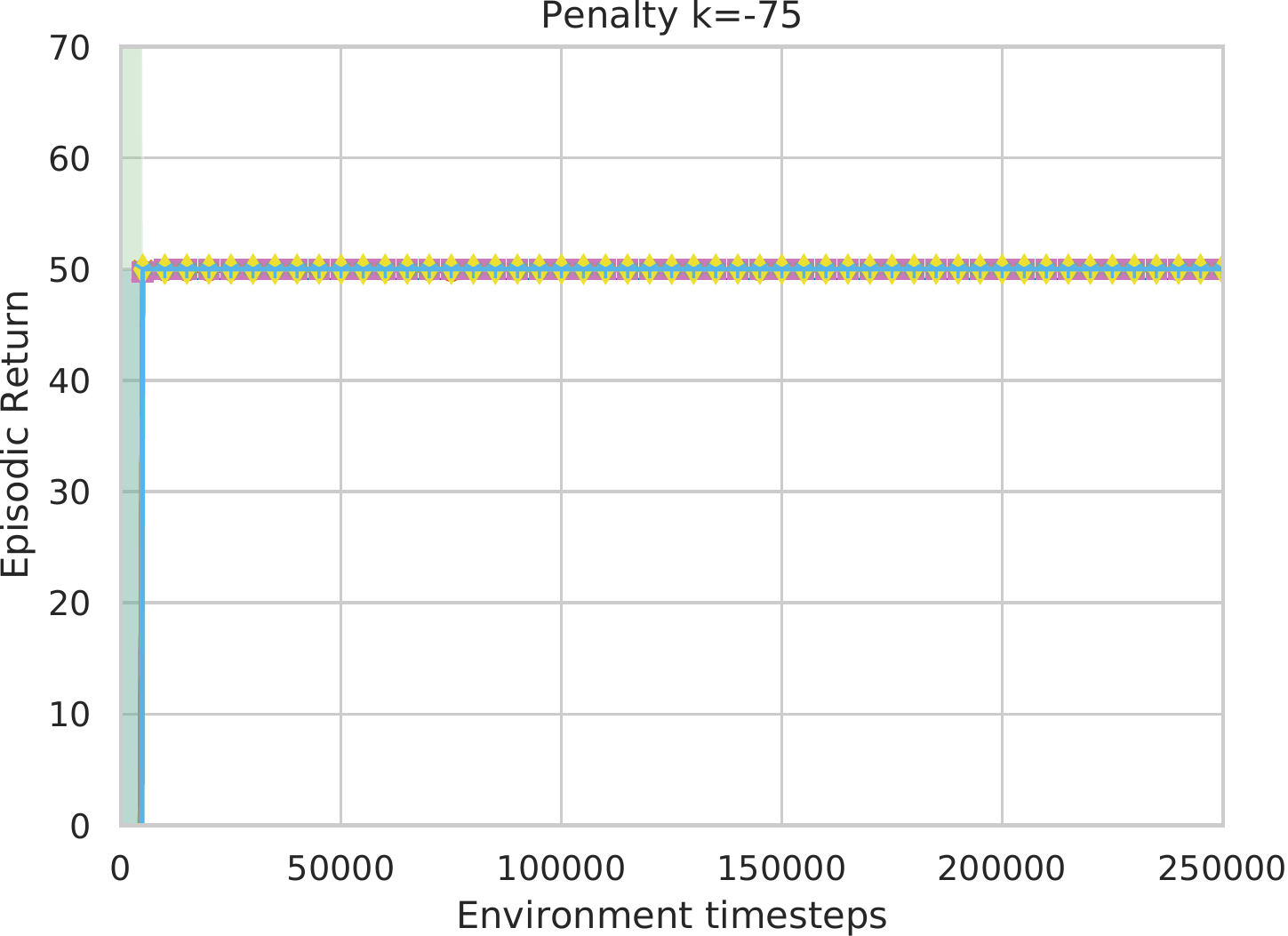}
    \end{subfigure}
     \hfill
     \begin{subfigure}{0.19\textwidth}
        \includegraphics[width=1.0\textwidth]{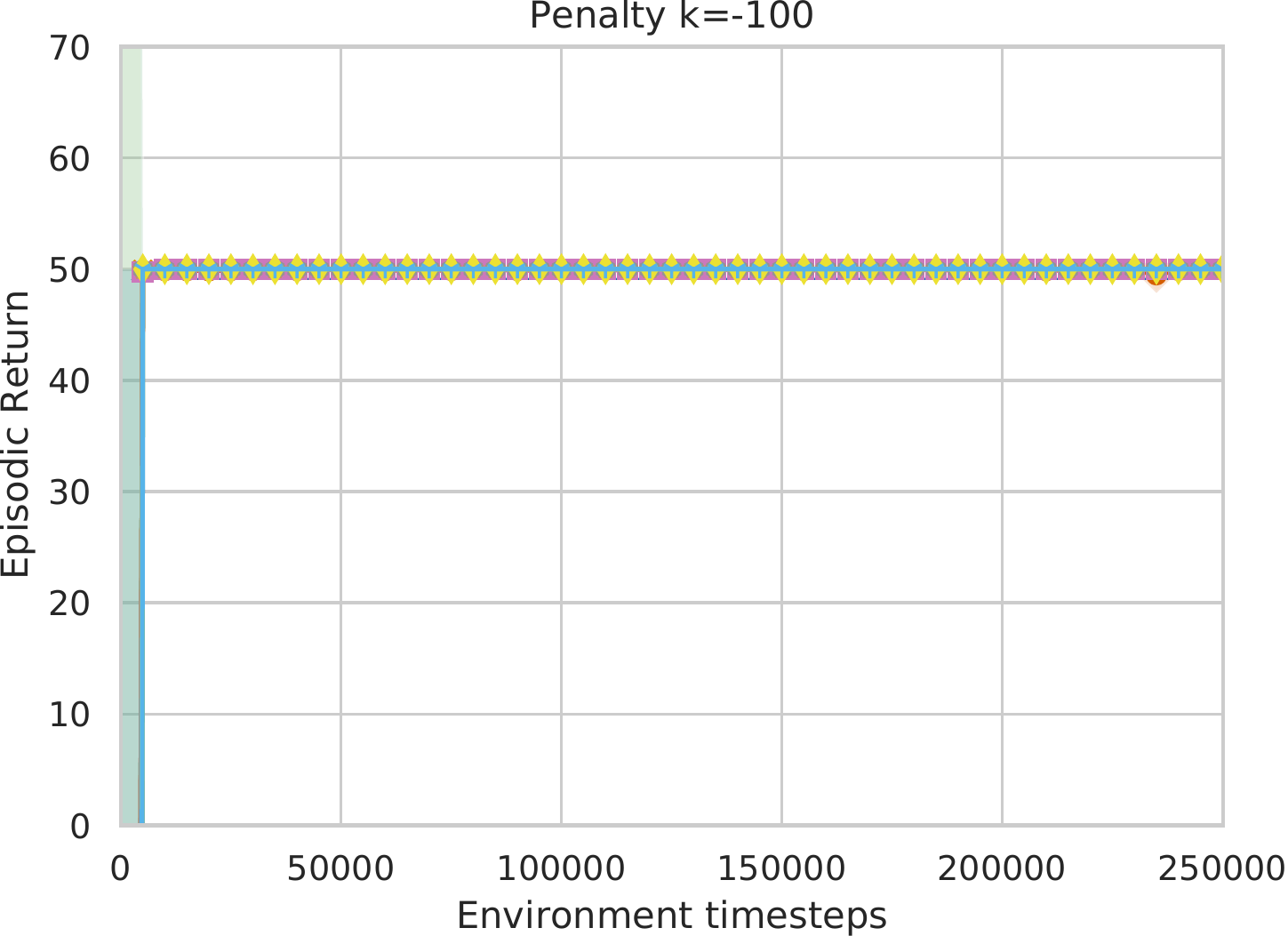}
    \end{subfigure}
     \hfill
        \begin{subfigure}{0.19\textwidth}
        \includegraphics[width=1.0\textwidth]{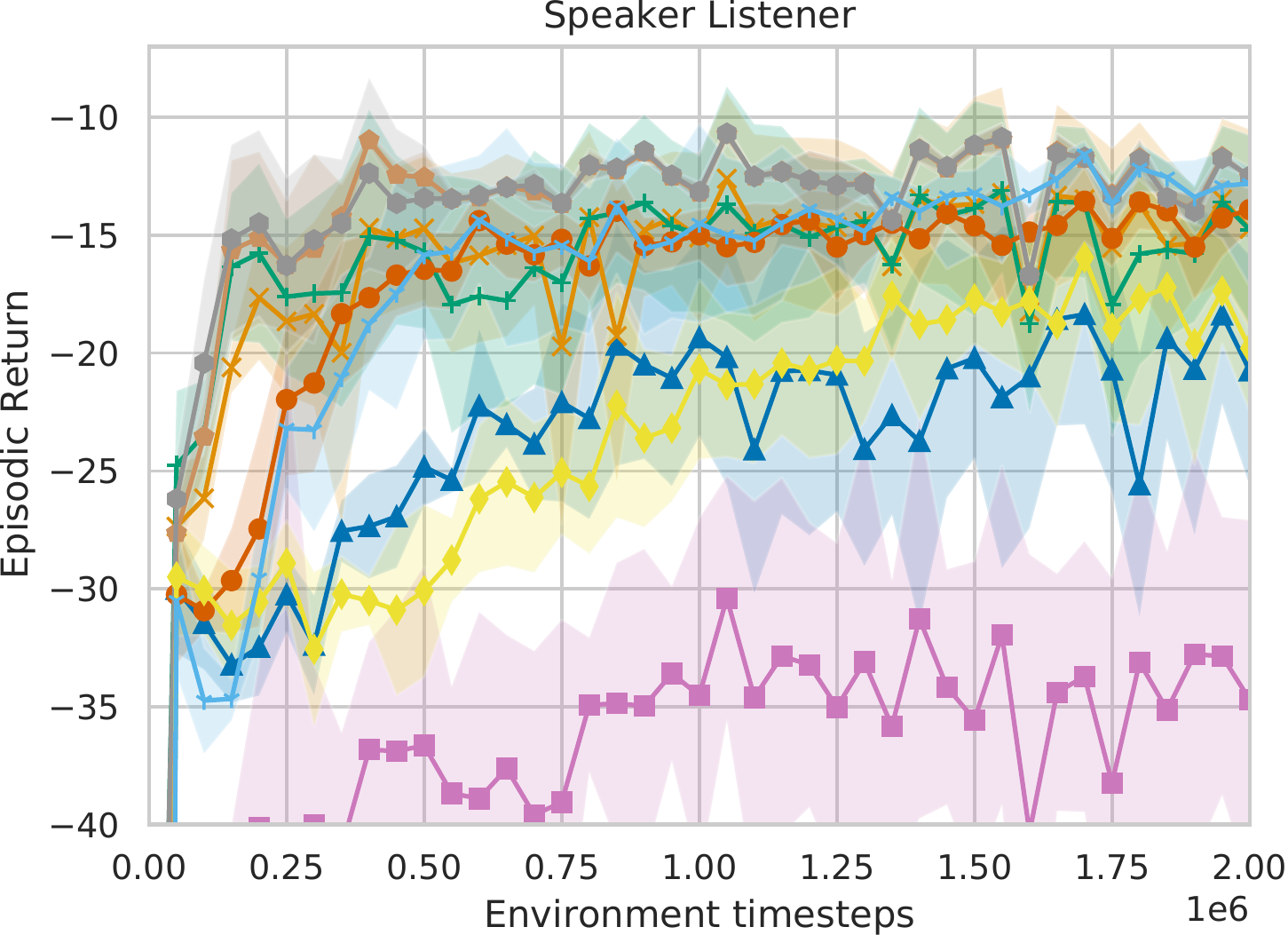}
    \end{subfigure}
    \hfill
    \begin{subfigure}{0.19\textwidth}
        \includegraphics[width=1.0\textwidth]{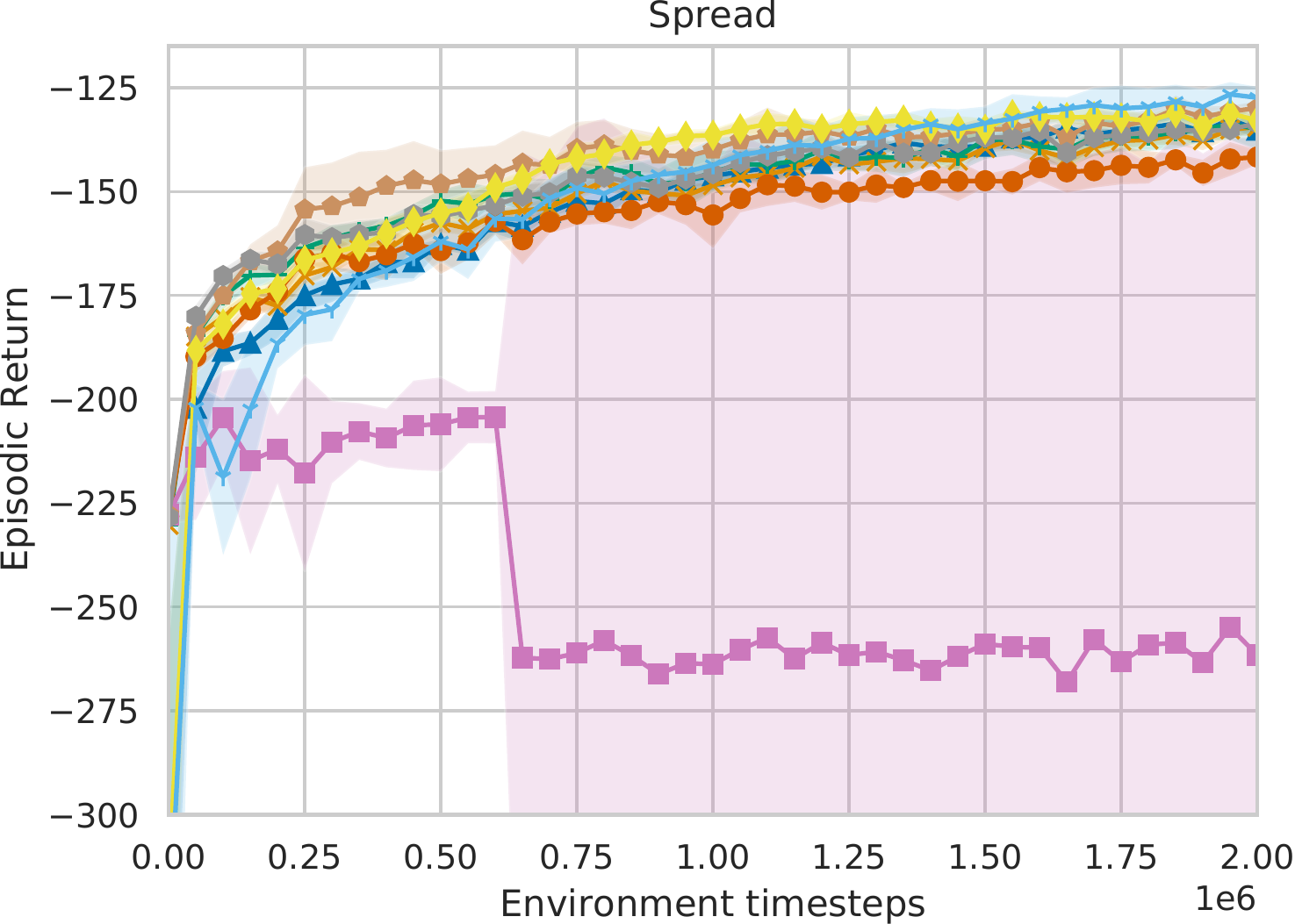}
    \end{subfigure}
     \hfill
    \begin{subfigure}{0.19\textwidth}
        \includegraphics[width=1.0\textwidth]{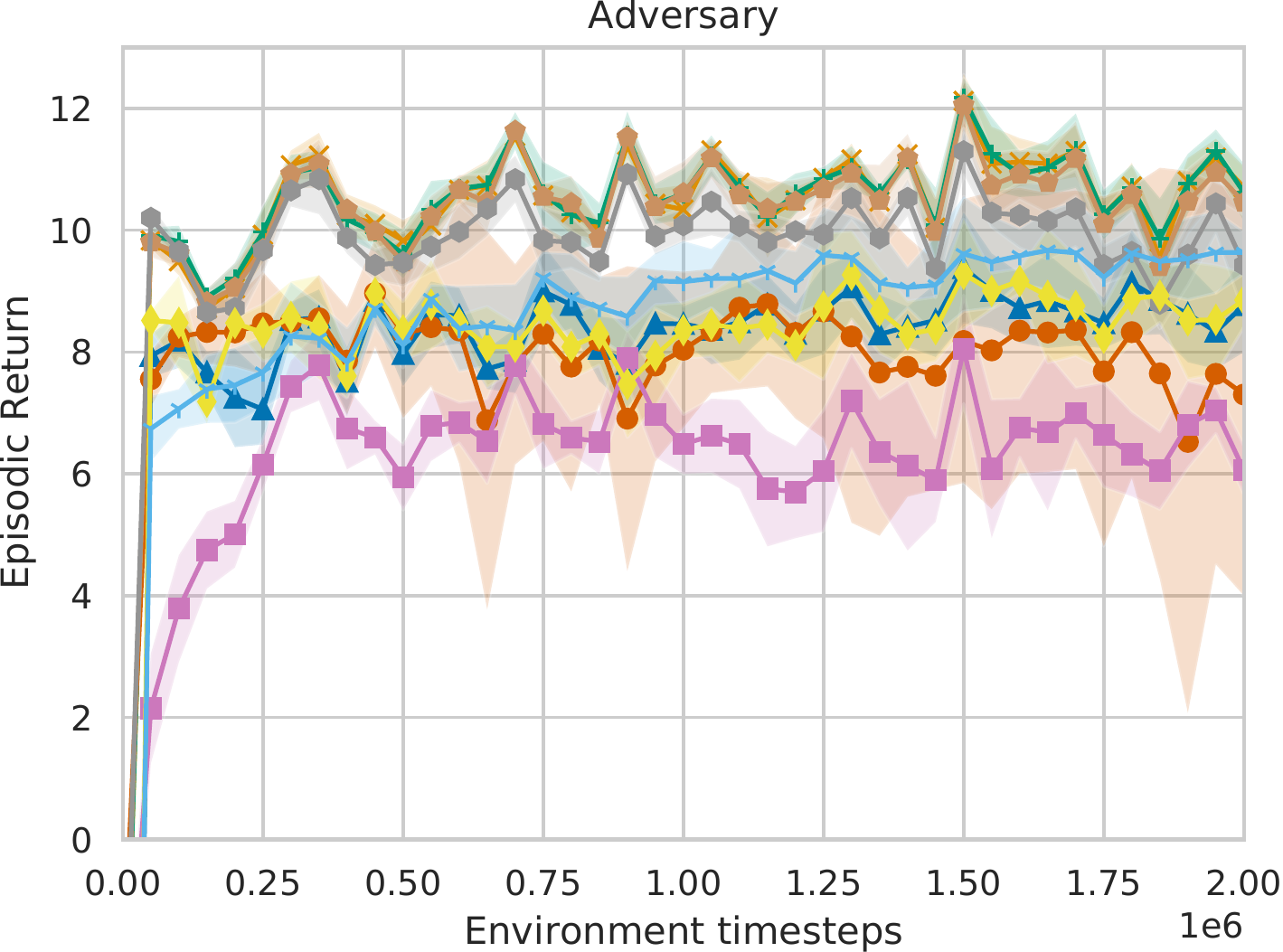}
    \end{subfigure}
     \hfill
    \begin{subfigure}{0.19\textwidth}
        \includegraphics[width=1.0\textwidth]{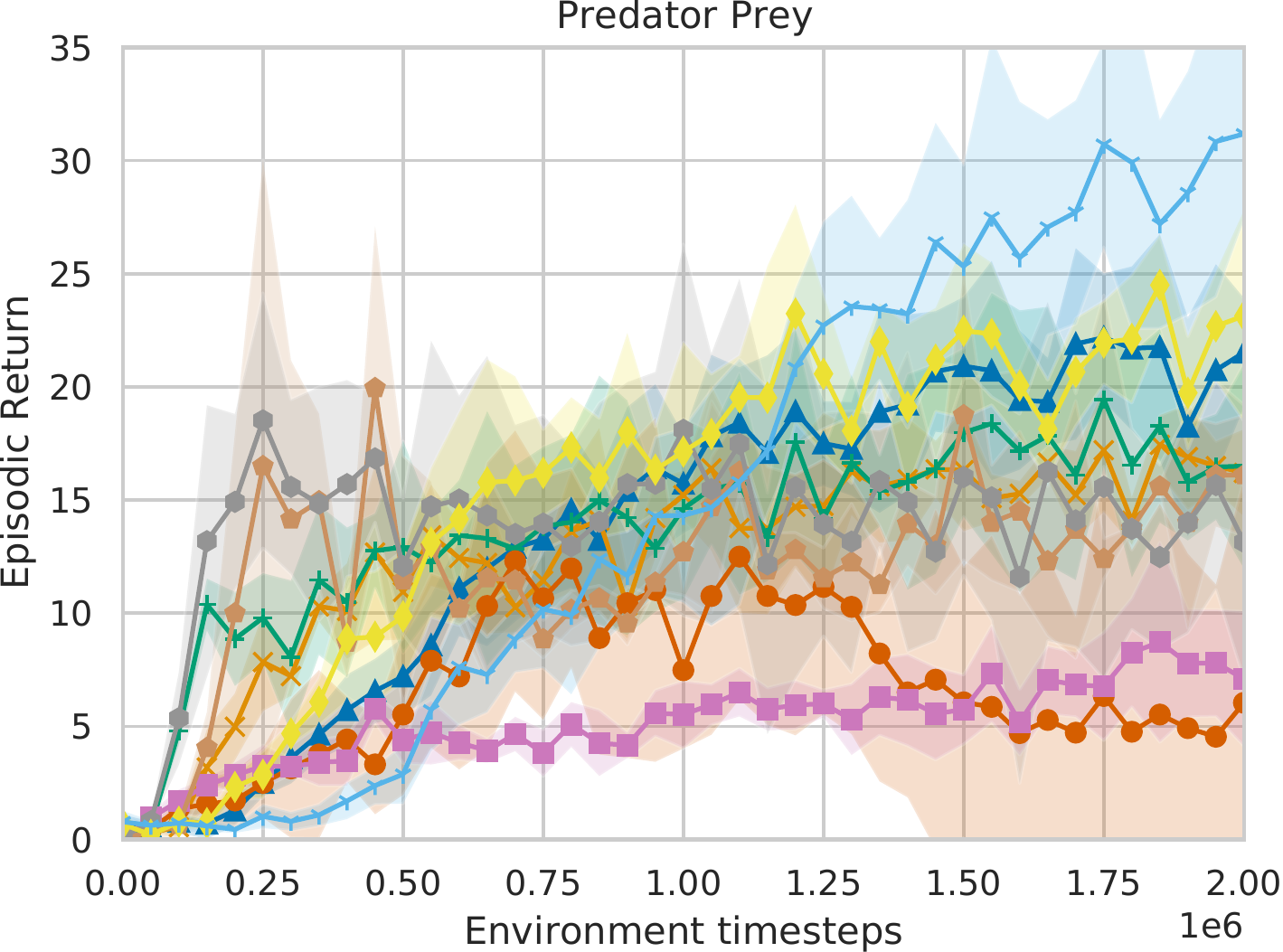}
    \end{subfigure}
      \hfill
    \begin{subfigure}{0.19\textwidth}
        \includegraphics[width=1.0\textwidth]{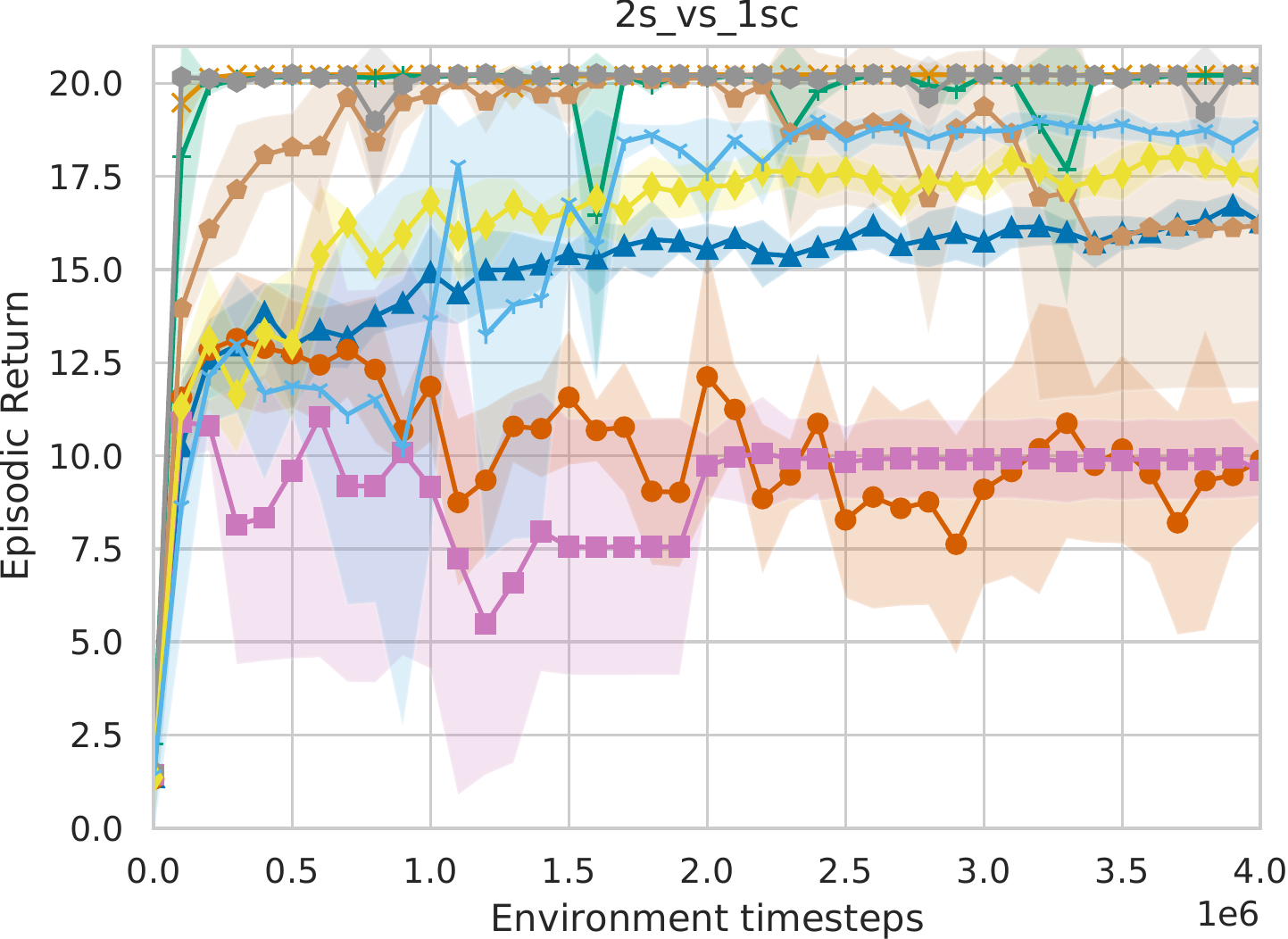}
    \end{subfigure}
     \hfill
    \begin{subfigure}{0.19\textwidth}
        \includegraphics[width=1.0\textwidth]{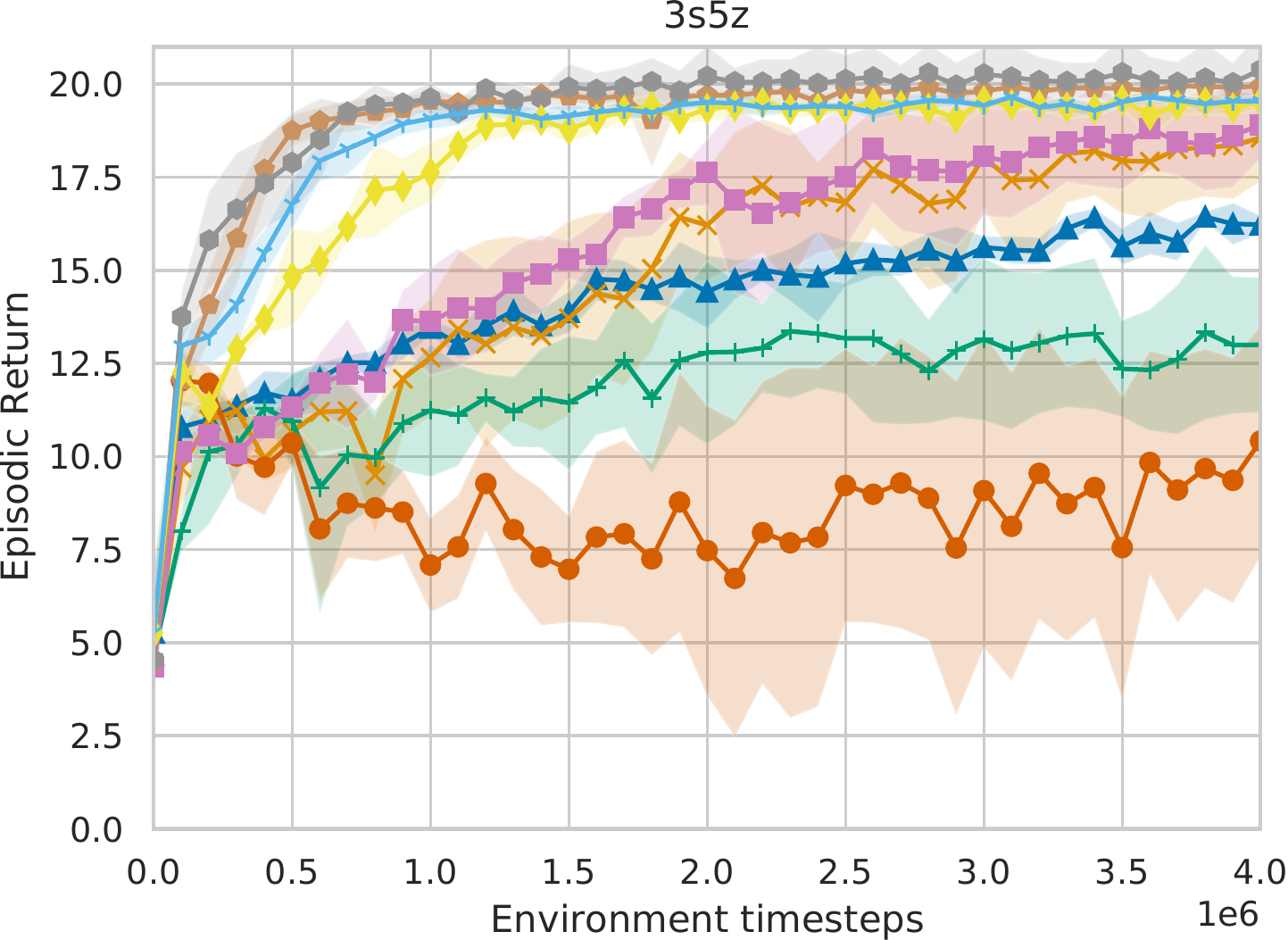}
    \end{subfigure}
     \hfill
      \hfill
    \begin{subfigure}{0.19\textwidth}
        \includegraphics[width=1.0\textwidth]{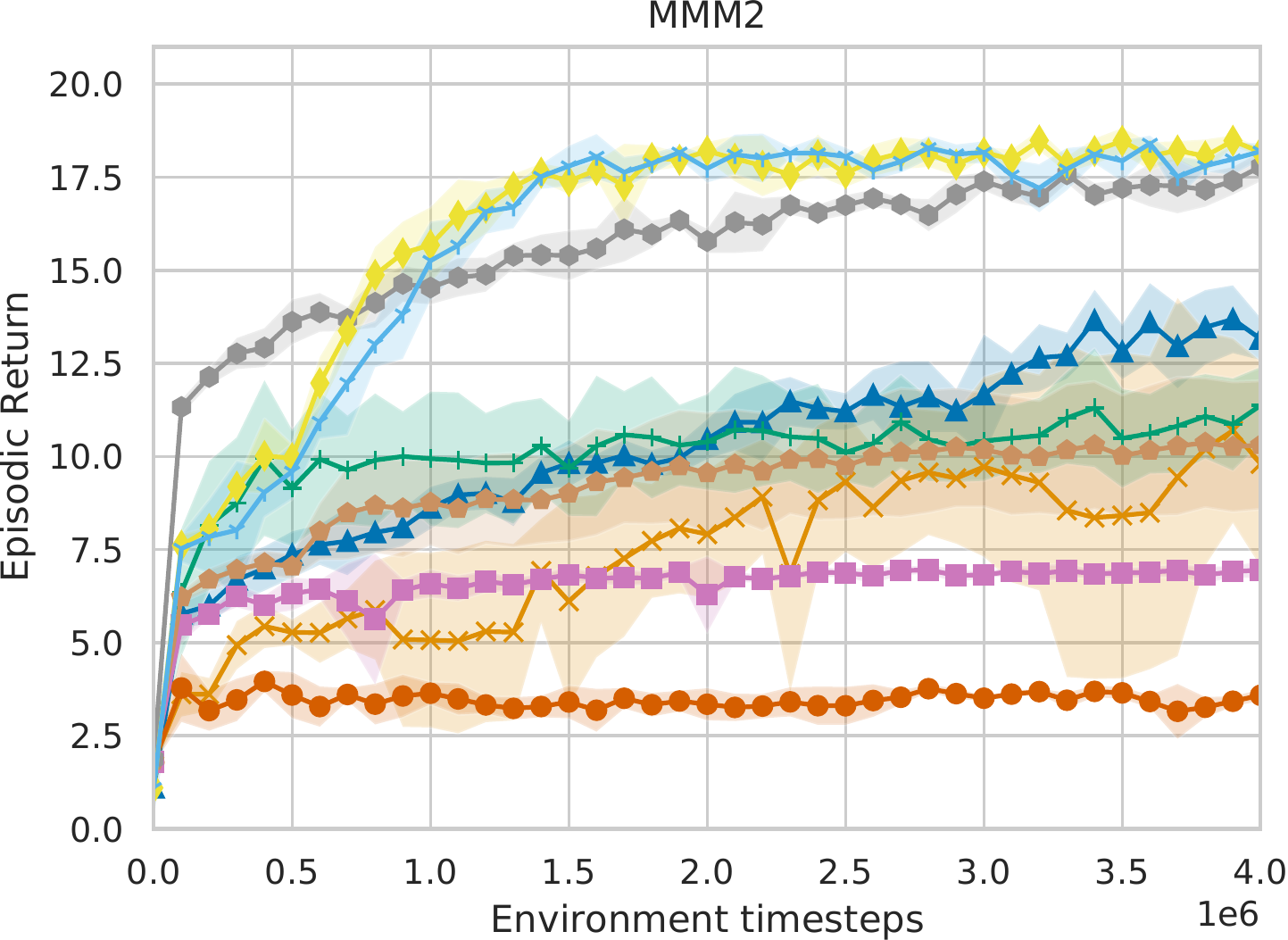}
    \end{subfigure}
     \hfill
      \hfill
    \begin{subfigure}{0.19\textwidth}
        \includegraphics[width=1.0\textwidth]{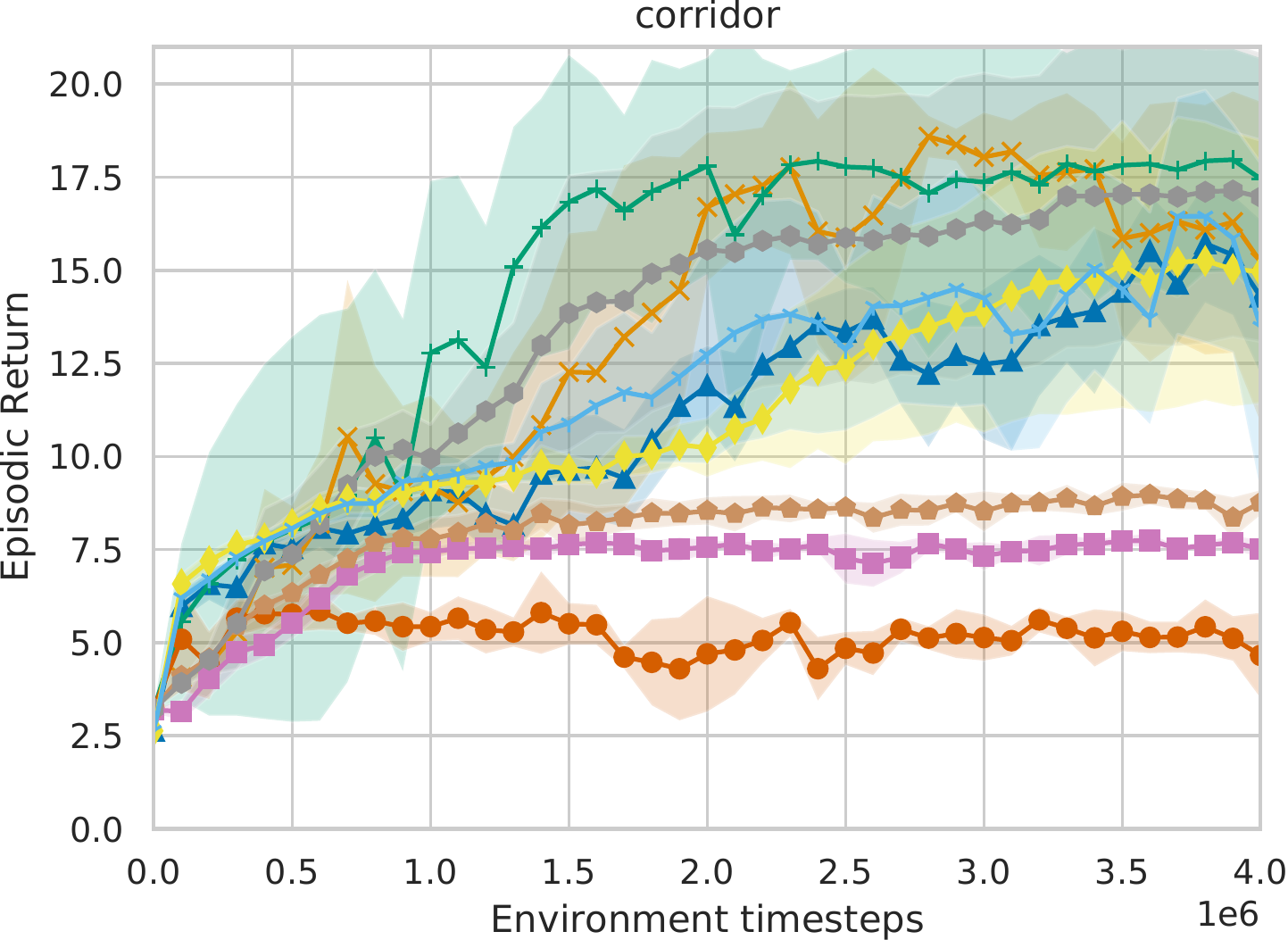}
    \end{subfigure}
     \hfill
      \hfill
    \begin{subfigure}{0.19\textwidth}
        \includegraphics[width=1.0\textwidth]{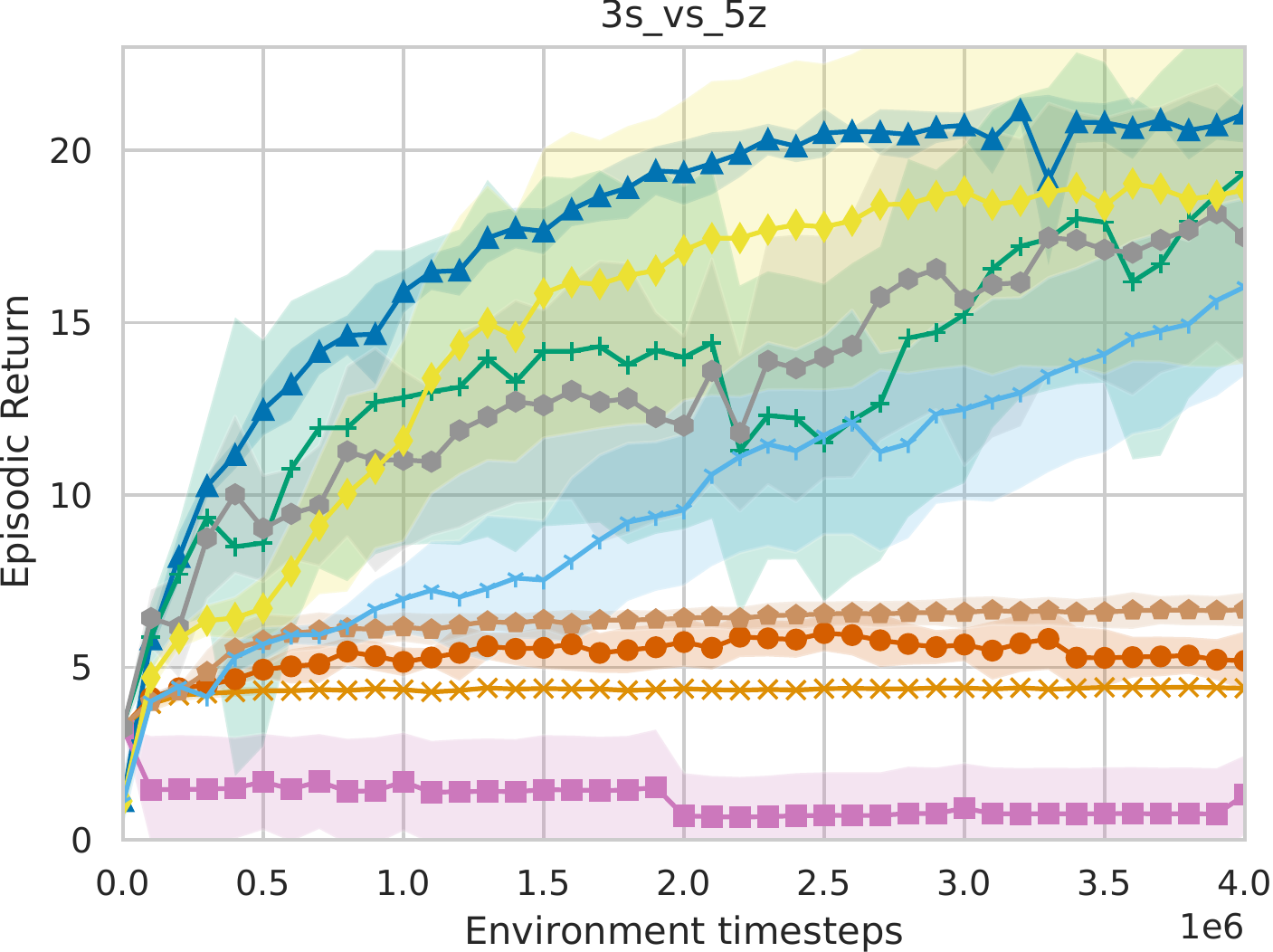}
    \end{subfigure}
     \hfill
    \begin{subfigure}{0.19\textwidth}
        \includegraphics[width=1.0\textwidth]{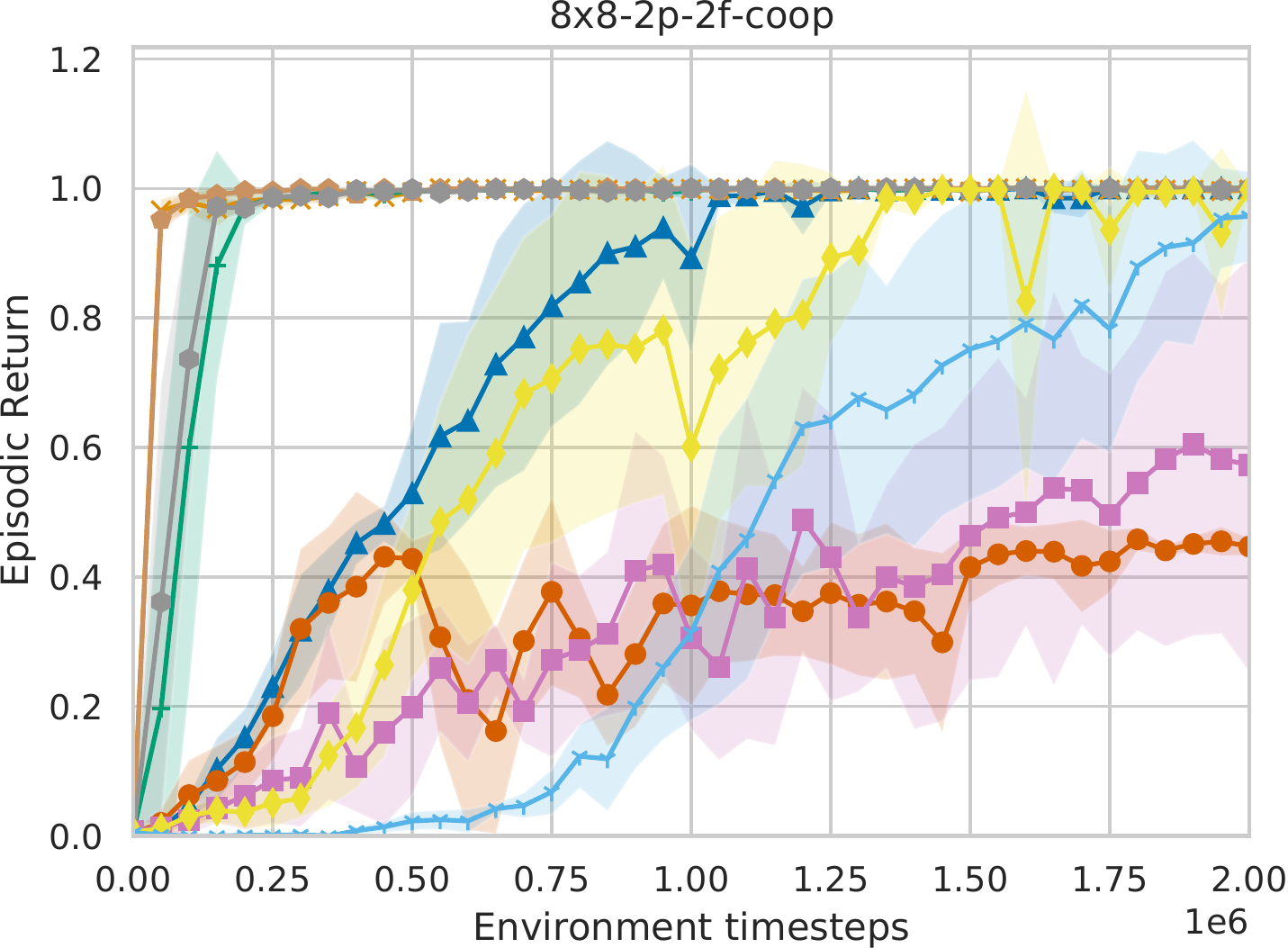}
    \end{subfigure}
     \hfill
    \begin{subfigure}{0.19\textwidth}
        \includegraphics[width=1.0\textwidth]{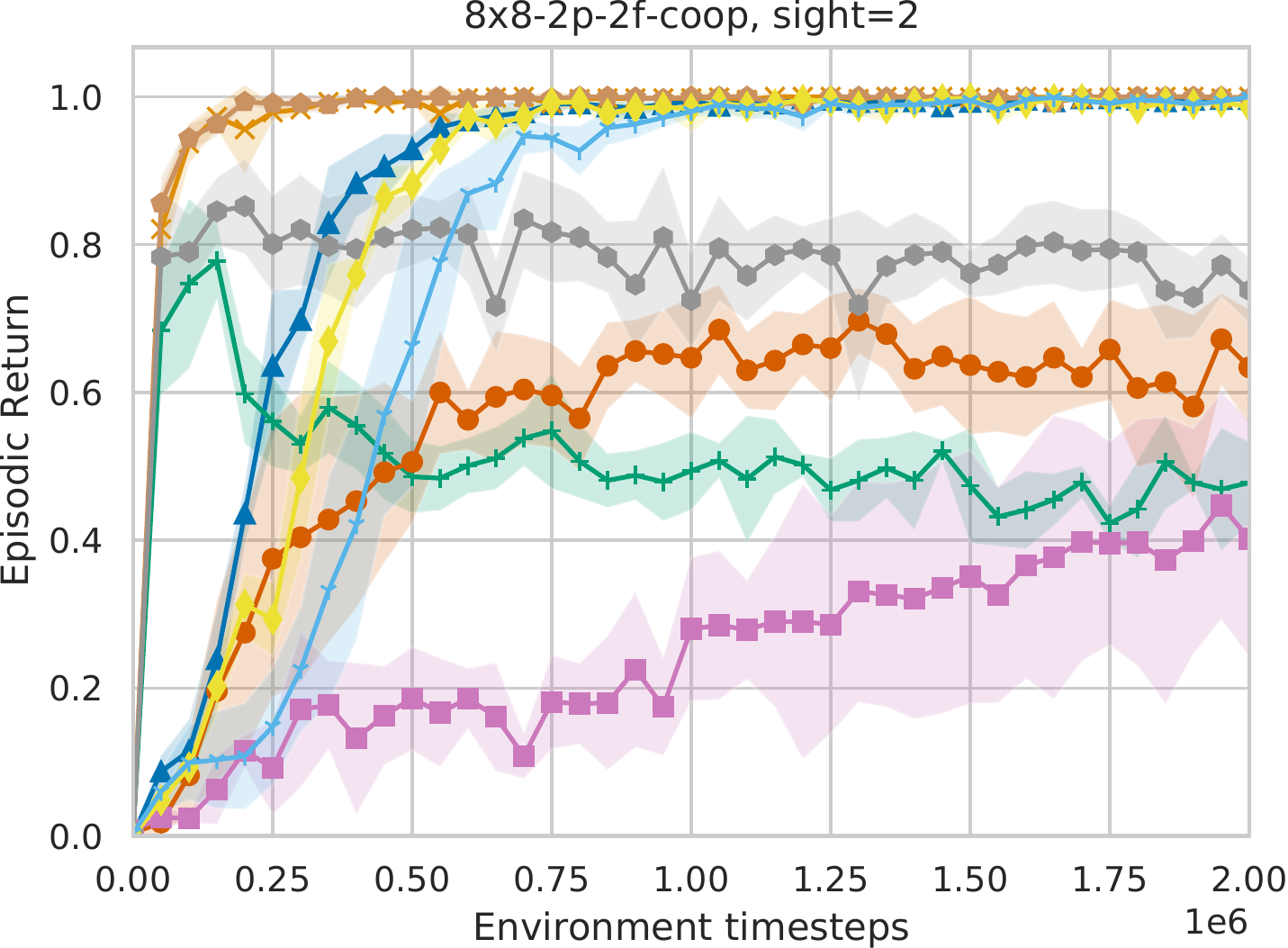}
    \end{subfigure}
     \hfill
    \begin{subfigure}{0.19\textwidth}
        \includegraphics[width=1.0\textwidth]{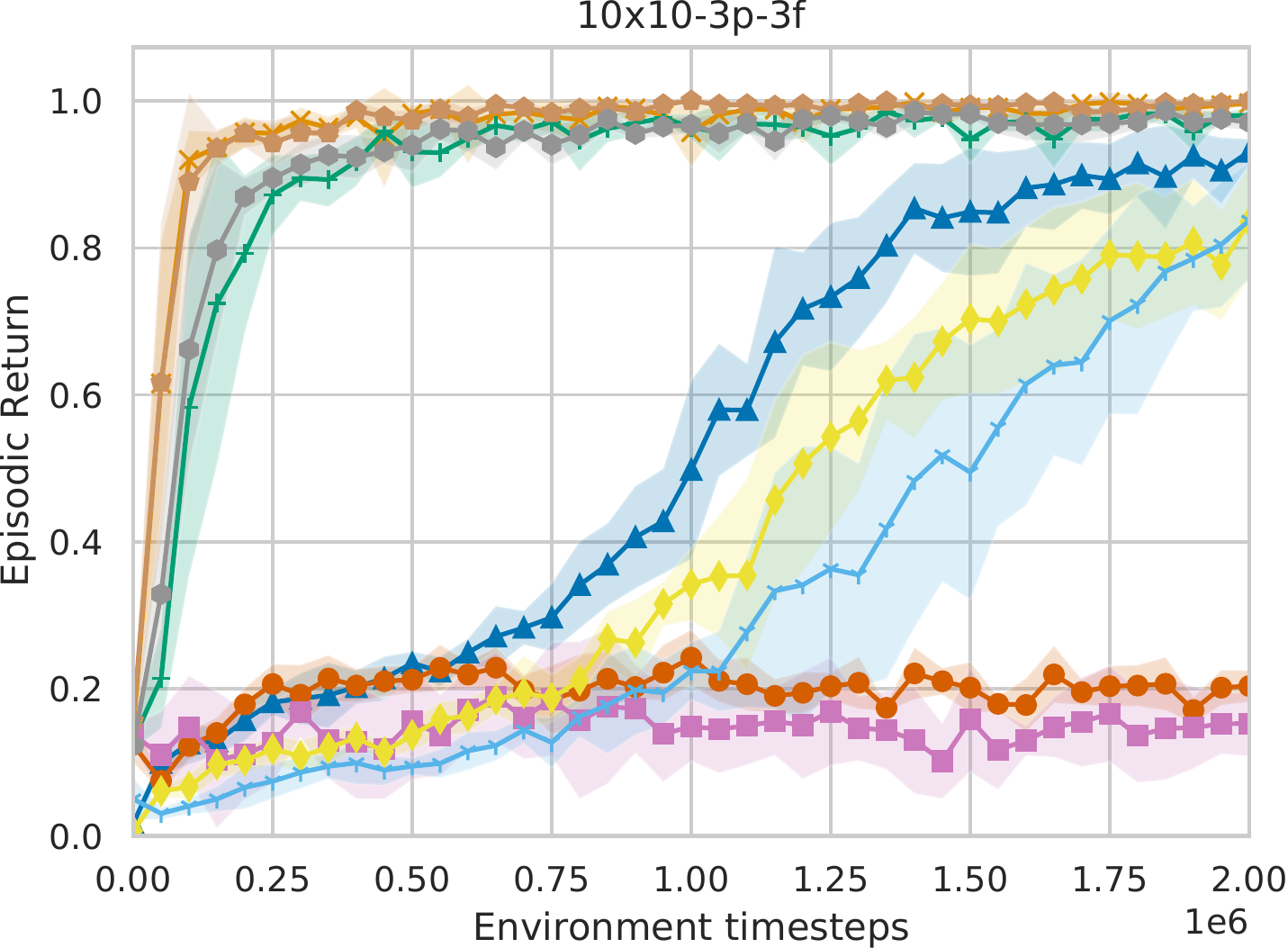}
    \end{subfigure}
     \hfill
    \begin{subfigure}{0.19\textwidth}
        \includegraphics[width=1.0\textwidth]{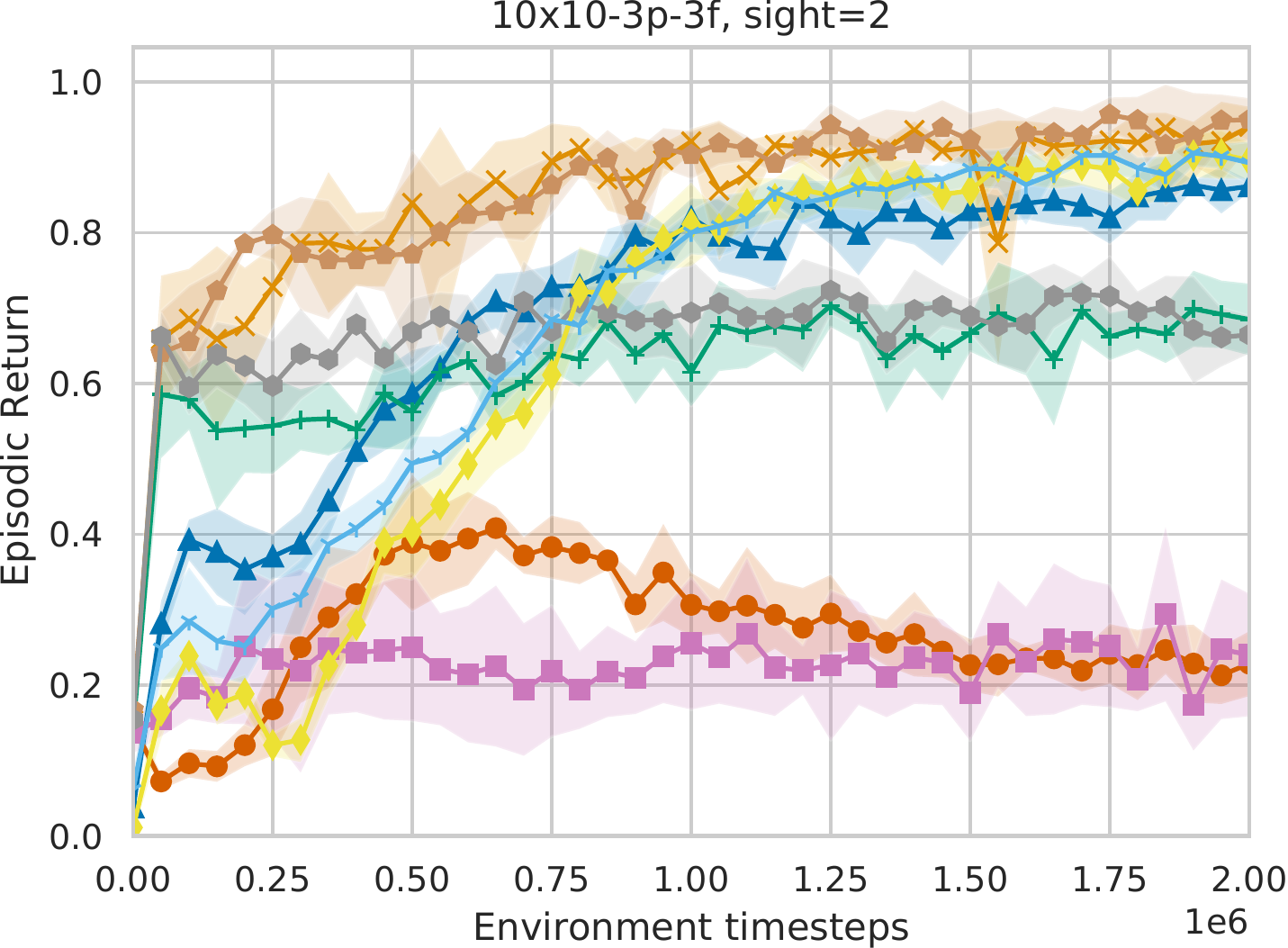}
    \end{subfigure}
     \hfill
    \begin{subfigure}{0.19\textwidth}
        \includegraphics[width=1.0\textwidth]{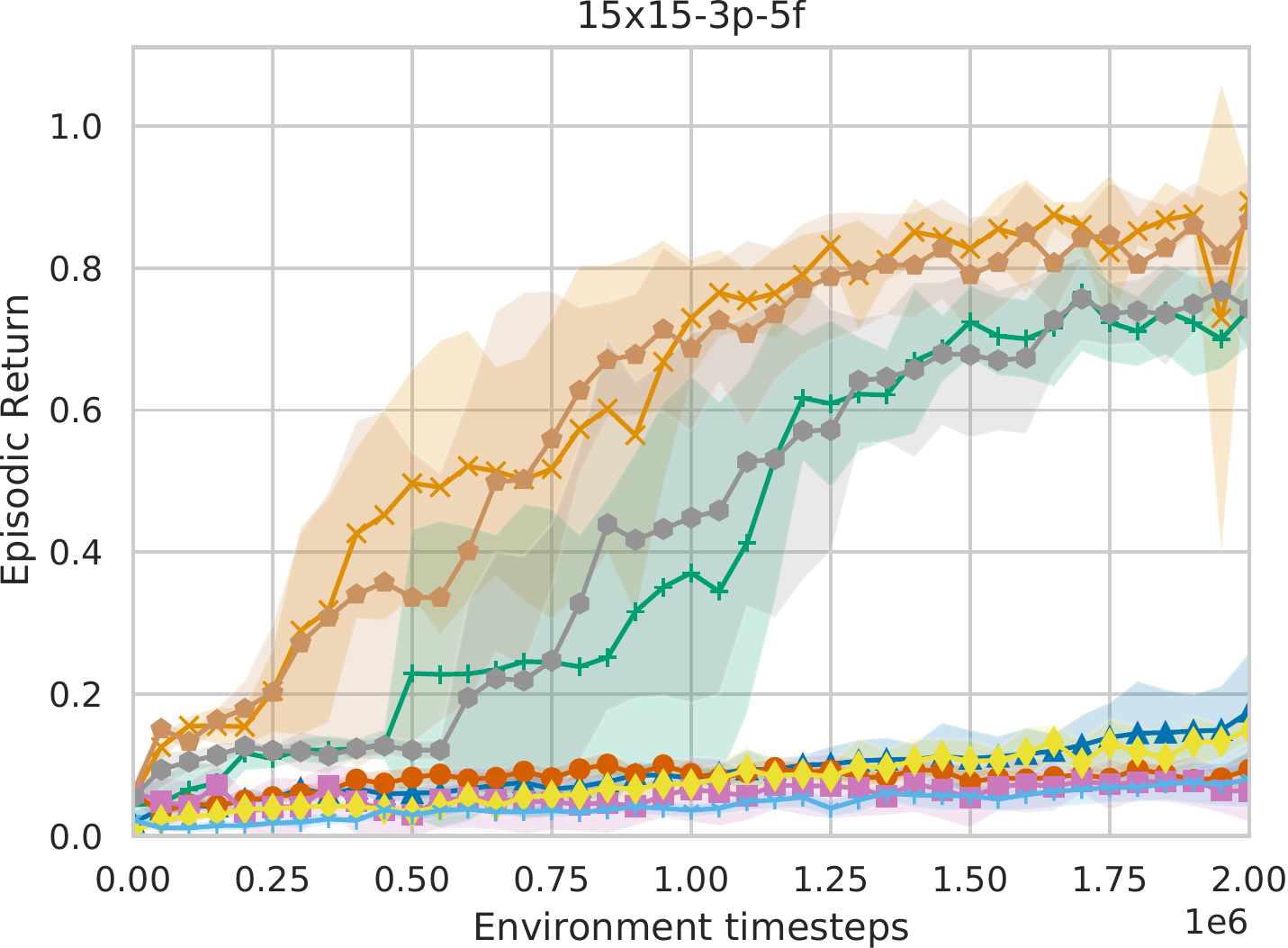}
    \end{subfigure}
    \hfill
    \begin{subfigure}{0.19\textwidth}
        \includegraphics[width=1.0\textwidth]{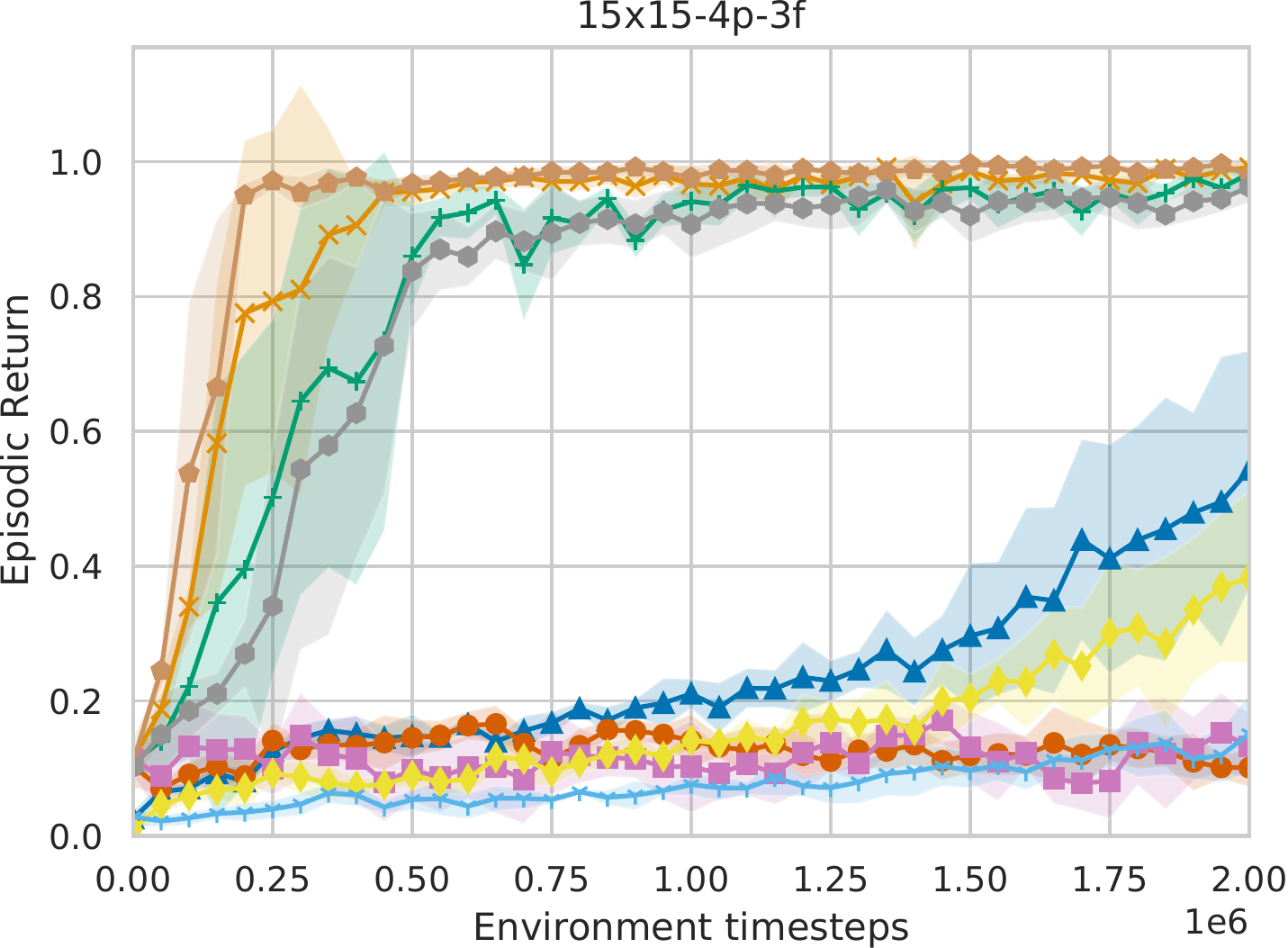}
    \end{subfigure}
    \hfill
    \begin{subfigure}{0.19\textwidth}
    \centering
        \includegraphics[width=1.0\textwidth]{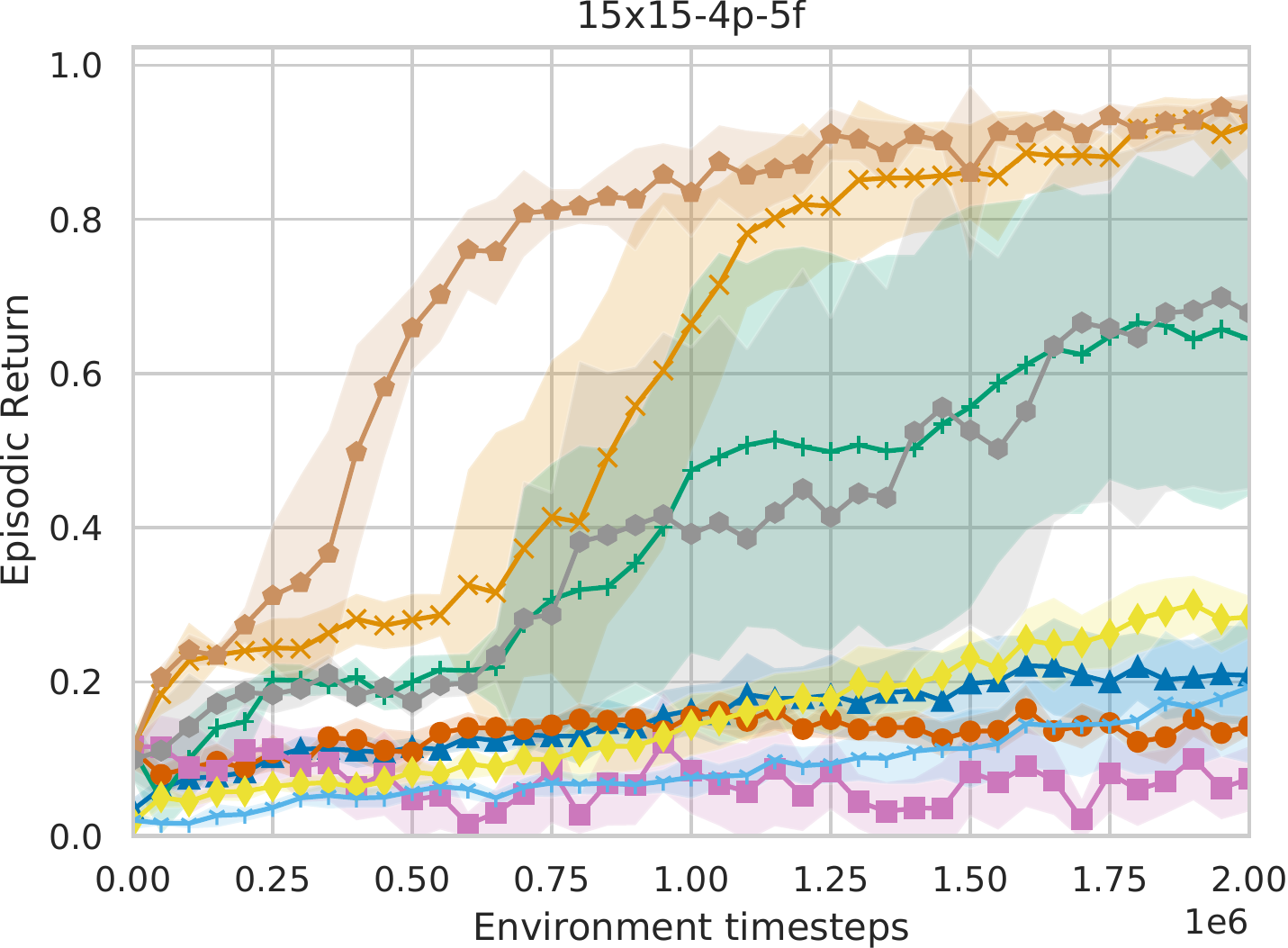}
    \end{subfigure}
    \hfill
    \begin{subfigure}{0.19\textwidth}
        \includegraphics[width=1.0\textwidth]{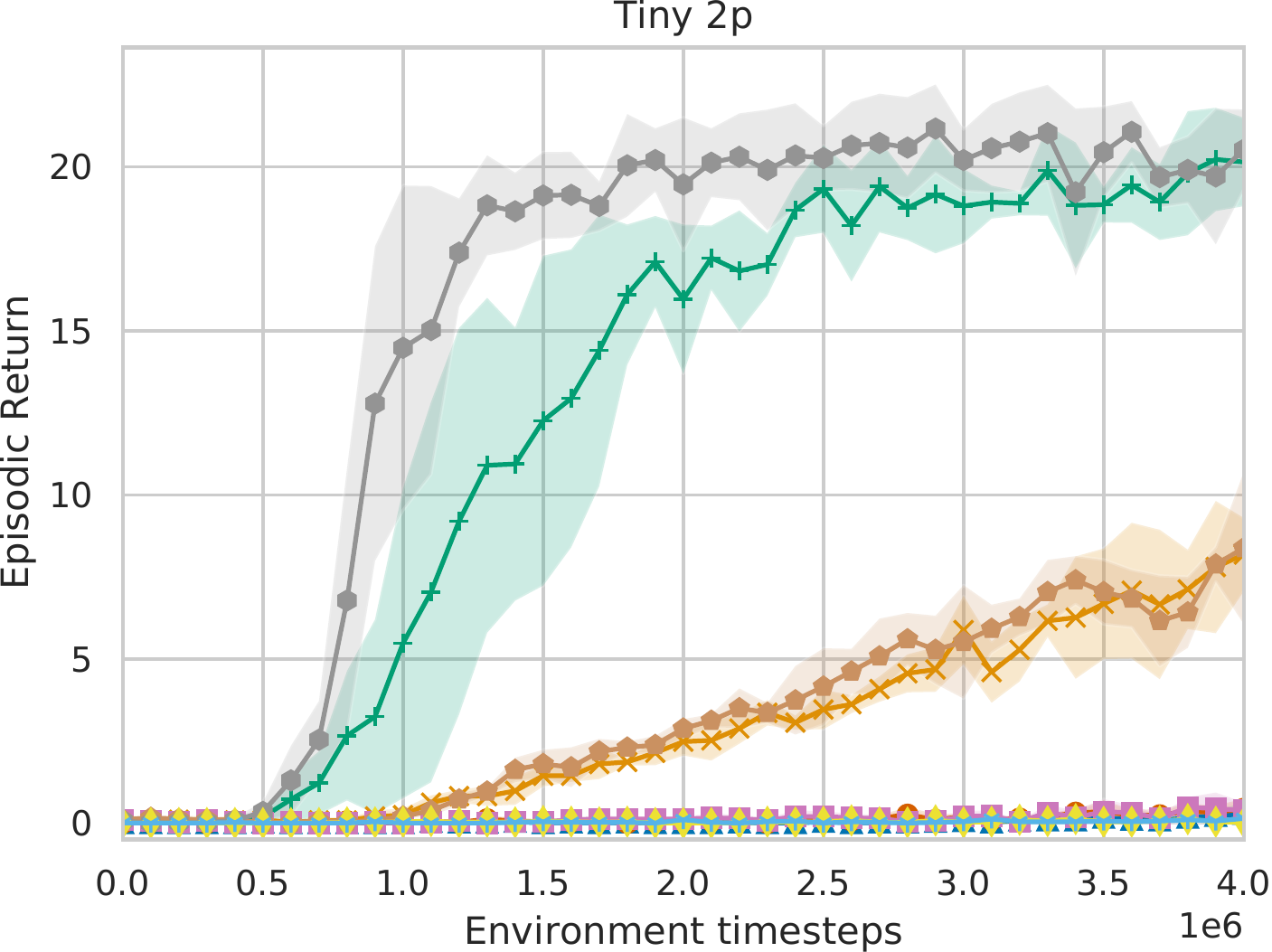}
    \end{subfigure}
    \hfill
    \begin{subfigure}{0.19\textwidth}
        \includegraphics[width=1.0\textwidth]{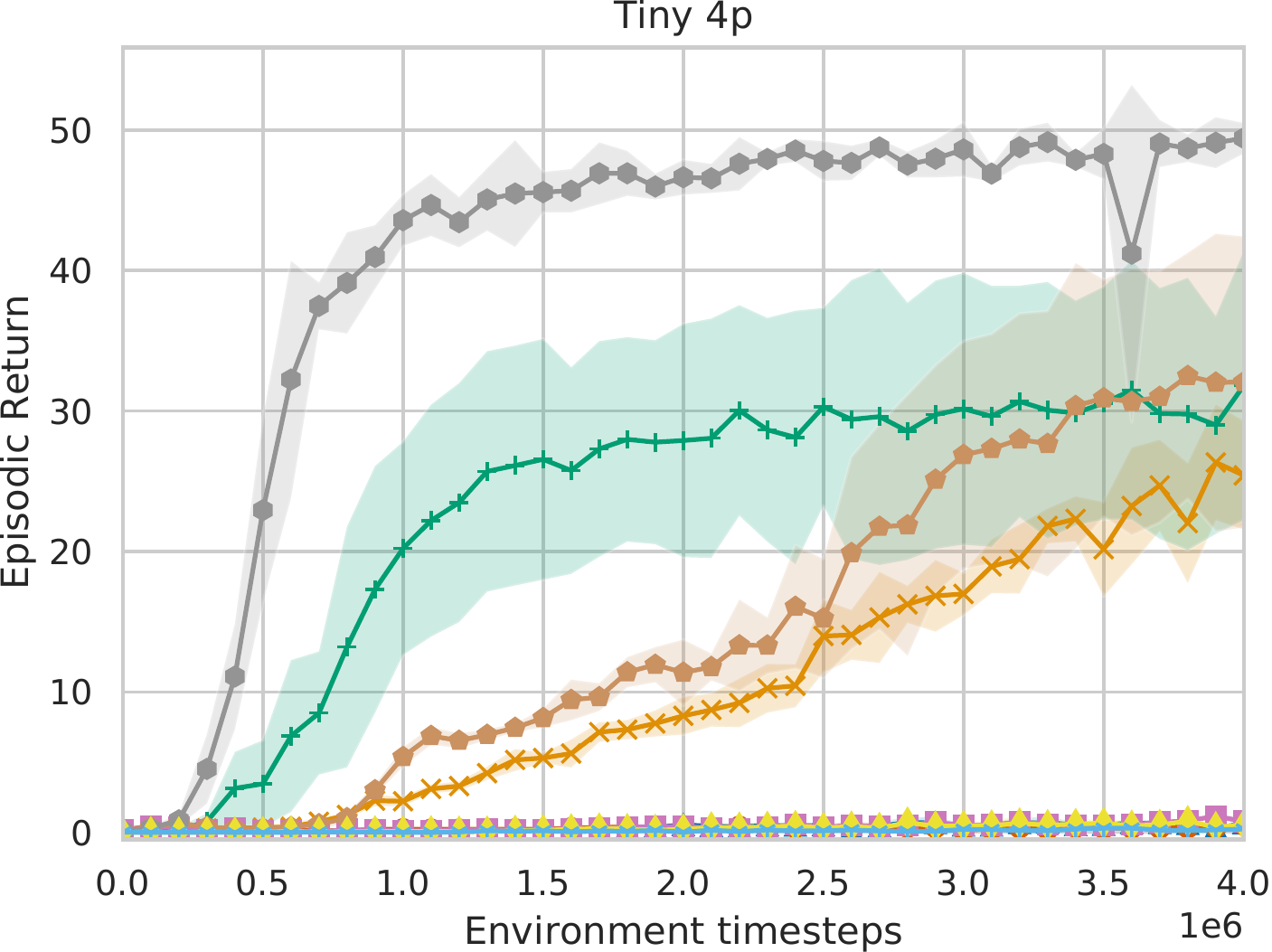}
    \end{subfigure}
    \hfill
    \begin{subfigure}{0.19\textwidth}
        \includegraphics[width=1.0\textwidth]{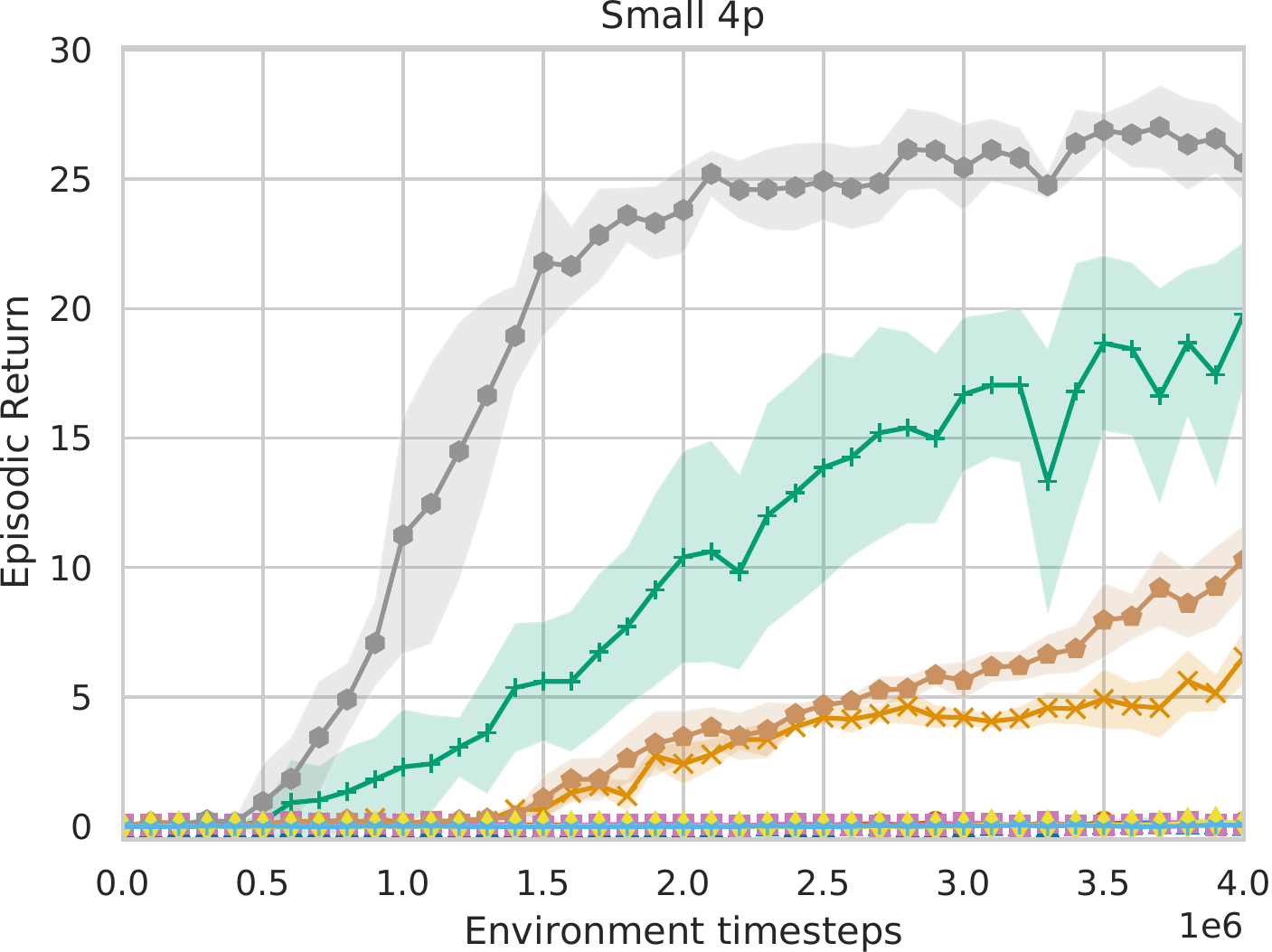}
    \end{subfigure}
   \begin{subfigure}{1\textwidth}
        \centering
        \includegraphics[trim={0cm 0cm 0.0cm, 10.1cm}, clip, width=0.4\textwidth]{images/return_plots/legends.pdf}
    \end{subfigure}

    \caption{Episodic returns of all algorithms with parameter sharing in all environments  showing the mean and the $95\%$ confidence interval over five different seeds.}
    \label{fig:performance}

\end{figure}

\Cref{fig:performance} presents the evaluation returns that are achieved during training by the nine algorithms with parameter sharing in the 25 different tasks.

\section{SMAC Win-Rates}
\label{sec:smac_win}

\begin{table}[H]

\centering
\caption{\label{tab:win_rate_smac} Maximum win-rate and 95\% confidence interval over five seeds for all nine algorithms with parameter sharing in all SMAC tasks.}
\resizebox{\linewidth}{!}{
\robustify\bf
\begin{tabular}{ l S S  S  S  S  S  S S S S}
\toprule
\textbf{Tasks \textbackslash Algs. }  &             {IQL} &           {IA2C} &             {IPPO} &              {MADDPG} &              {COMA} &              {MAA2C} &               {MAPPO} &             {VDN} &           {QMIX} \\
\midrule
2s\_vs\_1sc &   0.61(4) &   1.00(0) &   1.00(0) &  0.21(20) &  0.34(41) &  1.00(0) &   1.00(0) &   0.74(4) &   0.85(3) \\
 3s5z      &   0.39(4) &  0.72(23) &  0.17(13) &  0.15(30) &  0.81(19) &  0.99(1) &   0.96(1) &   0.92(5) &   0.94(1) \\
  corridor  &  0.44(20) &   0.80(8) &  0.82(25) &   0.00(0) &   0.00(0) &  0.00(0) &  0.68(34) &  0.44(37) &  0.53(29) \\
 MMM2      &   0.27(8) &  0.14(17) &  0.15(15) &   0.00(0) &   0.00(0) &  0.01(1) &   0.73(7) &   0.89(4) &   0.89(4) \\
 3s\_vs\_5z &  0.67(17) &   0.00(0) &  0.72(43) &   0.00(0) &   0.00(0) &  0.00(0) &  0.41(43) &  0.62(31) &  0.43(37) \\
\bottomrule
\end{tabular}
    }
\end{table}

\begin{table}[H]
\centering
\caption{\label{tab:win_rate_smac_ns} Maximum win-rate and 95\% confidence interval over five seeds for all nine algorithms without parameter sharing in all SMAC tasks.}
\resizebox{\linewidth}{!}{
\robustify\bf
\begin{tabular}{ l S S  S  S  S  S  S S S S}
\toprule
\textbf{Tasks \textbackslash Algs. }  &             {IQL} &           {IA2C} &             {IPPO} &              {MADDPG} &              {COMA} &              {MAA2C} &               {MAPPO} &             {VDN} &           {QMIX} \\
\midrule
2s\_vs\_1sc &  0.49(12) &   1.00(0) &   0.99(1) &  0.06(10) &  0.79(40) &     0.96(4) &  1.00(0) &  0.64(11) &   0.84(2) \\
3s5z      &  0.48(23) &  0.27(25) &  0.27(25) &   0.13(9) &   0.91(8) &     0.87(7) &  0.94(3) &   0.81(6) &  0.70(11) \\
corridor  &   0.05(6) &  0.31(39) &  0.17(30) &   0.00(0) &   0.01(2) &    0.06(12) &  0.01(1) &  0.08(11) &  0.40(32) \\
MMM2      &   0.04(8) &  0.06(13) &   0.00(0) &   0.00(0) &   0.00(0) &     0.00(0) &  0.00(0) &   0.58(4) &  0.23(16) \\
3s\_vs\_5z  &   0.00(0) &   0.00(0) &   0.00(0) &   0.00(0) &   0.00(0) &     0.00(0) &  0.00(0) &  0.11(22) &  0.23(24) \\
\bottomrule
\end{tabular}
}
\end{table}

\section{Hyperparameter Optimisation}
\label{sec:hyper_search}

\begin{table}[H]
    \centering
    \caption{Range of hyperparameters that was evaluated in each environment. N/A means that this hyperparameter was not optimised, and that we used one that was either proposed in the original paper or was found to be the best in the rest of the environments. If only one value is presented it means that this hyperparameter was used for all algorithms in this task.}
    \resizebox{\linewidth}{!}{\begin{tabular}{l ccccc}
    \toprule
     {} & Matrix Games & MPE & SMAC & LBF & RWARE \\
     \midrule
        hidden dimension & 64/128 &  64/128 &  64/128 &  64/128 &64/128\\
        learning rate  & 0.0001/0.0003/0.0005 & 0.0001/0.0003/0.0005 & 0.0005 & 0.0001/0.0003/0.0005 & 0001/0.0003/0.0005\\
        reward standardisation & True/False & True/False & True/False & True/False &True/False \\
        network type & FC & FC/GRU & FC/GRU & FC/GRU & FC/GRU \\
        evaluation epsilon & 0.0/0.05 & 0.0/0.05 & 0.0/0.05 & 0.0/0.05 & 0.0/0.05 \\
        epsilon anneal & 50,000/200,000 & 50,000/200,000 & 50,000 & 50,000/200,000 &  50,000/200,000\\
        target update & 200(hard)/0.01(soft) &  200(hard)/0.01(soft) & N/A & 200(hard)/0.01(soft) & 200(hard)/0.01(soft) \\
        entropy coefficient & 0.01/0.001 & 0.01/0.001 & N/A & 0.01/0.001 &  0.01/0.001\\
        n-step & 5/10 & 5/10 & N/A & 5/10 &  5/10\\
     \bottomrule 
    \end{tabular}}
    \label{tab:hyper_range}
\end{table}

The parameters of each algorithms are optimised for each environment in one of its tasks and are kept constant for the rest of the tasks within the same environment. Each combination of hyperparameters is evaluated for three different seeds.
The combination of hyperparameters that achieved the maximum evaluation, averaged over the three seeds, is used for producing the results presented in this work. \Cref{tab:hyper_range} presents the range of hyperparameters we evaluated in each environment, on the respective applicable algorithms. In general, all algorithms were evaluated in approximately the same number of hyperparameter combination for each environment to ensure consistency. To reduce the computational cost, the hyperparameter search was limited in SMAC compared to the other environments. However, several of the evaluated algorithms have been previously evaluated in SMAC and their best hyperparameters are publicly available in their respective papers.

\section{Selected Hyperparameters}

\label{sec:hyper}

In this section we present the hyperparameters used in each task. In the off-policy algorithms we use an experience replay to break the correlation between consecutive samples \citep{lin1992self, mnih2015human}. In the on-policy algorithms we use parallel synchronous workers to break the correlation between consecutive samples \citep{mnih2015human}.
The size of the experience replay is either 5K episodes or 1M samples, depending on which is smaller in terms of used memory. Exploration in Q-based algorithms is done with epsilon-greedy, starting with $\epsilon=1$ and linearly reducing it to 0.05. Additionally, in Q-based algorithms we select action with epsilon-greedy (with a small epsilon value) to ensure that the agents are not stuck. The evaluation epsilon is the hyperparameter that is optimised during the hyperparameter optimisation, with possible values between 0 and 0.05. 
In the stochastic policy algorithms, we perform exploration by sampling their categorical policy.  During execution, in the stochastic policy algorithms, we sample their policy instead of computing the action that maximises the policy. 
The computation of the temporal difference targets is done using the Double Q-learning \citep{hasselt2010double} update rule. In IPPO and MAPPO the number of update epochs per training batch is 4 and the clipping value of the surrogate objective is 0.2.

Tables \ref{tab:hyper_iql} and \ref{tab:hyper_iql_ns} present the hyperparameters in all environments for the IQL algorithm with and without parameter sharing respectively.

\begin{table}[H]
    \centering
    \caption{Hyperparameters for IQL with parameter sharing.}
    \begin{tabular}{l ccccc}
    \toprule
     {} & Matrix Games & MPE & SMAC & LBF & RWARE \\
     \midrule
        hidden dimension & 128 & 128 &128 & 128 &64\\
        learning rate  & 0.0003 & 0.0005 & 0.0005 & 0.0003 &0.0005\\
        reward standardisation & True & True &False & True &True \\
        network type & FC & FC & GRU & GRU & FC\\
        evaluation epsilon & 0.0 & 0.0 & 0.05 & 0.05 &0.05 \\
        epsilon anneal & 50,000 & 200,000 &50,000 & 200,000 &  50,000\\
        target update & 200 (hard) & 0.01 (soft) & 200 (hard) & 200 (hard) & 0.01 (soft) \\
     \bottomrule 
    \end{tabular}
    \label{tab:hyper_iql}
\end{table} 

\begin{table}[H]
    \centering
    \caption{Hyperparameters for IQL without parameter sharing.}
    \begin{tabular}{l ccccc}
    \toprule
     {} & Matrix Games & MPE & SMAC & LBF & RWARE \\
     \midrule
        hidden dimension & 64 & 128 & 64 & 64 &64\\
        learning rate  & 0.0001 & 0.0005 & 0.0005& 0.0003 &0.0005\\
        reward standardisation & True & True & True & True &True \\
        network type & FC & FC & GRU & GRU & FC\\
        evaluation epsilon & 0.0 & 0.0 & 0.05& 0.05 &0.05 \\
        epsilon anneal & 50,000 & 200,000 & 50,000 & 50,000 &  50,000 \\
        target update & 0.01 (soft) & 0.01 (soft) & 200 (hard) & 200 (hard) & 0.01 (soft) \\
     \bottomrule 
    \end{tabular}
    \label{tab:hyper_iql_ns}
\end{table}

Tables \ref{tab:hyper_iac} and \ref{tab:hyper_iac_ns} present the hyperparameters in all environments for the IA2C algorithm with and without parameter sharing respectively. 
\begin{table}[H]
    \centering
    \caption{Hyperparameters for IA2C with parameter sharing.}
    \begin{tabular}{l ccccc}
    \toprule
     {} & Matrix Games & MPE & SMAC & LBF & RWARE \\
     \midrule
        hidden dimension & 64 & 64 & 128 & 128 &64\\
        learning rate  & 0.0005 & 0.0005 & 0.0005 & 0.0005 &0.0005\\
        reward standardisation & True & True & True & True &True \\
        network type & FC & GRU & FC & GRU & FC\\
        entropy coefficient & 0.01 & 0.01 &0.01 & 0.001 &0.01 \\
        target update & 0.01 (soft) & 0.01 (soft) &0.01 (soft) & 0.01 (soft) & 0.01 (soft) \\
        n-step & 5 & 5 & 5 & 5 & 5 \\
     \bottomrule 
    \end{tabular}
    \label{tab:hyper_iac}
\end{table}

\begin{table}[H]
    \centering
    \caption{Hyperparameters for IA2C without parameter sharing.}
    \begin{tabular}{l ccccc}
    \toprule
     {} & Matrix Games & MPE & SMAC & LBF & RWARE \\
     \midrule
        hidden dimension & 128 & 128 & 64 & 64 &64\\
        learning rate  & 0.0001 & 0.0005 & 0.0005 & 0.0005 &0.0005\\
        reward standardisation & True & True & True & True &True \\
        network type & FC & FC & FC& GRU & FC\\
        entropy coefficient & 0.01 & 0.01 & 0.01 & 0.01 &0.01 \\
        target update & 200 (hard) & 0.01 (soft) & 0.01 (soft) & 0.01 (soft) & 0.01 (soft) \\
        n-step & 5 & 10 & 5 & 5 & 5 \\
     \bottomrule 
    \end{tabular}
    \label{tab:hyper_iac_ns}
\end{table}

\clearpage

Tables \ref{tab:hyper_ippo} and \ref{tab:hyper_ippo_ns} present the hyperparameters in all environments for the IPPO algorithm with and without parameter sharing respectively. 

\begin{table}[H]
    \centering
    \caption{Hyperparameters for IPPO with parameter sharing.}
    \begin{tabular}{l ccccc}
    \toprule
     {} & Matrix Games & MPE & SMAC & LBF & RWARE \\
     \midrule
        hidden dimension & 64 & 64 & 128 & 128 &128\\
        learning rate  & 0.0005 & 0.0003 & 0.0005 & 0.0003 &0.0005\\
        reward standardisation & True & True & False & False &False \\
        network type & FC &  GRU & GRU & FC & GRU\\
        entropy coefficient & 0.001 & 0.01 & 0.001 & 0.001 &0.001 \\
        target update & 0.01 (soft) & 0.01 (soft) & 0.01 (soft) & 200 (hard) & 0.01 (soft) \\
        n-step & 5 & 5 & 10 & 5 & 10 \\
     \bottomrule 
    \end{tabular}
    \label{tab:hyper_ippo}
\end{table}

\begin{table}[H]
    \centering
    \caption{Hyperparameters for IPPO without parameter sharing.}
    \begin{tabular}{l ccccc}
    \toprule
     {} & Matrix Games & MPE & SMAC & LBF & RWARE \\
     \midrule
        hidden dimension & 64 & 128 & 64 & 128 &128\\
        learning rate  & 0.0005 & 0.0001 & 0.0005 & 0.0001 &0.0005\\
        reward standardisation & True & True & True & False &False \\
        network type & FC & FC & FC & GRU & FC\\
        entropy coefficient & 0.001 & 0.01 & 0.001 & 0.001 &0.001 \\
        target update & 0.01 (soft) &0.01 (soft) &0.01 (soft) & 200 (hard) & 0.01 (soft) \\
        n-step & 5 & 10 & 10 & 5 & 10 \\
     \bottomrule 
    \end{tabular}
    \label{tab:hyper_ippo_ns}
\end{table}

Tables \ref{tab:hyper_maddpg} and \ref{tab:hyper_maddpg_ns} present the hyperparameters in all environments for the MADDPG algorithm with and without parameter sharing respectively.

\begin{table}[H]
    \centering
    \caption{Hyperparameters for MADDPG with parameter sharing.}
    \begin{tabular}{l ccccc}
    \toprule
     {} & Matrix Games & MPE & SMAC & LBF & RWARE \\
     \midrule
        hidden dimension & 128 & 128 & 128 & 64 &64\\
        learning rate  & 0.0003 & 0.0005 & 0.0005& 0.0003 &0.0005\\
        reward standardisation & True & True &  False & True &False \\
        network type & FC & GRU & GRU & FC & FC\\
        actor regularisation &  0.001 & 0.001 & 0.01 & 0.001 &0.001 \\
        target update & 200 (hard) & 200 (hard) & 0.01 (soft) & 200 (hard) & 0.01 (soft) \\
     \bottomrule 
    \end{tabular}
    \label{tab:hyper_maddpg}
\end{table}

\begin{table}[H]
    \centering
    \caption{Hyperparameters for MADDPG without parameter sharing.}
    \begin{tabular}{l ccccc}
    \toprule
     {} & Matrix Games & MPE & SMAC & LBF & RWARE \\
     \midrule
        hidden dimension & 128 & 128 & 128 & 64 &64\\
        learning rate  & 0.0005 & 0.0005 & 0.0005 & 0.0003 &0.0005\\
        reward standardisation & True & True & True & True &False \\
        network type & FC & GRU & FC & FC & FC\\
        actor regularisation & 0.001 & 0.01 & 0.001 & 0.001 &0.001 \\
        target update & 200 (hard) & 0.01 (soft) &0.01 (soft) & 200 (hard) & 0.01 (soft) \\
     \bottomrule 
    \end{tabular}
    \label{tab:hyper_maddpg_ns}
\end{table}

\clearpage

Tables \ref{tab:hyper_coma} and \ref{tab:hyper_coma_ns} present the hyperparameters in all environments for the COMA algorithm with and without parameter sharing respectively.

\begin{table}[H]
    \centering
    \caption{Hyperparameters for COMA with parameter sharing.}
    \begin{tabular}{l ccccc}
    \toprule
     {} & Matrix Games & MPE & SMAC & LBF & RWARE \\
     \midrule
        hidden dimension & 64 & 64 & 128 & 128 &64\\
        learning rate  & 0.0005 & 0.0003 & 0.0005 & 0.0001 &0.0005\\
        reward standardisation & True & True & True & True &True \\
         network type & FC & GRU & FC & GRU & FC\\
        entropy coefficient & 0.01 & 0.001 & 0.01 & 0.001 &0.01 \\
        target update & 0.01 (soft) & 200 (hard) & 0.01 (soft) & 200 (hard) & 0.01 (soft) \\
        n-step & 5 & 10 & 5 & 10 & 5 \\
     \bottomrule 
    \end{tabular}
    \label{tab:hyper_coma}
\end{table}

\begin{table}[H]
    \centering
    \caption{Hyperparameters for COMA without parameter sharing.}
    \begin{tabular}{l ccccc}
    \toprule
     {} & Matrix Games & MPE & SMAC & LBF & RWARE \\
     \midrule
        hidden dimension & 128 & 128 & 128 & 128 &64\\
        learning rate  & 0.0003 & 0.0005 & 0.0005 & 0.0001 &0.0005\\
        reward standardisation & True & True & True & True &False \\
        network type & FC & GRU & GRU & GRU & FC\\
        entropy coefficient & 0.01 & 0.01 & 0.01 & 0.001 &0.01 \\
        target update & 0.01 (soft) & 0.01 (soft) &  0.01 (soft)& 0.01 (soft) & 0.01 (soft) \\
        n-step & 10 & 10 & 5 & 5 & 5 \\
     \bottomrule 
    \end{tabular}
    \label{tab:hyper_coma_ns}
\end{table}

Tables \ref{tab:hyper_mac} and \ref{tab:hyper_mac_ns} present the hyperparameters in all environments for the MAA2C algorithm with and without parameter sharing respectively.

\begin{table}[H]
    \centering
    \caption{Hyperparameters for MAA2C with parameter sharing.}
    \begin{tabular}{l ccccc}
    \toprule
     {} & Matrix Games & MPE & SMAC & LBF & RWARE \\
     \midrule
        hidden dimension & 128 & 128 & 128 & 128 &64\\
        learning rate  & 0.003 & 0.0005 & 0.0005 & 0.0005 &0.0005\\
        reward standardisation & True & True & True & True &True \\
        network type & FC & GRU & FC & GRU & FC\\
        entropy coefficient & 0.001 & 0.01 & 0.01& 0.01 &0.01 \\
        target update & 0.01 & 0.01 (soft) & 0.01 (soft)& 0.01 (soft) & 0.01 (soft) \\
        n-step & 10 & 5 & 5 & 10 & 5 \\
     \bottomrule 
    \end{tabular}
    \label{tab:hyper_mac}
\end{table}

\begin{table}[H]
    \centering
    \caption{Hyperparameters for MAA2C without parameter sharing.}
    \begin{tabular}{l ccccc}
    \toprule
     {} & Matrix Games & MPE & SMAC & LBF & RWARE \\
     \midrule
        hidden dimension & 64 & 128 & 128 & 128 &64\\
        learning rate  & 0.0005 & 0.0003 & 0.0005& 0.0005 &0.0005\\
        reward standardisation & True & True & True & True &True \\
        network type & FC & GRU & FC & GRU & FC\\
        entropy coefficient & 0.001 & 0.01 & 0.01 & 0.01 &0.01 \\
        target update & 0.01 (soft) & 0.01 (soft) & 0.01 (soft) & 0.01 (soft) & 0.01 (soft) \\
        n-step & 10 & 5 & 5 & 5 & 5 \\
     \bottomrule 
    \end{tabular}
    \label{tab:hyper_mac_ns}
\end{table}

\clearpage

Tables \ref{tab:hyper_mappo} and \ref{tab:hyper_mappo_ns} present the hyperparameters in all environments for the MAPPO algorithm with and without parameter sharing respectively.

\begin{table}[H]
    \centering
    \caption{Hyperparameters for MAPPO with parameter sharing.}
    \begin{tabular}{l ccccc}
    \toprule
     {} & Matrix Games & MPE & SMAC & LBF & RWARE \\
     \midrule
        hidden dimension & 64 & 64 & 64 & 128 &128\\
        learning rate  & 0.0005 & 0.0005 & 0.0005 & 0.0003 &0.0005\\
        reward standardisation & True & True & False & False &False \\
        network type & FC & FC & GRU & FC & FC\\
        entropy coefficient & 0.001 & 0.01 &0.001 & 0.001 &0.001 \\
        target update & 0.01 (soft) & 0.01 (soft) &0.01 (soft) & 0.01 (soft) & 0.01 (soft) \\
        n-step & 5 & 5 & 10 & 5 & 10 \\
     \bottomrule 
    \end{tabular}
    \label{tab:hyper_mappo}
\end{table}

\begin{table}[H]
    \centering
    \caption{Hyperparameters for MAPPO without parameter sharing.}
    \begin{tabular}{l ccccc}
    \toprule
     {} & Matrix Games & MPE & SMAC & LBF & RWARE \\
     \midrule
        hidden dimension & 64 & 128 & 64 & 128 &128\\
        learning rate  & 0.0005 & 0.0001 & 0.0005 & 0.0001 &0.0005\\
        reward standardisation & True & True & True & False &False \\
        network type & FC & FC  & GRU & FC & FC\\
        entropy coefficient & 0.001 &  0.01 & 0.001 & 0.001 &0.001 \\
        target update & 0.01 (soft) & 0.01 (soft) & 0.01 (soft) & 200 (hard) & 0.01 (soft) \\
        n-step & 5 & 5 & 10 & 10 & 10 \\
     \bottomrule 
    \end{tabular}
    \label{tab:hyper_mappo_ns}
\end{table}

Tables \ref{tab:hyper_vdn} and \ref{tab:hyper_vdn_ns} present the hyperparameters in all environments for the VDN algorithm with and without parameter sharing respectively.

\begin{table}[H]
    \centering
    \caption{Hyperparameters for VDN with parameter sharing.}
    \begin{tabular}{l ccccc}
    \toprule
     {} & Matrix Games & MPE & SMAC & LBF & RWARE \\
     \midrule
        hidden dimension & 64 & 128 & 128& 128 &64\\
        learning rate  & 0.0001 & 0.0005 & 0.0005 & 0.0003 &0.0005\\
        reward standardisation & True & True & True & True &True \\
        network type & FC & FC & GRU & GRU & FC\\
        evaluation epsilon & 0.0 & 0.0 & 0.05 & 0.0 &0.05 \\
        epsilon anneal & 200,000 & 50,000 & 50,000 & 200,000 &  50,000\\
        target update & 0.01 (soft) & 200 (hard) & 200 (hard) & 0.01 (soft) & 0.01 (soft) \\
     \bottomrule 
    \end{tabular}
    \label{tab:hyper_vdn}
\end{table}

\begin{table}[H]
    \centering
    \caption{Hyperparameters for VDN without parameter sharing.}
    \begin{tabular}{l ccccc}
    \toprule
     {} & Matrix Games & MPE & SMAC & LBF & RWARE \\
     \midrule
        hidden dimension & 128 & 128 & 64 & 64 &64\\
        learning rate  & 0.0005 & 0.0005 & 0.0005& 0.0001 &0.0005\\
        reward standardisation & True & True & True & True &True \\
        network type & FC & FC & GRU & GRU & FC\\
        evaluation epsilon & 0.0 & 0.0 & 0.05 & 0.05 &0.05 \\
        epsilon anneal & 50,000 & 50,000 & 50,000 & 50,000 &  50,000\\
        target update & 0.01 (soft) & 200 (hard) & 200 (hard) & 200 (hard) & 0.01 (soft) \\
     \bottomrule 
    \end{tabular}
    \label{tab:hyper_vdn_ns}
\end{table}

\clearpage

Tables \ref{tab:hyper_qmix} and \ref{tab:hyper_qmix_ns} present the hyperparameters in all environments for the QMIX algorithm with and without parameter sharing respectively.

\begin{table}[H]
    \centering
    \caption{Hyperparameters for QMIX with parameter sharing.}
    \begin{tabular}{l ccccc}
    \toprule
     {} & Matrix Games & MPE & SMAC & LBF & RWARE \\
     \midrule
        hidden dimension & 64 & 64 & 128 & 64 &64\\
        learning rate  & 0.0003 & 0.0005 & 0.005 & 0.0003 &0.0005\\
        reward standardisation & True & True & True & True &True \\
        network type & FC & GRU & GRU & GRU & FC\\
        evaluation epsilon & 0.0 & 0.0 & 0.05 & 0.05 &0.05 \\
        epsilon anneal & 200,000 & 200,000 & 50,000 & 200,000 &  50,000 \\
        target update & 0.01 (soft) & 0.01 (soft) & 200 (hard)& 0.01 (soft) & 0.01 (soft) \\
     \bottomrule 
    \end{tabular}
    \label{tab:hyper_qmix}
\end{table}

\begin{table}[H]
    \centering
    \caption{Hyperparameters for QMIX without parameter sharing.}
    \begin{tabular}{l ccccc}
    \toprule
     {} & Matrix Games & MPE & SMAC & LBF & RWARE \\
     \midrule
        hidden dimension & 128 & 128 &64 & 64 &64\\
        learning rate  & 0.0005 & 0.0003 & 0.0005 & 0.0001 &0.0003\\
        reward standardisation & True & True & True& True &True \\
        network type & FC & GRU & GRU & GRU & FC\\
        evaluation epsilon & 0.0 & 0.0 & 0.05 & 0.05 &0.05 \\
        epsilon anneal & 50,000 & 200,000 &  50,000 & 50,000 &  50,000\\
        target update & 0.01 (soft) & 0.01 (soft) & 200 (hard) & 0.01 (soft) & 0.01 (soft) \\
     \bottomrule 
    \end{tabular}
    \label{tab:hyper_qmix_ns}
\end{table}

\end{document}